\documentclass{article}

\usepackage[nonatbib,final]{neurips_2024}
\usepackage[usenames,dvipsnames,svgnames]{xcolor}
\usepackage[linesnumbered,ruled,vlined]{algorithm2e}

\usepackage{amsthm,amsmath}
\usepackage{amsfonts}

\usepackage[english]{babel}
\usepackage{afterpage}
\usepackage{framed}
\usepackage{scalerel}
\usepackage{float}

\usepackage{graphicx}
\usepackage[labelfont=bf]{caption}
\usepackage{booktabs}

\usepackage{listings}
\usepackage{fancyvrb}
\fvset{fontsize=\normalsize}

\usepackage[numbers,sort]{natbib}
\usepackage[colorlinks,
            linkcolor = blue,
            urlcolor  = blue,
            citecolor = blue,
            linktoc=all]{hyperref}
\usepackage[nameinlink]{cleveref}

\usepackage[acronym,nowarn,nohypertypes={acronym,notation}]{glossaries}
\setacronymstyle{long-sc-short}

\lstdefinestyle{mystyle}{
    commentstyle=\color{OliveGreen},
    keywordstyle=\color{BurntOrange},
    numberstyle=\tiny\color{black!60},
    stringstyle=\color{MidnightBlue},
    basicstyle=\ttfamily,
    breakatwhitespace=false,
    breaklines=true,
    captionpos=b,
    keepspaces=true,
    numbers=left,
    numbersep=5pt,
    showspaces=false,
    showstringspaces=false,
    showtabs=false,
    tabsize=2
}
\usepackage{tikz}
\usetikzlibrary{shapes,decorations,arrows,calc,arrows.meta,fit,positioning}
\tikzset{
    -Latex,auto,node distance =1 cm and 1 cm,semithick,
    state/.style ={circle, draw, minimum width = 0.7 cm},
    detstate/.style ={rectangle, draw, minimum width = 0.7 cm, minimum height = 0.7 cm},
    point/.style = {circle, draw, inner sep=0.04cm,fill,node contents={}},
    bidirected/.style={Latex-Latex,dashed},
    el/.style = {inner sep=2pt, align=left, sloped}
}

\usepackage{wrapfig}
\usepackage{mathtools}

\usepackage{mfirstuc}
\usepackage{xspace}

\usepackage[shortlabels]{enumitem}
\setlist[enumerate]{nosep}

\DeclareRobustCommand{\mb}[1]{\ensuremath{\boldsymbol{\mathbf{#1}}}}

\DeclareRobustCommand{\KL}[2]{\ensuremath{\mb{KL}\left[#1\;\|\;#2\right]}}

\DeclareMathOperator*{\argmax}{arg\,max}

\renewcommand{\mid}{~\vert~}

\newcommand{\mba}{\mb{a}}
\newcommand{\mbb}{\mb{b}}
\newcommand{\mbc}{\mb{c}}

\newcommand{\mbv}{{\mb{v}}}

\newcommand{\mbx}{\mb{x}}
\newcommand{\mby}{\mb{y}}
\newcommand{\mbz}{\mb{z}}

\newcommand{\mbA}{\mb{A}}
\newcommand{\mbB}{\mb{B}}

\newcommand{\mbE}{\mb{E}}

\newcommand{\mbF}{\mb{F}}

\newcommand{\mbH}{\mb{H}}
\newcommand{\mbI}{\mb{I}}
\newcommand{\mbJ}{\mb{J}}

\newcommand{\mbL}{\mb{L}}

\newcommand{\mbR}{\mb{R}}
\newcommand{\mbS}{{\mb{S}}}

\newcommand{\mbU}{\mb{U}}

\newcommand{\mbY}{\mb{Y}}

\newcommand{\mbalpha}{\mb{\alpha}}
\newcommand{\mbbeta}{\mb{\beta}}

\newcommand{\mbtheta}{{\mb{\theta}}}

\newcommand{\supp}{\textrm{supp}}

\newcommand{\cN}{\mathcal{N}}

\newcommand{\cE}{\mathcal{E}}

\newcommand{\E}{\mathbb{E}}

\newcommand{\cB}{\mathcal{B}}

\newcommand{\cV}{\mathcal{V}}

\newcommand{\hy}{\hat{\mby}}
\newcommand{\hx}{\hat{\mbx}}

\newcommand{\g}{\mid}

\newtheorem{prop}{Proposition}

\newtheorem{lemma}{Lemma}

\newtheorem{thm}{Theorem}

\newtheorem*{lemma*}{Lemma}

\crefname{lemma}{lemma}{lemmas}
\crefname{prop}{proposition}{propositions}

\newcommand{\indep}{\rotatebox[origin=c]{90}{$\models$}}
\newcommand{\nindep}{\rotatebox[origin=c]{90}{$\not\models$}}

\newcommand{\kld}{\mb{KL}}

\newcommand{\xex}{\mbx_{e(\mbx)}}
\newcommand{\xoex}{\xex}

\makeatletter
\DeclareRobustCommand{\pdot}{\mathbin{\mathpalette\pdot@\relax}}
\newcommand{\pdot@}[2]{
  \ooalign{
    $\m@th#1\circ$\cr
    \hidewidth$\m@th#1\cdot$\hidewidth\cr
  }
}
\makeatother
\newacronym{ELBO}{elbo}{\emph{evidence lower bound}}
\newacronym{POPELBO}{pop-elbo}{\emph{population evidence lower bound}}

\newacronym{SVI}{svi}{stochastic variational inference}
\newacronym{BUMPVI}{bump-vi}{bumping variational inference}

\newacronym{GMM}{gmm}{Gaussian mixture model}
\newacronym{LDA}{lda}{latent Dirichlet allocation}

\newacronym{SUTVA}{sutva}{stable unit treatment value assumption}

\newacronym{KSD}{ksd}{{kernelized Stein discrepancy}}
\newacronym{KCC-SD}{kcc-sd}{kernelized complete conditional Stein discrepancy}

\newacronym{llm}{llm}{large language model}

\newacronym{OPVI}{opvi}{operator variational inference}
\newacronym{SVGD}{svgd}{Stein variational gradient descent}

\newacronym{erm}{erm}{empirical risk minimization}

\newacronym{nurd}{nurd}{Nuisance-Randomized Distillation}
\newacronym{jtt}{jtt}{Just Train Twice}
\newacronym{lff}{lff}{Learning from Failure}
\newacronym{nli}{nli}{natural language inference}
\newacronym{poe}{poe}{product of experts}
\newacronym{dfl}{dfl}{debiased focus loss}

\newacronym{pr}{patch-rnd}{patch randomization}
\newacronym{nr}{ngram-rnd}{n-gram randomization}

\newacronym{ood}{ood}{out-of-distribution}

\newacronym{scam}{b-scam}{biased-model-based spurious-correlation-avoiding method}

\newacronym{pm}{prem-mask}{premise masking}
\newacronym{rm}{roi-mask}{region-of-interest masking}

\newacronym{roi}{roi}{region-of-interest}

\newacronym{ff}{freq-filt}{frequency filtering}

\newacronym{if}{int-filt}{intensity filtering}

\newacronym{cad}{cad}{counterfactually augmented data}

\newacronym{dgp}{dgp}{data generating process}
\newacronym{REAL-X}{real-x}{real-x}
\newacronym{EVAL-X}{eval-x}{eval-x}
\newacronym{LLM}{llm}{llm}
\newacronym{AI}{ai}{ai}

\newacronym{posenc}{posi}{position-based encoding}

\newacronym{predenc}{pred}{prediction-based encoding}

\newacronym{margenc}{marg}{marginal encoding}

\newcommand{\evalx}{\acrshort{EVAL-X}}
\newcommand{\realx}{\acrshort{REAL-X}}
\newacronym{stripe-x}{stripe-x}{strongly information-penalized evaluator}
\newcommand{\detx}{\acrshort{stripe-x}}

\newacronym{KL}{KL}{KL}

\newcommand{\encmeas}{\textsc{encode-meter}}
\usetikzlibrary{bending} 

\usepackage{pifont}
\usepackage{enumitem}
\usepackage{multicol}
\usepackage{listings}

\usepackage{tcolorbox}

\pgfkeys{/pgf/decoration/.cd,
         start color/.store in=\startcolor,
         start color=black,
         end color/.store in=\endcolor,
         end color=black,
         varying line width steps/.initial=100
}
\pgfdeclaredecoration{width and color change}{initial}{
 \state{initial}[width=0pt, next state=line, persistent precomputation={
   \pgfmathparse{\pgfdecoratedpathlength/\pgfkeysvalueof{/pgf/decoration/varying line width steps}}
   \let\increment=\pgfmathresult
   \def\x{0}
 }]{}
 \state{line}[width=\increment pt,   persistent postcomputation={
   \pgfmathsetmacro{\x}{\x+\increment}
   },next state=line]{
   \pgfmathparse{ifthenelse(\x<\pgfdecoratedpathlength-5mm,varyinglw(100*(\x/\pgfdecoratedpathlength)),
    varyinglw(100*((\pgfdecoratedpathlength-5mm)/\pgfdecoratedpathlength))*(\pgfdecoratedpathlength-\x)/14) )}
   \pgfsetlinewidth{\pgfmathresult pt}
   \pgfpathmoveto{\pgfpointorigin}
   \pgfmathsetmacro{\steplength}{1.4*\increment}
   \pgfpathlineto{\pgfqpoint{\steplength pt}{0pt}}
   \pgfmathsetmacro{\y}{100*(\x/\pgfdecoratedpathlength)}
   \pgfsetstrokecolor{\endcolor!\y!\startcolor}
   \pgfusepath{stroke}
 }
 \state{final}{
   \pgfsetlinewidth{\pgflinewidth}
   \pgfpathmoveto{\pgfpointorigin}
   \pgfmathsetmacro{\y}{100*(\x/\pgfdecoratedpathlength)}
   \color{\endcolor!\y!\startcolor}
   \pgfusepath{stroke}
 }
}

\theoremstyle{plain}

\newtheorem{definition}{Definition}

\theoremstyle{remark}

\usepackage[textsize=tiny]{todonotes}
\usepackage[T1]{fontenc}
\usepackage{soul}
\usepackage{subcaption}
\AtBeginDocument{
  \DeclareFontShape{T1}{ptm}{m}{scit}{<->ssub*ptm/m/sc}{}
}

\usepackage[utf8]{inputenc}
\usepackage{booktabs} 
\usepackage{amsfonts}
\usepackage{nicefrac}
\usepackage{bbm}

\newcommand{\roar}{\textsc{roar}}
\newcommand{\fresh}{\textsc{fresh}}
\newcommand*{\encdef}{\hyperref[{def:encoding}]{Def: Encoding}} 
\newcommand{\cmark}{{\textcolor{Green}{\ding{51}}}}
\newcommand{\xmark}{{\textcolor{Red}{\ding{55}}}}
\newcommand{\vxex}{\texttt{val}(\xex)}

\title{\textbf{Explanations that reveal all through the \\ definition of encoding}}

\newcommand{\structpoint}[1]{\textit{#1}}

\author{ Aahlad Puli\thanks{Equal contribution\vspace{-17pt}}, \hspace{5pt} Nhi Nguyen$^*$, \hspace{5pt} Rajesh Ranganath \\ New York University }
\date{}

\begin{document}

\maketitle

\vspace{-8pt} 
\begin{abstract}
\vspace{-10pt} 
Feature attributions attempt to highlight what inputs drive predictive power.
Good attributions or explanations are thus those that produce inputs that retain this predictive power; accordingly, evaluations of explanations score their quality of prediction.
However, evaluations produce scores better than what appears possible from the values in the explanation for a class of explanations, called encoding explanations.
Probing for encoding remains a challenge because there is no general characterization of what gives the extra predictive power.
We develop a definition of encoding that identifies this extra predictive power via conditional dependence and show that the definition fits existing examples of encoding. 
This definition implies, in contrast to encoding explanations, that non-encoding explanations contain all the informative inputs used to produce the explanation, giving them a "what you see is what you get" property, which makes them transparent and simple to use.
Next, we prove that existing scores (\textsc{roar, fresh}, \evalx{}) do not rank non-encoding explanations above encoding ones, and develop \detx{} which ranks them correctly.
%
After empirically demonstrating the theoretical insights, we use \detx{} to show that despite prompting an \textsc{llm} to produce non-encoding explanations for a sentiment analysis task, the \textsc{llm}-generated explanations encode.

\end{abstract}

\newcommand{\nn}[1]{\textcolor{orange}{(nn: #1)}}

\vspace{-13pt} 
\section{Introduction}
\vspace{-7pt} 

Artificial intelligence can unlock information in data that was previously unknown. 
In medicine, for example, using AI, researchers have shown that electrocardiograms are predictive of structural heart conditions \citep{elias2022deep} or new-onset diabetes \citep{jethani2022new}. 
Good predictions often lead one to ask what in the input is important for a prediction; this question is a driving factor behind research in interpretability and explainability
\citep{simonyan2014visualising, lundberg2017unified}. 
One primary direction in interpretability seeks to produce explanations that are subsets of the input that retain the predictability of the label. 
These types of explanations and interpretations are called feature attributions and have been used to find factors associated with debt defaults~\citep{tran2022explainable}, to demonstrate that detecting COVID-19 from chest radiographs can rely on non-physiological signals~\citep{degrave2021ai}, and to discover a new class of antibiotics~\citep{wong2023discovery}.

Several methods exist for producing feature attributions or explanations.
While some methods compute functions of model gradients \citep{selvaraju2017grad}
or look at predictability after removing features \citep{covert2021explaining},
other methods
attribute scores to different inputs
by treating them as players in a game \citep{lundberg2017unified,jethani2022fastshap} 
or amortize their explanations by learning a single model to select subsets for each instance~\citep{yoon2018invase}. 
Choosing one from the many feature attribution methods requires an evaluation.
There are, however, many approaches to evaluation itself: qualitative ones \citep{lage2019human,saporta2022benchmarking,crabbe2022benchmarking}, which are limited to cases where humans have precise knowledge about the inputs relevant to prediction, and quantitative ones~\citep{samek2016evaluating,petsiuk1806rise,dabkowski2017real,jain2020learning,lundberg2017unified,hooker2019benchmark,jethani2021have,jethani2022new}, which do not require human knowledge.

\begin{figure*}
\vspace{-15pt}
\centering
\includegraphics[width=\textwidth]{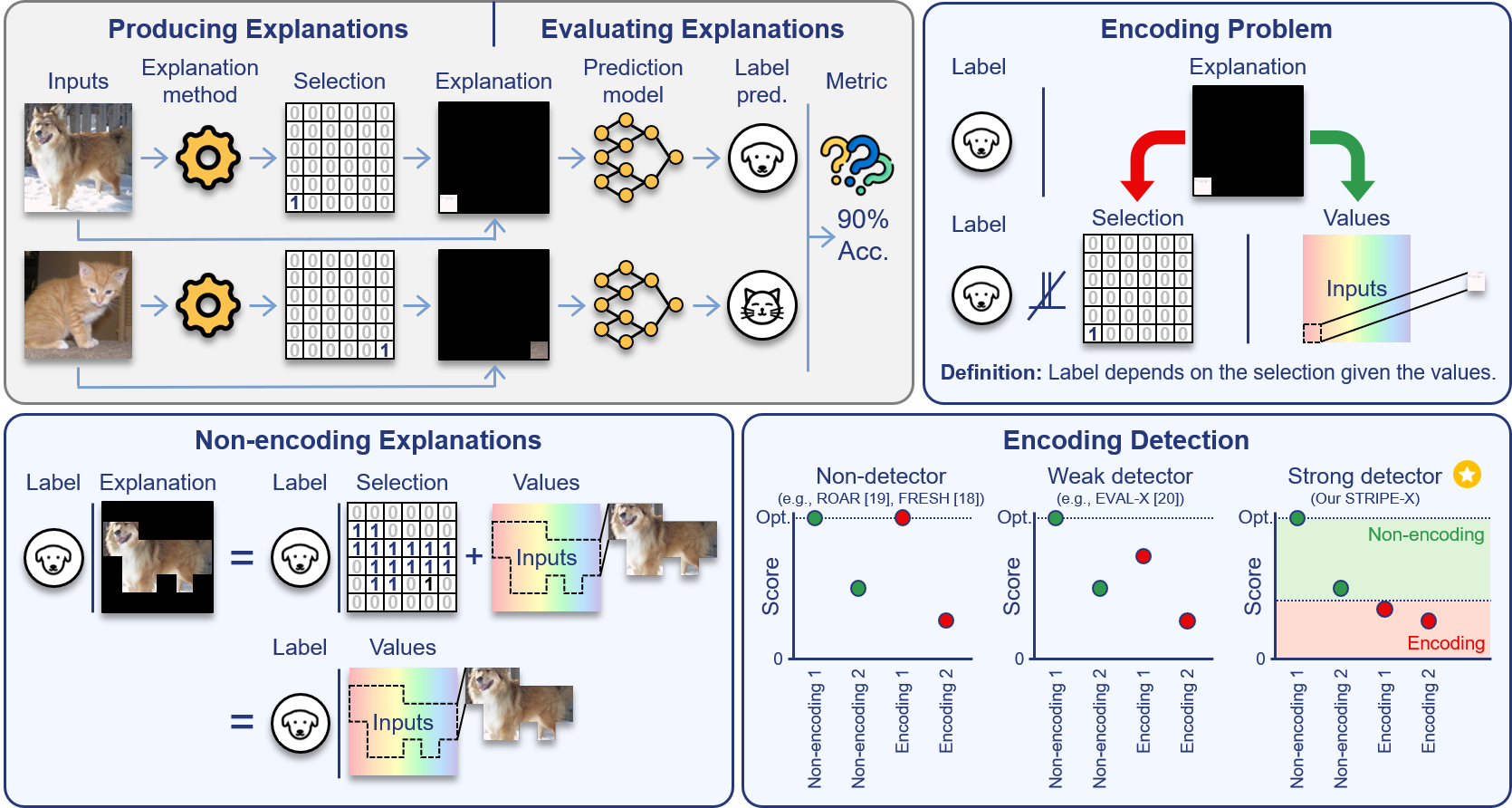}
\caption{
\small \textbf{Overview of the paper.} Explanations are produced to find inputs that are relevant to predicting a label.
However, explanations can predict the label well due to the selection being predictive of the label beyond the explanation's values.
Such explanations are called encoding.
In contrast, predicting instead from a non-encoding explanation is equivalent to predicting from the values in the explanation.
When explanations are evaluated purely based on the quality of prediction, encoding can go undetected.
We classify existing evaluations into non-detectors and weak detectors and develop a strong detector, called \detx{}.
}
\label{fig:overview}
\vspace{-10pt} 
\end{figure*}

\looseness=-1
Intuitively, a good evaluation method for feature attributions should assign higher scores to explanations that select inputs that are more predictive of the label.
However, evaluations that score explanations based on the predictability of the label from the explanation face one major challenge: \textit{encoding}.
Informally, an encoding explanation is one where the explanation predicts the label beyond what seems plausible from the values of the inputs themselves. The top left panel of \Cref{fig:overview} shows an explanation that predicts the label of dog or cat depending on whether the explanation is a pixel on the right half or left half of the image respectively.
Many explanation methods fit the description of encoding \citep{jethani2021have,hsia2023goodhart}.
Further, given that many evaluations only look at the quality of prediction, encoding can go undetected, rendering the evaluations ineffective at picking explanations.
In contrast, non-encoding explanations predict the label well only when the values in the explanation do, making them easy to reason about.

In addressing encoding, this work makes the following contributions:
\vspace{-6pt} 
\begin{itemize}[leftmargin=15pt, itemsep=0.05em]
\item \textbf{Develops a simple statistical definition of encoding} via a conditional dependence property.
\item Confirms the introduced definition captures all existing ad hoc encoding instances.
\item Shows that \textbf{non-encoding explanations are easy to use} because they retain all the predictive inputs used to build them, meaning that predictive non-encoding explanations reveal inputs that predict the label to their users, \textbf{and thus have a "what you see is what you get" property}.
\item Formalizes evaluations' sensitivity to encoding as \textit{weak detection} (optimal scoring explanations are non-encoding) and \textit{strong detection} (non-encoding explanations score above encoding ones).
\item Demonstrates that the evaluations \textsc{roar}~\cite{hooker2019benchmark} and \textsc{fresh}~\cite{jain2020learning} do not weakly detect encoding.
\item Proves that \acrshort{EVAL-X}~\citep{jethani2021have}  weakly detects encoding, but does not strongly detect encoding.
\item Develops \textbf{\acrshort{stripe-x}} and proves that it \textbf{strongly detects encoding}.
\item Uses \detx{} to show that despite prompting an \textsc{llm} to produce non-encoding explanations for a sentiment analysis task, the \textsc{llm}-generated explanations encode.
\end{itemize}
\vspace{-6pt} 
\Cref{fig:overview} provides an overview of this paper. 

\vspace{-5pt} 
\section{Evaluating explanations}
\vspace{-5pt} 

\newcommand{\nulltoken}{\texttt{null}}

We focus on explanation methods where the goal is to produce subsets of the input that predict the label~\citep{guyon2003introduction,chen2018learning}. 
Explanation methods of this form, also called feature attributions, saliency methods \citep{ribeiro2016should,selvaraju2017grad,lundberg2017unified}, or just "explanations," include thresholded rankings from Shapley values \citep{vstrumbelj2014explaining,jethani2023don}, \textsc{lime} \citep{ribeiro2016should}, and \acrshort{REAL-X} \citep{jethani2021have}.
\looseness=-1
With $\mby$ as the label and $\mbx \in \mbR^d$ as the inputs, let $q(\mby, \mbx)$ be the joint distribution over them.
An explanation method $e$ maps the inputs $\mbx$ to a binary selection mask $e(\mbx)$ over the inputs: $e: \mbR^{d} \rightarrow \{0,1\}^d$. 
The explanation $\xex$ is a pair: the \textit{selection} $e(\mbx)$ and the vector of explanation's \textit{values}.
For example, if $\mbx=[a,b,c]$ is three-dimensional and $e(\mbx)=[0, 1, 1]$, $\xex$ consists of the binary mask $e(\mbx)$ and the values associated with the inputs that correspond to the indices in $e(\mbx)$ with value $1$:
\[\xex = (e(\mbx),[b, c]).\]
We keep track of the indices because the same value can lead to different predictions depending on the index it appears at; for example, in predicting mortality from patient vital signs, a heart rate above $110$ can occur in healthy patients but a temperature of $110^\circ $F is almost always fatal.
Equivalently, like in existing work~\citep{samek2016evaluating,petsiuk1806rise,dabkowski2017real,jethani2021have}, one can choose $\xex$ to retain the values in the explanation in the same position and mask out those not selected: 
$\mbx_{e(\mbx)} = e(\mbx) \times \mbx + (1 - e(\mbx)) \times \texttt{mask-token}.$ 
For concision, we overload the word "explanation" to mean the explanation method instead of the random variable $\xex$ when it is clear from context.

Choosing between explanation methods requires evaluation. Explanation methods seek to return inputs that predict the label, so existing evaluations consider how well the explanation $\xex$ predicts the label $\mby$~\citep{samek2016evaluating,petsiuk1806rise,dabkowski2017real,jethani2021have}. 
To score explanations based on predictive power, an evaluation method  $\alpha(\cdot)$ takes as arguments both the explanation $e(\mbx)$ and the joint distribution $q(\mby, \mbx)$: $\alpha(q, e)$.
Without loss of generality let higher be better.

\subsection{Encoding: A disconnect between the predictiveness of explanations and the predictiveness of their values}\label{sec:enc-begin}

\looseness=-1 

We give a simple example of encoding to build intuition for the disconnect between predicting the label from the explanation and predicting the label from the explanation's values.
Imagine that the goal is to explain which set of vital signs signal bacterial pneumonia as the diagnosis compared to the common cold.
Consider the explanation method that selects the patient's height when the true probability of pneumonia is high given the whole set of observables (including labs, symptoms, and vital signs) and otherwise selects the patient's hair color.
Physiologically, height and hair color do not indicate that the patient has pneumonia, meaning that this explanation should not be highly predictive of the label.
However, by construction, pneumonia is likely exactly when the explanation selects height, and predicting the label from the explanation achieves the same accuracy as predicting with the full conditional $\argmax_{y\in \{\text{pneumonia, cold}\}} q(\mby=y\g \mbx)$.
Thus, despite the explanation method only selecting physiologically irrelevant inputs, the explanation predicts the label well.

Encoding examples such as the one above are neither contrived nor unique.
For example,
\citet{jethani2021have} show that certain procedures that learn to explain, when applied to MNIST digit classification, yield explanations that select a background, black pixel that predicts the label at an accuracy $>90\%$; (see Figure 1 in \citep{jethani2021have}).
Other examples of encoding explanations that predict better than what is expected from the explanation's values exist~\cite{jethani2021have,hsia2023goodhart}.
Encoding explanations should not score optimally under a good evaluation because the explanation selects inputs that do not appear to predict the label.
However, without a general characterization of the discrepancy in predictive power for encoding, finding explanations whose values predict well remains a challenge.
The next section develops a definition of encoding.

\section{Formalizing encoding}\label{sec:formalizing}

\newcommand{\econst}{e_{\texttt{const}}}

\newcommand{\xv}{\mbx_{\mbv}}
Intuitively, encoding is a phenomenon where the information about the label in the explanation $\xex$ exceeds what is known from the \textit{explanation's values}.
As the input $\mbx$ determines the explanation $\xex$, the quality of predicting the label $\mby$ from the explanation relies on the information about the label transmitted from $\mbx$ to $\xex$.
There are two pathways for this transmission; we elaborate below.

Denoting the values in a subset $\mbv$ by $\xv$, compare the event this subset takes the values $\mba$, i.e. $\xv = \mba$ to the event that the explanation's selection is $\mbv$ and that the explanation's values are $\mba$, i.e., $\xex=  (\mbv ,\mba)$.

\vspace{-2pt} 
\begin{enumerate}[leftmargin=20pt,itemsep=0.2em]
    \item Knowing that the explanation is $\xex = (\mbv, \mba)$ implies not only that the values in the explanation are determined as $\xv = \mba$, but also that the selection is determined as $e(\mbx) = \mbv$.
    \item In reverse, knowing that the values of a subset of inputs are $\xv = \mba$ and knowing the selection $e(\mbx) = \mbv$ implies that the explanation are $\xex = (\mbv, \mba)$.
\end{enumerate}

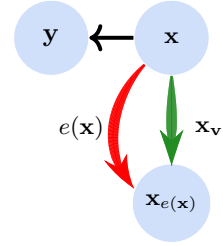
\begin{wrapfigure}[21]{R}{0.23\textwidth}
\centering
\begin{tikzpicture}[varying arrow/.style={-{Stealth[length=5mm,width=3.2mm,bend]},
color=\endcolor,
postaction={/utils/exec=\pgfsetarrows{-},decorate,decoration={width and color change}}},
circ/.style={draw=white, shape=circle, node distance=1cm,minimum width=1cm,fill=CornflowerBlue!30}
]
\begin{scope}[declare function={varyinglw(\x)=1+3*sin(1.8*\x);}]
\node[circ] (y)  {$\mby$};
\node[circ,right of=y, xshift=0.6cm] (x)  {\small$\mbx$};
\draw[->, line width=1.5pt] (x) to (y) ; 
\node[circ,below of=x,yshift=-1.2cm] (xex)  {\small$\xoex$};
\draw[varying arrow,/pgf/decoration/varying line width steps=180,
/pgf/decoration/start color=red,/pgf/decoration/end color=red] (x) 
to[out=-135,in=135,xshift=-2.1cm,yshift=0.2cm] ++ (xex);
\draw[varying arrow,/pgf/decoration/varying line width steps=180,
/pgf/decoration/start color=ForestGreen,/pgf/decoration/end color=ForestGreen] (x) 
to[out=-90,in=90,xshift=-1.6cm,yshift=0.5cm] ++ (xex);
 \node[right of=x,yshift=-1.2cm,xshift=-2.2cm]{\small${e(\mbx)}$};
 \node[right of=x,yshift=-1.2cm,xshift=-0.5cm]{\small $\xv$};
\end{scope}
\vspace{-10pt}
\end{tikzpicture}
\caption{\small 
\textbf{Intuition for encoding:}
There are two ways the information in the inputs $\mbx$ about the label $\mby$ is transmitted to the explanation $\xoex$: (1) through the values in the explanation and (2) the selection $e(\mbx)$ (in red). When the latter happens, the explanation is said to be \textit{encoding}.}
\label{fig:alt-pathways}
\end{wrapfigure}
Putting these two points together yields an equality between events:
\begin{align}\label{eq:event-eq}
    \{\mbx : \xex = (\mbv, \mba) \} = \{\mbx : e(\mbx) = \mbv\} \cap \{\mbx : \xv = \mba\}.
\end{align}

\looseness=-1
Thus, the two pathways for information between $\mbx$ and the explanation $\xex$ are the selection $e(\mbx)$ and explanation's values $\xv$; see \Cref{fig:alt-pathways}.
Existing work makes similar intuitive observations but stops short of formalizing the additional predictive power in an explanation $\xex$~\citep{jethani2021have,hsia2023goodhart}.

To formalize this extra predictive power, define the explanation indicator $\mbE_\mbv=\mathbbm{1}[e(\mbx) = \mbv]$. A little algebra in~\Cref{appsec:simplification} shows the explanation indicator $\mbE_\mbv$ provides the extra information:
\[q(\mby\g \xex = (\mbv, \mba)) = q(\mby\g \xv=\mba, \underbrace{\mbE_\mbv=1}_{\mathclap{\text{extra information in } \xex}}) \neq q(\mby \g \xv=\mba).\]
Building on this insight, we define encoding as a conditional \textit{dependence}:

\begin{definition}[\textbf{Encoding}]
    \label{def:encoding}
    The explanation $e(\mbx)$ is encoding if there exists an $\mbS$ where $q(\xex \in \mbS) > 0$ such that for every $(\mbv, \mba) \in \mbS$ :
    \begin{align}
        \mby \,\, \nindep \,\, \mbE_\mbv \g \xv = \mba.
        \label{eq:no-information-from-explanation}
    \end{align}
\end{definition}
\vskip -.05in
An example mathematical construction of an encoding explanation is provided in \Cref{sec:definition-example}.
The dependence in~\encdef{} means that for encoding explanations, there is a disconnect between how well the explanation $\xex$ predicts the label versus only the explanation's values $\xv$.
This disconnect means that evaluations that score explanations based on predictions from the explanation or their transformations~\citep{samek2016evaluating,petsiuk1806rise,dabkowski2017real,jethani2021have} can favor explanation methods that select inputs whose values have little relevance to predicting the label.

Beyond the disconnect in prediction, encoding explanations are undesirable as they conceal predictive inputs that 
nevertheless affect the explanation.
This concealment can lead to incorrect conclusions, such as that inputs outside the selection are irrelevant, or bewilderment because predictive inputs outside the explanation drive changes in the selection in ways that cannot be understood from the explanation itself. An example is provided in \Cref{sec:bewilderment-example}.

\textbf{Non-encoding explanations.} Conversely, for a non-encoding explanation, there exists no positive measure set of explanations $\xex$, where the explanation indicator has conditional dependence given the explanation's values. That is, for a set $\mbA$ where $q(\xex \in \mbA) = 1$, then for all $(\mbv,\mba) \in \mbA$
\[\mby \,\, \indep \,\, \mbE_\mbv \g \xv = \mba\]
which in turn guarantees
 \[q(\mby \g \xex=(\mbv,\mba)) = q(\mby\g \xv =\mba, \mbE_\mbv=1) = q(\mby\g \xv=\mba).
 \]
\Cref{appsec:wysiwyg} shows this. 
A simple example of a non-encoding explanation is a constant explanation that always picks the same subset of inputs, since a constant $\mbE_\mbv$ is independent of any variable.
The information for predicting the label in a non-encoding explanation lives in the explanation's values. Evaluations based on predictions from the explanation $\xex$ of non-encoding explanations will yield explanations where the input values $\xv$ predict the label.
\emph{In other words, non-encoding explanations reveal all the informative inputs they depend on, and \textbf{"what you see is what you get"} in the explanation. }

\subsection{Encoding explanations in the wild}\label{subsec:examples}

\encdef{} encompasses examples in the existing literature beyond the example in~\Cref{sec:enc-begin}.
In that example, the information about $\mby$ lies in the positions in the selection $e(\mbx)$, which motivates the name \gls{posenc}.
This section describes two other informal examples from the literature of encoding explanations, \gls{predenc}~\citep{hsia2023goodhart} and \gls{margenc}~\citep{jethani2021have}, and explains the intuition behind why they encode.
In the appendix, we develop formalizations of these types of encoding and show that these formulations meet \encdef{}.

\glsreset{predenc}
\glsreset{margenc}

\textbf{\Gls{predenc}.}
To understand how prediction-based encoding occurs, consider the task of sentiment analysis from movie reviews. Assume that reviews can either be of type "My day was terrible, but the movie was [ADJ1]." and "The movie was [ADJ2], but the day was not great." where adjective ADJ1 can be "good" or "not great" and adjective ADJ2 can be "not great" or "terrible". 
Due to common English parlance, "terrible" indicates bad sentiment more often than "not great". 
Then, in the example setup above, only seeing that the fourth word is "terrible" yields bad sentiment with higher probability than when only seeing that the phrase is "not great". 
However, the fourth word does not always describe the movie. An explanation can look at "not great" describing the movie as bad but then selects "terrible" to encode the bad sentiment. 
This explanation encodes because the selected word may not describe the movie but the selection predicts the sentiment.

\begin{wrapfigure}[16]{r}{0.5\textwidth}
\centering
\vspace{-5pt}
\def\scale{0.45}
\def\xshift{5.25*\scale}
\def\yshift{-1*\scale}
\def\ymidshift{-0.5*\scale}
\hspace{-10pt}
\begin{subfigure}[b]{0.17\textwidth}
\begin{tikzpicture}[
        >=latex',
        circ/.style={draw=white, shape=circle, node distance=1cm,minimum width=0.9cm,fill=CornflowerBlue!30},
        rect/.style={draw=black, shape=rectangle, node distance=1cm, line width=0.5pt},
        inner/.style={shape=rectangle, node distance=1cm, line width=0.0pt}]
    \node[rect,fill=red!40,minimum width=\scale cm,minimum height=2*\scale cm] at (0.5*\scale,-\scale) {};
    \node[inner] at (1.5*\scale,-0.5*\scale) 
        {\includegraphics[width=\scale cm,clip,trim={20 20 20 20}]{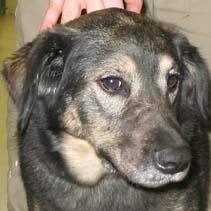}} (0.5*\scale,-\scale) {};
    \node[inner] at (1.5*\scale,-1.5*\scale) 
        {\includegraphics[width=\scale cm,clip,trim={20 60 60 20}]{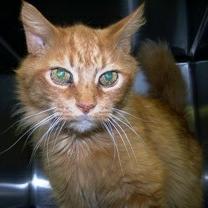}};
    \node[rect,draw,minimum width=2*\scale cm, minimum height=2*\scale cm] at (\scale,-\scale) {};
    \node[rect,fill=CornflowerBlue!30,minimum width=\scale cm,minimum height=2*\scale cm] at (3*\scale,-\scale) {};
    \node[inner] at (4*\scale,-0.5*\scale) 
        {\includegraphics[width=\scale cm,clip,trim={20 20 20 20}]{figures/dog.jpeg}};
    \node[inner] at (4*\scale,-1.5*\scale) 
        {\includegraphics[width=\scale cm,clip,trim={20 60 60 20}]{figures/cat.jpeg}};
    \node[rect,draw,minimum width=2*\scale cm, minimum height=2*\scale cm] at (3.5*\scale,-\scale) {};
    \node[inner,minimum height=6.25*\scale cm] at (\scale,-4.25*\scale) {\large cat};
    \node[inner] at (3.5*\scale,-4.25*\scale)  {\large dog};
    \draw[->, line width=1pt] (\scale,-2.25*\scale) -- (\scale,-3*\scale) {};
    \draw[->, line width=1pt] (3.5*\scale,-2.25*\scale) -- (3.5*\scale,-3*\scale) {};
\end{tikzpicture}
\vspace{-10pt}
\end{subfigure}
\begin{subfigure}[b]{0.3\textwidth}
	\begin{tikzpicture}[
        >=latex',
        circ/.style={draw=white, shape=circle, node distance=1cm,minimum width=0.9cm,fill=CornflowerBlue!30},
        rect/.style={draw=black, shape=rectangle, node distance=1cm, line width=0.5pt},
        inner/.style={shape=rectangle, node distance=1cm, line width=0.0pt}]
    \node[rect,fill=red!40,minimum width=\scale cm,minimum height=2*\scale cm] at (-0.5*\scale,-\scale) {};
    \node[inner] at (0.5*\scale,-0.5*\scale) 
        {\includegraphics[width=\scale cm,clip,trim={20 20 20 20}]{figures/dog.jpeg}};
    \node[inner] at (0.5*\scale,-1.5*\scale) 
        {\includegraphics[width=\scale cm,clip,trim={20 60 60 20}]{figures/cat.jpeg}};
    \node[rect,draw,minimum width=2*\scale cm, minimum height=2*\scale cm] at (0,-\scale) {};
    
    \node[rect,fill=black,minimum width=2*\scale cm, minimum height=2*\scale cm] at (0,-4.25*\scale) {};
    \node[rect,draw=black,minimum width=\scale cm, minimum height=\scale cm] at (-0.5*\scale,-3.75*\scale) {};
    \node[rect,draw,fill=black,minimum width=2*\scale cm, minimum height=2*\scale cm] at (2.5*\scale,-2.5*\scale) {};
    \node[rect,draw,fill=white,minimum width=\scale cm, minimum height=\scale cm] at (3*\scale,-3*\scale) {};
    \draw[-,dashed] (-1.5*\scale,0) -- (-1.5*\scale, -5.5*\scale) {};
    \node[rect,fill=CornflowerBlue!30,minimum width=\scale cm,minimum height=2*\scale cm] (blue) at (4.5*\scale,-\scale) {};
    \node[inner] at (5.5*\scale,-0.5*\scale) 
        {\includegraphics[width=\scale cm,clip,trim={20 20 20 20}]{figures/dog.jpeg}};
    \node[inner] at (5.5*\scale,-1.5*\scale) 
        {\includegraphics[width=\scale cm,clip,trim={20 60 60 20}]{figures/cat.jpeg}};
    \node[rect,draw,minimum width=2*\scale cm, minimum height=2*\scale cm] at (5*\scale,-\scale) {};
    \node[rect,draw=black,fill=black,minimum width=2*\scale cm, minimum height=2*\scale cm] at (5*\scale,-4.25*\scale) {};
    \node[rect,draw,fill=black,minimum width=2*\scale cm, minimum height=2*\scale cm] at (7.5*\scale,-2.5*\scale) {};
    \node[rect,draw,fill=white,minimum width=\scale cm, minimum height=\scale cm] at (8*\scale,-2*\scale) {};
    \draw[->, line width=1pt] (-0.2*\scale,-2.25*\scale) -- (-0.2*\scale,-3*\scale) {};
    \node[] at (0*\scale,-5.625*\scale) {\scriptsize $\xoex$};
    \draw[->, line width=1pt] (1.1*\scale,-0.75*\scale) -- (2.25*\scale,-1.25*\scale) {};
    \node[] at (2.5*\scale,-0.7*\scale) {\scriptsize $e(\mbx)$};
    \draw[->, line width=1pt] (2.25*\scale,-3.75*\scale) -- (1.1*\scale,-4.25*\scale) {};
    \draw[->, line width=1pt] (5*\scale,-2.25*\scale) -- (5*\scale,-3*\scale) {};
    \node[] at (5*\scale,-5.625*\scale) {\scriptsize $\xoex$};
    \draw[->, line width=1pt] (6.35*\scale,-0.75*\scale) -- (7.5*\scale,-1.25*\scale) {};
   \node[] at (7.75*\scale,-0.7*\scale) {\scriptsize $e(\mbx)$};
    \draw[->, line width=1pt] (7.5*\scale,-3.75*\scale) -- (6.35*\scale,-4.25*\scale) {};
    \node[inner] at (0.5*\scale,-4.75*\scale) 
        {\includegraphics[width=\scale cm,clip,trim={20 60 60 20}]{figures/cat.jpeg}};
    \node[inner] at (5.5*\scale,-3.75*\scale) 
        {\includegraphics[width=\scale cm,clip,trim={20 20 20 20}]{figures/dog.jpeg}};
\end{tikzpicture}
\vspace{-4pt}
\end{subfigure}
\vspace{-10pt}
\caption{
\small 
\textbf{Left}: Consider data where the color in the left half determines whether the label "cat", "dog") is produced from the top or bottom image on the right.
\textbf{Right:} A \acrshort{margenc} encoding explanation  that produces only the top or the bottom animal image based on the color.
The animal image alone says less about the label than knowing the animal image and the color.
Knowing the selection determines the color and thus provides additional information about the label.
}
\label{fig:encoding-diagram-marg}
\vspace{-10pt}
\end{wrapfigure}

\textbf{\Gls{margenc}.}
This type of encoding occurs when some inputs determine which other inputs determine the label.
For example, in \Cref{fig:encoding-diagram-marg}, the color determines whether the top right patch produces the label or the bottom right patch.
Inputs that \textit{control} where the label comes from are named \textit{control flow inputs}.
For a real-world example, consider the following example from \citet{jethani2021have}, where the goal is to predict mortality for patients with chest pain.
A lab value that checks for heart injury and acts like a control flow input is troponin.
Abnormal troponin indicates that cardiac issues exist and cardiac imaging would inform mortality.
Normal troponin on the other hand can indicate that chest pain is unrelated to cardiac health and a chest X-ray would instead inform mortality.
Selecting one image or the other, but not the control flow input, conceals information about why the image was relevant to the label.

\textbf{Formalization.}
In \Cref{appsec:example-proofs}, we provide mathematical formulations of each informal example and show that they fall under the definition of encoding in \encdef{}: position-based encoding (\Cref{sec:position-based-example}), prediction-based encoding (\Cref{sec:prediction-based-example}), and marginal encoding (\Cref{appsec:margenc}).
The key intuition behind all of these is that the explanation $e(\mbx)$ varies with inputs other than the selected ones, and these additional inputs provide information about the label beyond the selected ones.
Next, we turn to detecting encoding via quantitative evaluations.

\vspace{-5pt}
\section{Detecting encoding in explanations}
\label{sec:eval-x}
\vspace{-5pt} 
This section develops notions of sensitivity to encoding for evaluation methods, and uses the mathematical definition of encoding developed in the previous section to establish which methods detect encoding and which do not.
\citet{hsia2023goodhart} suggest that evaluation methods like \acrshort{EVAL-X} can be gamed to produce high scores for encoding explanations by optimizing the evaluation.
To study this case, we introduce the notion of \textit{weak detection.}
If the optimal score of an evaluation of explanations does not permit encoding, then that evaluation is said to weakly detect encoding:
\begin{definition}[\textbf{Weak detection of encoding}]
    An evaluation $\alpha(q, e)$ of explanations \textit{weakly detects encoding} if the optimal explanations $e^*$, i.e. $\,\, \alpha(q, e^*) = \max_e \alpha(q,e)$, are non-encoding.
\end{definition}
Weak detection provides a recipe for finding non-encoding explanations: find the explanation that achieves the maximum score of a weak detector.
However, such a recipe would only work when optimizing without constraints because weak detection does not require non-encoding explanations to have a better score than any encoding one. 
Requiring this leads to the definition of \textit{strong detection}. 
\begin{definition}[\textbf{Strong detection of encoding}]
    An evaluation $\alpha(q,e)$ \textit{strongly detects encoding} if for any encoding explanation $e$ and non-encoding explanation $e'$, $\alpha(q,e^\prime ) > \alpha(q,e)$.
\end{definition}
Evaluations that are not weak detectors cannot be strong detectors because they score some encoding explanation optimally.
\subsection{Do existing evaluation methods detect encoding?}\label{sec:weak}
Here, we consider whether several techniques for evaluating explanations: \roar{}~\citep{hooker2019benchmark}, \textsc{fresh}~\citep{jain2020learning}, and \acrshort{EVAL-X}~\citep{jethani2021have} can detect encoding. 
We analyze these evaluations on the following distribution $q$ 
\begin{align}
    \mbx & = [\mbx_1, \mbx_2, \mbx_3]  \sim \cB(0.5)^{\otimes 3 }, \qquad \quad
       \mby = \begin{cases}
                     \mbx_1 \quad \text{w.p. } 0.9 \quad  \text{else} \quad 1-\mbx_1 \quad  \text{ if }  \mbx_3 = 1,
                          \\
                    \mbx_2 \quad \text{w.p. } 0.9 \quad  \text{else} \quad 1- \mbx_2 \quad  \text{ if }  \mbx_3 = 0.
                \end{cases}  
    \label{eq:sim-example-main}
\end{align}
Consider the explanation $e_\text{encode}(\mbx) = \xi_1=[1,0,0]$ if $\mbx_3=1$ and $\xi_2=[0,1,0]$ otherwise; this encodes because $\mbx_3$ is used to create the explanation and $\mbx_3$ predicts the label conditional on $\mbx_1$ when $\mbE_{\xi_1}=1$.
This is a \gls{margenc} explanation (see~\Cref{subsec:examples}).

\paragraph{\textsc{roar} and \textsc{fresh} do not weakly detect encoding.}
\textsc{roar} evaluates explanations by predicting the label from the inputs not selected by the explanation, denoted as $\mbx_{-e(\mbx)}$; \textsc{roar} scores explanations optimally if the predictions from the remaining covariates are as random as predicting without any covariates at all.
In other words, \textsc{roar} checks how informative $\mbx_{-e(\mbx)}$ is of $\mby$ and provides the highest score when $\mby \indep \mbx_{-e(\mbx)}$. 
In contrast, \textsc{fresh} evaluates explanations by predicting the label from the explanation after removing all other inputs, denoted as $\vxex$. 
For example, assume we are given an input $\mbx =$ "Visually stunning. My favorite movie ever" and an explanation $e(\mbx)$ that selects the words "stunning" and "favorite". 
Then, the explanation is $\xex = ([0, 1, 0, 1, 0, 0], [\text{"stunning"}, \text{"favorite"}])$, whereas $\vxex = [\text{"stunning"},  \text{"favorite"}, \texttt{pad-token}, \texttt{pad-token}, \texttt{pad-token}, \texttt{pad-token}]$, which drops the information about where the selected words are in the input. See \Cref{appsec:props-roar-fail} for a formal definition of $\vxex$.
\textsc{fresh} 
checks how predictive 
$q(\mby \g \vxex{})$ 
is and assigns an optimal score if the prediction 
is as good as that of $q(\mby\g \mbx)$.
These conditions hold for $e_\textrm{encode}(\mbx)$ in \cref{eq:sim-example-main}:
\newcommand{\proproarfail}{
For the \gls{dgp} in~\cref{eq:sim-example-main}, \textsc{roar} and \textsc{fresh} assign their respective optimal scores to the encoding explanation ${e_\textrm{encode}(\mbx)}$.
}

\begin{prop}\label{prop:roar-indep}
\proproarfail{}
\end{prop}
The proof is in \Cref{appsec:props-roar-fail}.
The intuition is that the encoding explanation $e_\textrm{encode}(\mbx)$ always selects the input that informs the label given the control flow $\mbx_3$; removing the only conditionally informative input means that $\mbx_{-e_\textrm{encode}(\mbx)}$ has no information about $\mby$.
In turn, \textsc{roar} scores an encoding explanation $\mbx_{-e_\textrm{encode}(\mbx)}$ optimally, meaning it does not even weakly detect encoding.
In addition, $\vxex{}$ provides the exact same information about the label regardless of which position it came from. 
As a result, $\mbx \indep \mby \g \vxex{}$, so \fresh{} scores $e_{\text{encode}}(\mbx)$ optimally.
Even though \fresh{} attempts to drop the information about the selection $\mbv = e(\mbx)$ during evaluation, $\vxex$ remains a function of $\xex = (\mbv, \mba)$, so extra information can still be transmitted through the selection $\mbv$.
Thus, \textsc{roar} and \textsc{fresh} are not weak detectors of encoding.

\paragraph{\acrshort{EVAL-X} weakly detects encoding but not strongly.}
\acrshort{EVAL-X}~\citep{jethani2022fastshap,jethani2023don} is an evaluation method and is sometimes called the surrogate model score.
The \acrshort{EVAL-X} score with log-probabilities is
\begin{align}
    \acrshort{EVAL-X}(q, e) :=  \E_{(\mbv, \mba) \sim q(\xex)}\E_{q(\mby \g \xex = (\mbv, \mba))}\left[\log q\left(\mby \g \mbx_\mbv =\mba\right)\right].
    \label{eq:eval-x-log-prob}
\end{align}
This score measures the expected log-likelihood of the labels given the input values chosen by the explanation method $e$ and is grounded in the sampling distribution $q$. Log-likelihoods are maximized by matching the true distribution, this leads to \acrshort{EVAL-X}'s weak detection:

\newcommand{\optencthm}{If $e(\mbx)$ is \acrshort{EVAL-X} optimal, 
then $e(\mbx)$ is not encoding.}

\begin{thm}
\label{thm:opt-removes-encoding}
\optencthm{}
\end{thm}
\Cref{sec:proof-eval-x} gives a proof. The proof 
shows that at optimality, the prediction from the \textit{values} of explanation has to match the prediction from the full inputs.
In turn, given the values there is no additional information in $\mbx$ about $\mby$, which means the explanation indicator $\mbE_\mbv$ is independent of $\mby$; this violates \encdef{}, which proves the non-encoding nature of \evalx{-}optimal explanations.

To test strong detection for \acrshort{EVAL-X}, we consider explanations constrained to select one input.
Such reductive constraints appear in practice because the goal of producing an explanation is often to aid humans who benefit from reduced complexity. 
Such constraints prohibit explanations from reaching \acrshort{EVAL-X}'s optimal score. Compare $e_{encode}(\mbx)$ with a non-encoding constant explanation:
\newcommand{\proptwo}{Let $e_{\mbc}(\mbx) = \xi_3$. Then, for the \gls{dgp} in~\cref{eq:sim-example-main}, $\evalx{}(q,e_{\textrm{encode}}) > \evalx{}(q,e_{\mbc})$.}
\begin{prop}\label{prop:evalx-weak}
\proptwo{}	
\end{prop}
Thus, \acrshort{EVAL-X} is not a strong detector. The intuition is that the first two coordinates $\mbx_1$, $\mbx_2$ predict the label when selected by $e_\textrm{encode}$, while the control flow feature does not predict the label.
\evalx{} not being a strong detector means that optimizing \acrshort{EVAL-X} over a reductive set may yield an encoding explanation. In this case, $e_\textrm{encode}$ is one of the \evalx{-}optimal reductive explanations (\Cref{lemma:optimal-reductive}).

\subsection{\texorpdfstring{\acrshort{stripe-x}}{}: a strong detector of encoding}

Encoding explanations induce the dependence between the label $\mby$ and the identity of the selection $\mbE_\mbv=\mathbbm{1}[e(\mbx)=\mbv]$ given the values in the explanation $\mbx_\mbv$ (\encdef{}). This dependence can be tested for by building on conditional independence tests~\citep{zhang2011kernel, sudarshan2020deep, sudarshan2023diet}. Rather than testing, direct quantification of dependence can be useful for when combining with other scores, which can be done using instantaneous conditional mutual information:
\begin{align}
    \label{eq:defonepen}
    \phi_q(e) :=  \E_{(\mbv, \mba) \sim q(\mbx_{e(\mbx)})}\mbI\left( \mbE_\mbv; \mby \g \mbx_\mbv = \mba  \right) \quad (\encmeas{}). 
\end{align}

\encmeas{} is $0$ only when \encdef{} does not hold:
\newcommand{\encmeaszero}{\encmeas{} $\phi_q(e)=0$ if and only if $e$ is not encoding.} 
\begin{prop}\label{prop:encmeaszero}
\encmeaszero{}
\end{prop}
\vskip -.05in

\glsreset{stripe-x}
\begin{wraptable}[13]{r}{4.55cm}
\vspace{-12pt}
{\small 
\begin{tabular}{lcc}
\toprule
 Method 		 & Weak & Strong\\
\midrule
\textsc{roar}~\citep{hooker2019benchmark} & \xmark & \xmark\\
\textsc{fresh}~\citep{jain2020learning} & \xmark & \xmark \\
\acrshort{EVAL-X}~\citep{jethani2021have} &\cmark  & \xmark \\
\midrule
\acrshort{stripe-x} & \cmark  & \cmark \\
\bottomrule
\end{tabular}
\vspace{-5pt}
\caption{\small The weak and strong detection properties of different evaluation methods. Existing scores like 
\textsc{roar}~\citep{hooker2019benchmark} and \textsc{fresh}~\citep{jain2020learning}, are not weak detectors, which in turn means they are not strong detectors either.}
\label{tab:encoding-comparison}
}
\end{wraptable}
The proof is in \Cref{appsec:hy}.
Combining \evalx{} with \encmeas{} weighed by $\alpha$ yields a method we call the \gls{stripe-x}:
\begin{align}
\label{eq:stripe-x-main}
\detx{}_\alpha(q,e) := \evalx{(q,e)}
  - \alpha \phi_q(e).
\end{align}
For a large enough $\alpha$, the added penalty term pushes down the scores of encoding explanations below that of all non-encoding ones, meaning that \detx{} is a strong detector of encoding:
\newcommand{\strongdet}{
With finite $\mbH(\mby \g \mbx)$ and $\mbH(\mby)$,
for any explanation that encodes $e$ and any that does not encode $e^\prime$,
there exists an $\alpha^*$ such that $ \forall \alpha > \alpha^*
\,\, \acrshort{stripe-x}_\alpha(q, e^\prime) \,\,
				> 
			\acrshort{stripe-x}_\alpha(q, e).$
}

\begin{thm}
\label{thm:strong-det}
	\strongdet{}
\end{thm}
The proof is in \Cref{appsec:hy}.
The intuition behind the proof is that for a large enough $\alpha$, the \detx{} scores for any encoding explanations will be dominated by the information term, and thus will become smaller than any non-encoding explanation whose score is lower bounded by the negative marginal entropy, $-\mbH_q(\mby)$.
\Cref{tab:encoding-comparison} summarizes the weak and strong detection properties of different evaluations.

\textbf{Estimating \acrshort{stripe-x}.}
The first component of \detx{} is \evalx{}.
Computing \acrshort{EVAL-X} (\cref{eq:eval-x-log-prob}) requires an estimate of the predictive distribution of the label $\mby$ given $\mbx_\mbv$, $q(\mby \g \mbx_\mbv)$ \citep{jethani2021have}. Estimation can be done in two ways. The first way makes use of a surrogate model trained to predict the label from different random subsets using masked tokens \citep{jethani2021have, covert2021explaining}. The second way to compute \acrshort{EVAL-X} (\cref{eq:eval-x-log-prob}) relies on conditional generative models \citep{brown2020language,ramesh2022hierarchical}.
Both hyperparameters and a combination of the estimators can be chosen to maximize the average log-likelihood on a held-out validation set across random input subsets.

To estimate the second part of \detx{}, the \encmeas{}, 
first expand the mutual information terms in \encmeas{}, $\phi_q(e)$, in terms of expected $\kld$:
\begin{align}
\label{eq:klform-detx}
	\phi_q(e) =  \E_{(\mbv, \mba) \sim q(\mbx_{e(\mbx)})}
	\E_{\mby \sim q(\mby\g \mbx_\mbv = \mba)}
\KL{q(\mbE_\mbv\g \mbx_\mbv=\mba,\mby)}{q(\mbE_\mbv\g \mbx_\mbv=\mba)}.
\end{align}
The outer expectation can be estimated using samples from the data and the inner expectation over $\mby$ can be estimated using the \acrshort{EVAL-X} model $q(\mby \g \mbx_\mbv)$.
The distributions over $\mbE_\mbv$ can be estimated using a classifier of $\mbE_\mbv$ that randomly masks the label and masks different subsets of the inputs.
Further details and a generative way to estimate \detx{} are in \Cref{appsec:detxestimate} and \Cref{appsec:gen-way-desc}; full algorithms are given in~\Cref{appsec:algorithm-boxes}.

\textbf{\detx{} in practice.}
Using \detx{} to choose between explanations is straightforward: pick the one with the larger score.
However, like other evaluations that use learned models, misestimation can pose a problem. With large $\alpha$, non-encoding explanations with misestimated \encmeas{} will have bad \detx{} scores, while with small $\alpha$ some encoding explanations can have good scores. Across all experiments, we set $\alpha=20$, which yielded \detx{} scores for known encoding explanations worse than known non-encoding explanations.

\section{Experiments}\label{sec:experiments}
This section consists of two parts. 
The first part demonstrates the weak and strong detection capabilities of the evaluations \roar{}, \evalx{}, and \detx{} in a simulated setting and on an image recognition task.
To demonstrate these capabilities, we run these evaluations on instantiations of \gls{posenc}, \gls{predenc}, and \gls{margenc}.
Additionally, we evaluate an existing method that learns to explain under a reductive constraint, called \realx{}~\citep{jethani2021have}.
The second part shows how \detx{} enables discovering encoding explanations in the wild, without specific knowledge of the \gls{dgp} or the method that produced the explanation.
We employ \detx{} to uncover encoding in explanations generated by a \gls{llm} for predicting sentiments from movie reviews.

\subsection{Empirically studying the detection of encoding in a simulated setting}
We construct two examples with binary labels $\mby$: one discrete input $\mbx$ and one that is a hybrid of continuous and discrete components.
Both use one binary input in $\mbx \in \{0, 1\}^5$ as a control flow variable and switch the inputs that $\mby$ depends on.
In both \glspl{dgp}, $\mby$ only depends on $\mbx_1$ if $\mbx_3=1$, and only on $\mbx_2$ if $\mbx_3=0$; this means that $\mbx_4,\mbx_5$ are purely noise.
For both \glspl{dgp}, $\mby$ is sampled per the following distribution where $\mbx_3$ determines the subset the $\mby$ depends on
\begin{align}
q(\mby = 1 \g \mbx)  = \mathbbm{1}[\mbx_3=1] q(\mby \g \mbx_1, \mbx_3) + \mathbbm{1}[\mbx_3=0]q(\mby \g \mbx_2, \mbx_3).
\label{eq:experiment-simulation}
\end{align}
Thus, \acrshort{EVAL-X}$^*$ is achieved by an explanation of size $2$: $e(\mbx) = \xi_1 + \xi_3$ if $\mbx_3=1$ else $e(\mbx) = \xi_2 + \xi_3$.
See \Cref{appsec:sim-enc-exp-details} for details; the exact \acrshortpl{dgp} are given in \cref{eq:bern-dgp} and \cref{eq:cnts-dgp}.

\textbf{Encoding explanations.}
\Cref{tab:def-encoding} describes the encoding explanations we consider for this setting.
In~\Cref{appsec:sim-enc-exp-details}, we check that \encdef{} holds for these explanations in the discrete \gls{dgp} by estimating the role of the unselected inputs in affecting the explanation and the role of $\mbE_\mbv$ in predicting $\mby$ beyond $\mbx_\mbv$; a characterization of \encdef{} to support this check is in \Cref{lemma:defeq}.
\setlength{\tabcolsep}{4pt}
\begin{table}[t]
\centering
\vspace{-20pt}
\small
\begin{tabular}{c|c|c} \textbf{\gls{posenc}} 
& \textbf{\gls{predenc}}  
& \textbf{\gls{margenc}} 
	\\
\midrule
$e(\mbx) 
	 = \begin{cases}
	\xi_4  \text{ if }  \pi(\mbx) > 0.5,
	\\
	\xi_5  \text{ else}.
	\end{cases}$
&
$e(\mbx) 
	 = \begin{cases}
	 \underset{M:|M|\leq 1}{\argmax}
	\,\pi(\mbx_M)  \, \text{ if }  \, \pi(\mbx) > 0.5,
	\\
	 \underset{M:|M|\leq 1}{\argmax}\, 1 - \pi(\mbx_M)  \, \text{ else}.
	\end{cases}$
& 
$e(\mbx) 
	 = \begin{cases}
	\xi_1    \text{ if }  \mbx_3=1,
	\\
	\xi_2   \text{ else}.
	\end{cases}
	$
	\vspace{8pt}
\end{tabular}
\caption{\small Here, $\pi(\mbx) = q(\mby =1 \g \mbx)$. Different encoding explanation methods that we consider.}
\label{tab:def-encoding}
\vspace{-15pt}
\end{table}

\begin{wrapfigure}[28]{r}{5.5cm}
\vspace{-13pt}
\begin{subfigure}[b]{0.39\textwidth}
\includegraphics[width=1\linewidth]{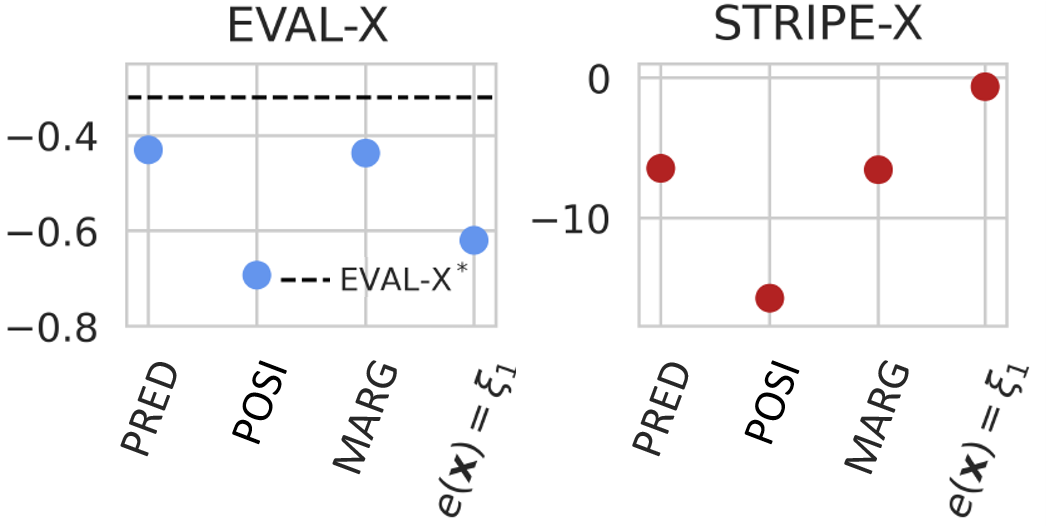}
\vspace{-14pt}
\caption{\small Results: discrete \gls{dgp}.}
\label{fig:discrete}
\end{subfigure}
\begin{subfigure}[b]{0.39\textwidth}
\vspace{4pt}
\includegraphics[width=1\linewidth]{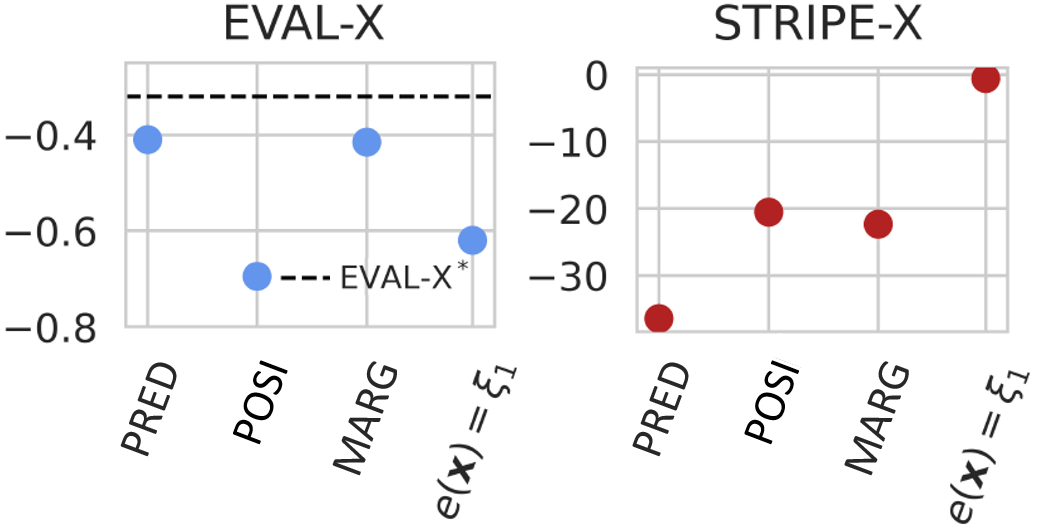}
\vspace{-14pt}
\caption{\small Results: hybrid \gls{dgp}. }
\label{fig:hybrid}
\end{subfigure}
\vspace{-15pt}
\caption{\small 
\acrshort{EVAL-X} and \detx{} scores of the $3$ encoding constructions and the non-encoding constant explanation $(e(\mbx) = \xi_1)$, for both \glspl{dgp}.
\evalx{}, being only a weak detector, assigns suboptimal scores to all encoding explanations ($<$), but scores some encoding explanations above the constant explanation.
On the other hand, \detx{}, being a strong detector, pushes down the scores of all the encoding explanations below that of the non-encoding constant explanation that always selects $\mbx_1$.
}
\vspace{-10pt}
\end{wrapfigure}
\setlength{\tabcolsep}{6pt}

\textbf{\textsc{roar} and \textsc{fresh} fails to weakly detect encoding.}
To empirically test the analysis about \roar{} and \textsc{fresh}, we study whether the two evaluations weakly detect encoding. 
In this study, we compare each evaluation's score on the all-inputs explanation, which is optimal, to the score assigned to \gls{margenc}.
\gls{margenc} ignores $\mbx_3$ which is required to produce the label $\mby$ in \cref{eq:experiment-simulation}.
\roar{} log-likelihoods for \gls{margenc} and the all-inputs explanation are approximately $-\mbH(\mby)=-0.69$ for both \glspl{dgp}.
In addition, the value of the input that \textsc{marg} selects alone
contains all the information about the label regardless of whether \textsc{marg} selects $\mbx_1$ or $\mbx_2$.
Thus, \fresh{} log-likelihoods for \textsc{marg} and the all-inputs explanation are both approximately $-0.29$ for both \textsc{dgp}s.
This result validates that \textsc{roar} and \textsc{fresh} are not weak detectors because they do not separate the optimal explanation from all encoding explanations.

\textbf{\evalx{} is a weak detector of encoding but not a strong detector.}
\acrshort{EVAL-X} log-likelihood scores are given in blue in \Cref{fig:discrete,fig:hybrid}. 
\evalx{}, being a weak detector, scores the encoding constructions (\gls{posenc}, \gls{predenc}, and \gls{margenc}) strictly lower than the log-likelihood of the optimal explanation $\acrshort{EVAL-X}^*$.
However, the \evalx{} score for the \gls{margenc} explanation is $-0.4$, which is above the score of $-0.6$ achieved by a non-encoding explanation $e(\mbx)=\xi_1$; 
\structpoint{thus, \evalx{} is not a strong detector.}

\textbf{Strong detector \detx{} prices out all the encoding explanations.}
\Cref{fig:discrete,fig:hybrid} report \detx{} scores for the same set of explanations as above; \detx{} scores are shown in red.
\structpoint{Strong detector \detx{} scores the non-encoding explanations above the negative entropy $-\mbH_q(\mby)=-0.69$ and scores every encoding construction under that threshold.}

\subsection{Detecting encoding on images of dogs and cats}

The goal of this section is to study the encoding detection capabilities of \textsc{roar}, \evalx{}, and \detx{} on real data.
We consider an image recognition task like the one in~\Cref{fig:encoding-diagram-marg} with labels and images from the \texttt{cats\_vs\_dogs} dataset from the Tensorflow package~\citep{tensorflow2015-whitepaper}.
We break images of size $64\times 64$ into $4$ patches each of size $32\times 32$.
In left-right then top-down order, let $\mbx_1, \mbx_2, \mbx_3, \mbx_4$ be 
the upper left, upper right, bottom left, and bottom right patches respectively; $\mbx_1, \mbx_3$ capture color, and $\mbx_2, \mbx_4$ are the animal images.
With $\texttt{annot(image)}$ denoting the annotation in the \texttt{cats\_vs\_dogs} dataset the \texttt{image} having a \texttt{dog} or a \texttt{cat}, the label is assigned as:
\begin{align*}
\mby & = \mathbbm{1}[\mbx_1=\texttt{blue}]\, \times\, \mathbbm{1}[\texttt{annot(image $\mbx_2$)=dog}]
    + \mathbbm{1}[\mbx_1=\texttt{red}]\,\times\,\mathbbm{1}[\texttt{annot(image $\mbx_4$)=dog}] 
\end{align*}
\begin{wraptable}[20]{r}{6.6cm}
\small
\centering
\vspace{-10pt}
\begin{tabular}{@{}rcccc@{}}
\toprule
      & 
        \textsc{roar}{} 
&
        \fresh{} 
&
        \evalx{} 
& 
        \detx{} 
\\ \midrule
\texttt{opt} 
    & $\boldsymbol{0.69}$
    & $\boldsymbol{-0.23}$
    &  $\boldsymbol{-0.27}$ 
    & $\boldsymbol{-0.31}$   \\
\texttt{fixed}
        & ${0.59}$
        & $-0.64$ 
        &  $-0.64$ 
        & ${-0.64}$   \\
\midrule
\gls{posenc}
                & $0.51$
                & $-0.69$ 
                &  $-0.70$ 
                & $-5.98$   \\
\gls{predenc}
                & $\boldsymbol{0.69}$
                & $\boldsymbol{-0.23}$
                &  ${-0.51}$ 
                & $-1.40$   \\
 \gls{margenc} 
                & $\boldsymbol{0.69}$
                & $\boldsymbol{-0.23}$
                &  ${-0.53}$
                & $-1.02$  
                \\  
                 \bottomrule
\end{tabular}
\caption{\small 
\roar{}, \fresh{}, \evalx{}, and \detx{} scores for the image recognition experiment.
Higher is better.
\roar{} and \fresh{} score two encoding explanations \gls{predenc} and \gls{margenc} as high as the \texttt{optimal} explanation, meaning they are not even weak detectors.
\evalx{} being only a weak detector scores \gls{posenc}, \gls{predenc}, and \gls{margenc} all worse than the \texttt{optimal} explanation under both \acrshort{EVAL-X} but not the non-encoding constant explanation ($e(\mbx)=\xi_4$), denoted \texttt{fixed}.
\acrshort{stripe-x} being a strong detector scores the non-encoding explanations above the negative marginal entropy $-\mbH_q(\mby)=-0.69$ and scores every encoding construction under that threshold.
}
\label{tab:real}
\end{wraptable}
We consider three encoding explanations (\gls{posenc}, \gls{predenc}, \gls{margenc}) and two non-encoding ones: 1) \texttt{optimal}, which selects the color and the patch that produces the label as dictated by the color, and 2) denoted \texttt{fixed}, which always outputs the bottom right patch $\mbx_4$.
\Cref{appsec:catsvdogs} gives details.

We report the scores assigned to each explanation by \roar{}, \evalx{}, and \detx{} in \Cref{tab:real}.
\roar{} scores two encoding explanations \gls{predenc} and \gls{margenc} as high as the \texttt{optimal} explanation, meaning it is not even a weak detector.
\gls{posenc}, \gls{predenc}, and \gls{margenc} all score worse than the \texttt{optimal} explanation under both the weak detector \acrshort{EVAL-X} and the strong detector \acrshort{stripe-x}.
However, \acrshort{EVAL-X} scores one non-encoding explanation (\texttt{fixed}) worse than two encoding ones, meaning it is not a strong detector.
Being a strong detector, \acrshort{stripe-x} scores the \texttt{fixed} explanation above the negative marginal entropy $-\mbH_q(\mby)=-0.69$ and scores every encoding construction under that threshold.

\textbf{Evaluating explanations produced by \realx{}~\citep{jethani2021have}.}
We ran \realx{} to learn explanations for the simulated setting and the image recognition task.
In the simulated setting, \realx{} is run to select one input;~\Cref{appsec:sim-enc-exp-details} gives details.
In the image recognition task, \realx{} is run to select one of the four patches as an explanation; \Cref{appsec:catsvdogs} gives details.
In both the simulated setting (see~\Cref{fig:discrete,fig:hybrid}) and the image recognition task (see~\Cref{tab:real}), \realx{} fails to achieve the optimal \evalx{} score while achieving a \detx{} score below the threshold of negative marginal entropy $-\mbH_q(\mby)=-0.69$.
Upon investigation, we found that \realx{} produced an explanation that matched the \gls{margenc} construction on at least $80\%$ of the inputs in the simulated setting.
On the image recognition task, \realx{} explanation matched the \gls{margenc} explanation on the whole dataset.
In both cases, \detx{}, being a strong detector, correctly alerts that the \realx{} explanation encodes.

\subsection{Encoding in \texorpdfstring{\acrshort{llm}}{}-generated explanations}

One can detect encoding in any explanation by checking if the \detx{} score falls below the negative marginal entropy.
Recent work uses \glspl{llm} to produce explanations; e.g.~\citep{li2022explanations} prompt an \gls{llm} to generate explanations for reasoning tasks which are later used to improve smaller models.
If the \gls{llm} explanation encodes, the smaller model can falsely ignore the informative inputs the larger model's explanation depends on and yet does not reveal.
In this section, we evaluate explanations generated by an \gls{llm}, \href{https://github.com/meta-llama/llama3}{Llama 3}, for a sentiment analysis task.
We consider reviews that take one of two forms: with \texttt{ADJ1} and \texttt{ADJ2} as adjectives, the review is
\vspace{-4pt}
\begin{itemize}[leftmargin=25pt,itemsep=1pt]
	\item  \texttt{\small 'My day was <ADJ1> and the movie was <ADJ2>. that is it'}
	   $\quad $ or
	\item  \texttt{\small 'My day was <ADJ1> and the movie was <ADJ2>. oh wait, reverse the adjectives'}.
\end{itemize}
The second sentence in the review acts as a "control flow" input and determines whether \texttt{ADJ1} or \texttt{ADJ2} describes the sentiment about the movie.
We prompt Llama 3 (see~\Cref{appsec:prompts}) to predict the sentiment and select a few words from the review that were important for that sentiment; the selected parts form the generated explanation.
To discourage encoding, the prompt explicitly instructs the \gls{llm} to select all the words that the \gls{llm} based the selection on; such an explanation, by~\Cref{lemma:defeq}, would be non-encoding.
On the $5$ most common selections $e(\mbx)$ generated by the \gls{llm}, we compute the \evalx{} score and the \encmeas{} $\phi_q(e)$.
The resulting \detx{} score is $-2.78$, falling short of the negative entropy $-\mbH_q(\mby)=-0.69$, meaning the~\gls{llm} encodes.
We investigated why.

As an example, consider the review \texttt{\small 'My day was resplendent and the movie was hollow. That is it.'}; the \gls{llm} selects only \texttt{\small hollow} in the explanation.
However, the \gls{llm} instead selects \texttt{\small resplendent} when \texttt{\small That is it} is switched to \texttt{\small oh wait, reverse the adjectives.}
Such occurrences are common.
On $>70\%$ of the data, the \gls{llm} selects the word that describes the movie but does not select the second sentence in the review which controls which adjective describes the movie; this is akin to \gls{margenc} encoding.
Thus, the \gls{llm}-generated explanation encodes by looking at the control flow input in the second sentence to find the correct adjectives, but failing to select the control flow input.
Such an explanation falsely indicates that only the adjectives are relevant to predicting the label.
In contrast, a non-encoding explanation would, in addition to the adjective that describes the movie, reveal control flow words that indicate which adjective predicts the label.

\looseness=-1
In summary, despite being instructed to include all the words that were looked at when producing the explanation, the \gls{llm} encodes.
Building non-encoding explanations with \glspl{llm} may require an extensive search over prompts or finetuning guided by scores from \detx{}.

\vspace{-3pt}
\section{Discussion}
\vspace{-3pt}
When an explanation is encodes, predictions from the explanation become disconnected from predictions from the values in the explanations.
Such explanations can select values with little relevance to the label and yet score highly on the many existing predictive evaluations.
We develop a simple statistical definition of encoding.
Inverting this definition shows that when non-encoding explanations predict the label, users know the values of those inputs selected in the explanation predict the label.
We then show that existing evaluations are either non-detectors (\textsc{roar}\citep{hooker2019benchmark},\textsc{fresh}~\citep{jain2020learning}) or only weak detectors (\evalx{}~\citep{jethani2021have}).
Motivated by this, we introduce a new strong detector, \detx{}.
After empirically demonstrating the detection capabilities (or lack thereof) of said evaluations, we use \detx{} to discover encoding in \gls{llm}-generated explanations.

\textbf{More related work.}
Other investigations into evaluating explanations focused on label leakage \citep{jethani2023don,hase2020leakage} and faithfulness \citep{bastings2021will,zhou2022feature,ju2021logic,jain2020learning,jacovi2020towards}.
Label leakage is similar to encoding in that additional information is in the explanation, but focuses on explanations that have access to both the inputs and the observed label; we leave extending~\encdef{} to leakage to the future.
Faithfulness, intuitively, asks that the explanation reflect the process of how a label is predicted from the inputs; a formalization does not exist.
\citet{jacovi2020towards} note the need to define faithfulness formally.
Encoding explanations are not faithful to the process of making an explanation because predictive inputs outside those selected by the explanation control the explanation.

\textbf{Limitations and the future.}
Using misestimated models in evaluations (like \acrshort{EVAL-X}) may lead to mistakes (see~\Cref{appsec:estimation-error-example} for an example). 
The retinal fundus experiment from \citet{jethani2023don} is an example where misestimation leads to reductive explanations scoring higher than using the full input.
Misestimation can be due to poor uncertainty or due to dependence on shortcut features.
One fruitful direction is to use better uncertainty estimates, like conformal inference~\citep{lei2018distribution} or calibration~\citep{goldstein2020x}, or employ robustness methods~\citep{puli2021out,puli2023don} to ameliorate errors due to misestimation.
Another direction is use tricks like \textsc{reinforce}-style gradients to construct non-encoding explanations by optimizing \detx{}.
Explanations that output subsets may not always help humans interpret the mechanism of the prediction. 
For example, imagine one wants to understand why a model correctly answers the question "Who won the ski halfpipe at the X-games 3 years after her debut in 2021?" with "Eileen Gu".
A subset explanation may return "3 years after her debut in 2021" and "ski halfpipe", but that does not help a human interpret how the model predicts.
A better interpretation would be to make the model output, "3 years after 2021 is 2024. Eileen Gu won in 2024, and debuted in 2021."
Such explanations can also encode information about the prediction in the text produced as a rationale~\citep{atanasova2023faithfulness}.
An important direction here would be to extend the definitions of weak and strong detectors of encoding to evaluations of free-text rationales.

\textbf{Data versus Model Explanations.}
Even with the formal definitions of explanation methods, there is a question about what is being explained: the data or the model.
These two concepts often get blended together in the literature \citep{yoon2018invase,jethani2021have}. 
We clarify this point and abstract the choice away as two different ways to produce the joint distribution $q(\mby, \mbx)$.
In \textit{data explanation}, the distribution under which a feature attribution method seeks to output a subset of inputs that predict the label should be the population distribution of the data \citep{chen2018learning}.
If, instead, the goal is \textit{model explanation}, the goal should not be to highlight inputs that predict the label well in samples of the data; rather it should be \emph{to predict the label well in samples from the model.} 
Formally, a model with parameters $\mbtheta$ is a conditional distribution, $p_\mbtheta(\mby \g \mbx)$. 
To target a model explanation, a feature attribution method would aim to output a subset of inputs that predict the label under the distribution $F(\mbx) p_\mbtheta(\mby \g \mbx)$.

\newpage 
\section*{Acknowledgements}
This work was partly supported by the NIH/NHLBI Award R01HL148248, NSF Award 1922658 NRT-HDR:FUTURE Foundations, Translation, and Responsibility for Data Science, NSF CAREER Award 2145542, NSF Award 2404476, ONR N00014-23-1-2634, Google DeepMind, and Apple.
The authors would like to thank Yoav Wald, the NeurIPS 2024 reviewers and the NeurIPS 2024 area chair for helpful feedback.

\bibliographystyle{unsrtnat}
\bibliography{ref}

\begin{thebibliography}{43}
\providecommand{\natexlab}[1]{#1}
\providecommand{\url}[1]{\texttt{#1}}
\expandafter\ifx\csname urlstyle\endcsname\relax
  \providecommand{\doi}[1]{doi: #1}\else
  \providecommand{\doi}{doi: \begingroup \urlstyle{rm}\Url}\fi

\bibitem[Elias et~al.(2022)Elias, Poterucha, Rajaram, Moller, Rodriguez, Bhave, Hahn, Tison, Abreau, Barrios, et~al.]{elias2022deep}
Pierre Elias, Timothy~J Poterucha, Vijay Rajaram, Luca~Matos Moller, Victor Rodriguez, Shreyas Bhave, Rebecca~T Hahn, Geoffrey Tison, Sean~A Abreau, Joshua Barrios, et~al.
\newblock Deep learning electrocardiographic analysis for detection of left-sided valvular heart disease.
\newblock \emph{Journal of the American College of Cardiology}, 80\penalty0 (6):\penalty0 613--626, 2022.

\bibitem[Jethani et~al.(2022{\natexlab{a}})Jethani, Puli, Zhang, Garber, Jankelson, Aphinyanaphongs, and Ranganath]{jethani2022new}
Neil Jethani, Aahlad Puli, Hao Zhang, Leonid Garber, Lior Jankelson, Yindalon Aphinyanaphongs, and Rajesh Ranganath.
\newblock New-onset diabetes assessment using artificial intelligence-enhanced electrocardiography.
\newblock \emph{arXiv preprint arXiv:2205.02900}, 2022{\natexlab{a}}.

\bibitem[Simonyan et~al.(2014)Simonyan, Vedaldi, and Zisserman]{simonyan2014visualising}
Karen Simonyan, Andrea Vedaldi, and Andrew Zisserman.
\newblock Visualising image classification models and saliency maps.
\newblock \emph{Deep Inside Convolutional Networks}, 2, 2014.

\bibitem[Lundberg and Lee(2017)]{lundberg2017unified}
Scott~M Lundberg and Su-In Lee.
\newblock A unified approach to interpreting model predictions.
\newblock \emph{Advances in neural information processing systems}, 30, 2017.

\bibitem[Tran et~al.(2022)Tran, Le, Nguyen, and Nguyen]{tran2022explainable}
Kim~Long Tran, Hoang~Anh Le, Thanh~Hien Nguyen, and Duc~Trung Nguyen.
\newblock Explainable machine learning for financial distress prediction: evidence from vietnam.
\newblock \emph{Data}, 7\penalty0 (11):\penalty0 160, 2022.

\bibitem[DeGrave et~al.(2021)DeGrave, Janizek, and Lee]{degrave2021ai}
Alex~J DeGrave, Joseph~D Janizek, and Su-In Lee.
\newblock Ai for radiographic covid-19 detection selects shortcuts over signal.
\newblock \emph{Nature Machine Intelligence}, 3\penalty0 (7):\penalty0 610--619, 2021.

\bibitem[Wong et~al.(2023)Wong, Zheng, Valeri, Donghia, Anahtar, Omori, Li, Cubillos-Ruiz, Krishnan, Jin, et~al.]{wong2023discovery}
Felix Wong, Erica~J Zheng, Jacqueline~A Valeri, Nina~M Donghia, Melis~N Anahtar, Satotaka Omori, Alicia Li, Andres Cubillos-Ruiz, Aarti Krishnan, Wengong Jin, et~al.
\newblock Discovery of a structural class of antibiotics with explainable deep learning.
\newblock \emph{Nature}, pages 1--9, 2023.

\bibitem[Selvaraju et~al.(2017)Selvaraju, Cogswell, Das, Vedantam, Parikh, and Batra]{selvaraju2017grad}
Ramprasaath~R Selvaraju, Michael Cogswell, Abhishek Das, Ramakrishna Vedantam, Devi Parikh, and Dhruv Batra.
\newblock Grad-cam: Visual explanations from deep networks via gradient-based localization.
\newblock In \emph{Proceedings of the IEEE international conference on computer vision}, pages 618--626, 2017.

\bibitem[Covert et~al.(2021)Covert, Lundberg, and Lee]{covert2021explaining}
Ian Covert, Scott Lundberg, and Su-In Lee.
\newblock Explaining by removing: A unified framework for model explanation.
\newblock \emph{Journal of Machine Learning Research}, 22\penalty0 (209):\penalty0 1--90, 2021.

\bibitem[Jethani et~al.(2022{\natexlab{b}})Jethani, Sudarshan, Covert, Lee, and Ranganath]{jethani2022fastshap}
Neil Jethani, Mukund Sudarshan, Ian~Connick Covert, Su-In Lee, and Rajesh Ranganath.
\newblock Fastshap: Real-time shapley value estimation.
\newblock In \emph{International Conference on Learning Representations}, 2022{\natexlab{b}}.

\bibitem[Yoon et~al.(2018)Yoon, Jordon, and van~der Schaar]{yoon2018invase}
Jinsung Yoon, James Jordon, and Mihaela van~der Schaar.
\newblock Invase: Instance-wise variable selection using neural networks.
\newblock In \emph{International Conference on Learning Representations}, 2018.

\bibitem[Lage et~al.(2019)Lage, Chen, He, Narayanan, Kim, Gershman, and Doshi-Velez]{lage2019human}
Isaac Lage, Emily Chen, Jeffrey He, Menaka Narayanan, Been Kim, Samuel~J Gershman, and Finale Doshi-Velez.
\newblock Human evaluation of models built for interpretability.
\newblock In \emph{Proceedings of the AAAI Conference on Human Computation and Crowdsourcing}, volume~7, pages 59--67, 2019.

\bibitem[Saporta et~al.(2022)Saporta, Gui, Agrawal, Pareek, Truong, Nguyen, Ngo, Seekins, Blankenberg, Ng, et~al.]{saporta2022benchmarking}
Adriel Saporta, Xiaotong Gui, Ashwin Agrawal, Anuj Pareek, Steven~QH Truong, Chanh~DT Nguyen, Van-Doan Ngo, Jayne Seekins, Francis~G Blankenberg, Andrew~Y Ng, et~al.
\newblock Benchmarking saliency methods for chest x-ray interpretation.
\newblock \emph{Nature Machine Intelligence}, 4\penalty0 (10):\penalty0 867--878, 2022.

\bibitem[Crabb{\'e} et~al.(2022)Crabb{\'e}, Curth, Bica, and van~der Schaar]{crabbe2022benchmarking}
Jonathan Crabb{\'e}, Alicia Curth, Ioana Bica, and Mihaela van~der Schaar.
\newblock Benchmarking heterogeneous treatment effect models through the lens of interpretability.
\newblock \emph{Advances in Neural Information Processing Systems}, 35:\penalty0 12295--12309, 2022.

\bibitem[Samek et~al.(2016)Samek, Binder, Montavon, Lapuschkin, and M{\"u}ller]{samek2016evaluating}
Wojciech Samek, Alexander Binder, Gr{\'e}goire Montavon, Sebastian Lapuschkin, and Klaus-Robert M{\"u}ller.
\newblock Evaluating the visualization of what a deep neural network has learned.
\newblock \emph{IEEE transactions on neural networks and learning systems}, 28\penalty0 (11):\penalty0 2660--2673, 2016.

\bibitem[Petsiuk et~al.(2018)Petsiuk, Das, and Saenko]{petsiuk1806rise}
V~Petsiuk, A~Das, and K~Saenko.
\newblock Rise: Randomized input sampling for explanation of black-box models.
\newblock \emph{arXiv preprint arXiv:1806.07421}, 2018.

\bibitem[Dabkowski and Gal(2017)]{dabkowski2017real}
Piotr Dabkowski and Yarin Gal.
\newblock Real time image saliency for black box classifiers.
\newblock \emph{Advances in neural information processing systems}, 30, 2017.

\bibitem[Jain et~al.(2020)Jain, Wiegreffe, Pinter, and Wallace]{jain2020learning}
Sarthak Jain, Sarah Wiegreffe, Yuval Pinter, and Byron~C Wallace.
\newblock Learning to faithfully rationalize by construction.
\newblock \emph{arXiv preprint arXiv:2005.00115}, 2020.

\bibitem[Hooker et~al.(2019)Hooker, Erhan, Kindermans, and Kim]{hooker2019benchmark}
Sara Hooker, Dumitru Erhan, Pieter-Jan Kindermans, and Been Kim.
\newblock A benchmark for interpretability methods in deep neural networks.
\newblock \emph{Advances in neural information processing systems}, 32, 2019.

\bibitem[Jethani et~al.(2021)Jethani, Sudarshan, Aphinyanaphongs, and Ranganath]{jethani2021have}
Neil Jethani, Mukund Sudarshan, Yindalon Aphinyanaphongs, and Rajesh Ranganath.
\newblock Have we learned to explain?: How interpretability methods can learn to encode predictions in their interpretations.
\newblock In \emph{International Conference on Artificial Intelligence and Statistics}, pages 1459--1467. PMLR, 2021.

\bibitem[Hsia et~al.(2023)Hsia, Pruthi, Singh, and Lipton]{hsia2023goodhart}
Jennifer Hsia, Danish Pruthi, Aarti Singh, and Zachary~C Lipton.
\newblock Goodhart's law applies to nlp's explanation benchmarks.
\newblock \emph{arXiv preprint arXiv:2308.14272}, 2023.

\bibitem[Guyon and Elisseeff(2003)]{guyon2003introduction}
Isabelle Guyon and Andr{\'e} Elisseeff.
\newblock An introduction to variable and feature selection.
\newblock \emph{Journal of machine learning research}, 3\penalty0 (Mar):\penalty0 1157--1182, 2003.

\bibitem[Chen et~al.(2018)Chen, Song, Wainwright, and Jordan]{chen2018learning}
Jianbo Chen, Le~Song, Martin Wainwright, and Michael Jordan.
\newblock Learning to explain: An information-theoretic perspective on model interpretation.
\newblock In \emph{International conference on machine learning}, pages 883--892. PMLR, 2018.

\bibitem[Ribeiro et~al.(2016)Ribeiro, Singh, and Guestrin]{ribeiro2016should}
Marco~Tulio Ribeiro, Sameer Singh, and Carlos Guestrin.
\newblock " why should i trust you?" explaining the predictions of any classifier.
\newblock In \emph{Proceedings of the 22nd ACM SIGKDD international conference on knowledge discovery and data mining}, pages 1135--1144, 2016.

\bibitem[{\v{S}}trumbelj and Kononenko(2014)]{vstrumbelj2014explaining}
Erik {\v{S}}trumbelj and Igor Kononenko.
\newblock Explaining prediction models and individual predictions with feature contributions.
\newblock \emph{Knowledge and information systems}, 41:\penalty0 647--665, 2014.

\bibitem[Jethani et~al.(2023)Jethani, Saporta, and Ranganath]{jethani2023don}
Neil Jethani, Adriel Saporta, and Rajesh Ranganath.
\newblock Don’t be fooled: label leakage in explanation methods and the importance of their quantitative evaluation.
\newblock In \emph{International Conference on Artificial Intelligence and Statistics}, pages 8925--8953. PMLR, 2023.

\bibitem[Zhang et~al.(2011)Zhang, Peters, Janzing, and Sch{\"o}lkopf]{zhang2011kernel}
Kun Zhang, Jonas Peters, Dominik Janzing, and Bernhard Sch{\"o}lkopf.
\newblock Kernel-based conditional independence test and application in causal discovery.
\newblock In \emph{Proceedings of the Twenty-Seventh Conference on Uncertainty in Artificial Intelligence}, pages 804--813, 2011.

\bibitem[Sudarshan et~al.(2020)Sudarshan, Tansey, and Ranganath]{sudarshan2020deep}
Mukund Sudarshan, Wesley Tansey, and Rajesh Ranganath.
\newblock Deep direct likelihood knockoffs.
\newblock \emph{Advances in neural information processing systems}, 33:\penalty0 5036--5046, 2020.

\bibitem[Sudarshan et~al.(2023)Sudarshan, Puli, Tansey, and Ranganath]{sudarshan2023diet}
Mukund Sudarshan, Aahlad Puli, Wesley Tansey, and Rajesh Ranganath.
\newblock Diet: Conditional independence testing with marginal dependence measures of residual information.
\newblock In \emph{International Conference on Artificial Intelligence and Statistics}, pages 10343--10367. PMLR, 2023.

\bibitem[Brown et~al.(2020)Brown, Mann, Ryder, Subbiah, Kaplan, Dhariwal, Neelakantan, Shyam, Sastry, Askell, et~al.]{brown2020language}
Tom Brown, Benjamin Mann, Nick Ryder, Melanie Subbiah, Jared~D Kaplan, Prafulla Dhariwal, Arvind Neelakantan, Pranav Shyam, Girish Sastry, Amanda Askell, et~al.
\newblock Language models are few-shot learners.
\newblock \emph{Advances in neural information processing systems}, 33:\penalty0 1877--1901, 2020.

\bibitem[Ramesh et~al.(2022)Ramesh, Dhariwal, Nichol, Chu, and Chen]{ramesh2022hierarchical}
Aditya Ramesh, Prafulla Dhariwal, Alex Nichol, Casey Chu, and Mark Chen.
\newblock Hierarchical text-conditional image generation with clip latents, 2022.
\newblock \emph{URL https://arxiv. org/abs/2204.06125}, 7, 2022.

\bibitem[Abadi et~al.(2015)Abadi, Agarwal, Barham, Brevdo, Chen, Citro, Corrado, Davis, Dean, Devin, Ghemawat, Goodfellow, Harp, Irving, Isard, Jia, Jozefowicz, Kaiser, Kudlur, Levenberg, Man\'{e}, Monga, Moore, Murray, Olah, Schuster, Shlens, Steiner, Sutskever, Talwar, Tucker, Vanhoucke, Vasudevan, Vi\'{e}gas, Vinyals, Warden, Wattenberg, Wicke, Yu, and Zheng]{tensorflow2015-whitepaper}
Mart\'{\i}n Abadi, Ashish Agarwal, Paul Barham, Eugene Brevdo, Zhifeng Chen, Craig Citro, Greg~S. Corrado, Andy Davis, Jeffrey Dean, Matthieu Devin, Sanjay Ghemawat, Ian Goodfellow, Andrew Harp, Geoffrey Irving, Michael Isard, Yangqing Jia, Rafal Jozefowicz, Lukasz Kaiser, Manjunath Kudlur, Josh Levenberg, Dan Man\'{e}, Rajat Monga, Sherry Moore, Derek Murray, Chris Olah, Mike Schuster, Jonathon Shlens, Benoit Steiner, Ilya Sutskever, Kunal Talwar, Paul Tucker, Vincent Vanhoucke, Vijay Vasudevan, Fernanda Vi\'{e}gas, Oriol Vinyals, Pete Warden, Martin Wattenberg, Martin Wicke, Yuan Yu, and Xiaoqiang Zheng.
\newblock {TensorFlow}: Large-scale machine learning on heterogeneous systems, 2015.
\newblock URL \url{http://tensorflow.org/}.
\newblock Software available from tensorflow.org.

\bibitem[Li et~al.(2022)Li, Chen, Shen, Chen, Zhang, Li, Wang, Qian, Peng, Mao, et~al.]{li2022explanations}
Shiyang Li, Jianshu Chen, Yelong Shen, Zhiyu Chen, Xinlu Zhang, Zekun Li, Hong Wang, Jing Qian, Baolin Peng, Yi~Mao, et~al.
\newblock Explanations from large language models make small reasoners better.
\newblock \emph{arXiv preprint arXiv:2210.06726}, 2022.

\bibitem[Hase et~al.(2020)Hase, Zhang, Xie, and Bansal]{hase2020leakage}
Peter Hase, Shiyue Zhang, Harry Xie, and Mohit Bansal.
\newblock Leakage-adjusted simulatability: Can models generate non-trivial explanations of their behavior in natural language?
\newblock \emph{arXiv preprint arXiv:2010.04119}, 2020.

\bibitem[Bastings et~al.(2021)Bastings, Ebert, Zablotskaia, Sandholm, and Filippova]{bastings2021will}
Jasmijn Bastings, Sebastian Ebert, Polina Zablotskaia, Anders Sandholm, and Katja Filippova.
\newblock " will you find these shortcuts?" a protocol for evaluating the faithfulness of input salience methods for text classification.
\newblock \emph{arXiv preprint arXiv:2111.07367}, 2021.

\bibitem[Zhou et~al.(2022)Zhou, Booth, Ribeiro, and Shah]{zhou2022feature}
Yilun Zhou, Serena Booth, Marco~Tulio Ribeiro, and Julie Shah.
\newblock Do feature attribution methods correctly attribute features?
\newblock In \emph{Proceedings of the AAAI Conference on Artificial Intelligence}, volume~36, pages 9623--9633, 2022.

\bibitem[Ju et~al.(2021)Ju, Zhang, Yang, Jiang, Liu, and Zhao]{ju2021logic}
Yiming Ju, Yuanzhe Zhang, Zhao Yang, Zhongtao Jiang, Kang Liu, and Jun Zhao.
\newblock Logic traps in evaluating attribution scores.
\newblock \emph{arXiv preprint arXiv:2109.05463}, 2021.

\bibitem[Jacovi and Goldberg(2020)]{jacovi2020towards}
Alon Jacovi and Yoav Goldberg.
\newblock Towards faithfully interpretable nlp systems: How should we define and evaluate faithfulness?
\newblock \emph{arXiv preprint arXiv:2004.03685}, 2020.

\bibitem[Lei et~al.(2018)Lei, G’Sell, Rinaldo, Tibshirani, and Wasserman]{lei2018distribution}
Jing Lei, Max G’Sell, Alessandro Rinaldo, Ryan~J Tibshirani, and Larry Wasserman.
\newblock Distribution-free predictive inference for regression.
\newblock \emph{Journal of the American Statistical Association}, 113\penalty0 (523):\penalty0 1094--1111, 2018.

\bibitem[Goldstein et~al.(2020)Goldstein, Han, Puli, Perotte, and Ranganath]{goldstein2020x}
Mark Goldstein, Xintian Han, Aahlad Puli, Adler Perotte, and Rajesh Ranganath.
\newblock X-cal: Explicit calibration for survival analysis.
\newblock \emph{Advances in neural information processing systems}, 33:\penalty0 18296--18307, 2020.

\bibitem[Puli et~al.(2021)Puli, Zhang, Oermann, and Ranganath]{puli2021out}
Aahlad~Manas Puli, Lily~H Zhang, Eric~Karl Oermann, and Rajesh Ranganath.
\newblock Out-of-distribution generalization in the presence of nuisance-induced spurious correlations.
\newblock \emph{ICLR 2022}, 2021.

\bibitem[Puli et~al.(2023)Puli, Zhang, Wald, and Ranganath]{puli2023don}
Aahlad~Manas Puli, Lily Zhang, Yoav Wald, and Rajesh Ranganath.
\newblock Don’t blame dataset shift! shortcut learning due to gradients and cross entropy.
\newblock \emph{Advances in Neural Information Processing Systems}, 36, 2023.

\bibitem[Atanasova et~al.(2023)Atanasova, Camburu, Lioma, Lukasiewicz, Simonsen, and Augenstein]{atanasova2023faithfulness}
Pepa Atanasova, Oana-Maria Camburu, Christina Lioma, Thomas Lukasiewicz, Jakob~Grue Simonsen, and Isabelle Augenstein.
\newblock Faithfulness tests for natural language explanations.
\newblock \emph{arXiv preprint arXiv:2305.18029}, 2023.

\end{thebibliography}

\newpage

\appendix

\section{Theoretical Details}

\subsection{Expressing \texorpdfstring{$q(\mby \g \xoex)$}{}  in terms of the values and the identity of the explanation}\label{appsec:simplification}

To express $q(\mby \g \xoex)$, we use the following equivalence of events from \cref{eq:event-eq}
\begin{align}\tag{\cref{eq:event-eq}}
    \{\mbx : \xex = (\mbv, \mba) \} = \{\mbx : e(\mbx) = \mbv\} \cap \{\mbx : \xv = \mba\}.
\end{align}
Then, intuitively, conditioning on the event that $\xex=(\mbv,\mba)$ gives you the same information as conditioning on the events $e(\mbx) = \mbv$ and $\xv = \mba$ simultaneously.
We make this formal below. 

For discrete $\mbx$, for any $\mbv, \mba$ such that the probability $q(\xex = (\mbv, \mba)) > 0$, define the LHS and RHS of \cref{eq:event-eq} as $B_\mbv, C_\mbv$ respectively.
Then, the conditionals $q(\mby \g \xex =(\mbv, \mba))$ and $q(\mby\g  \mbE_\mbv=1, \xv = \mba)$ exist and can be written as follows:
\[q(\mby\g \xex = (\mbv,\mba)) = q(\mby\g \mbx\in B_\mbv) \qquad q(\mby\g \mbE_\mbv=1, \xv = \mba) = q(\mby\g \mbx\in C_\mbv).\]
These two conditionals are equal because $B_\mbv=C_\mbv$.

The same kind of result holds for general random vectors (discrete or continuous) $\mbx$ but is a little more involved because $B_\mbv$ may be non-empty while $q(\mbx\in B_\mbv)=0$ and the equality of conditional densities/probabilities need to be written via measure theory. 
Assume the regular conditional probabilities $q(\mby\g \xex)$ and $q(\mby, \mbE_\mbv\g \mbx_\mbv), q(\mby\g \mbE_\mbv, \mbx_\mbv)$ and $q(\mbE_\mbv \g \mbx_\mbv)$ are defined almost everywhere in their respective probability measures.
Take any  $\mbS_\mbv \subseteq \{\xv : e(\mbx)=\mbv\}$ where $q(\xv\in \mbS_\mbv)>0$ and $q(e(\mbx) = \mbv)>0$.
Consider any measurable sets $\mbY$ over $\mby$ and $\mbB_\mbv(\mbS_\mbv):= \{(\mbv, \mba) : \mba \in \mbS_\mbv\}$ over $\xex$. Now, by definition of regular conditional probability measures, joint probabilities are obtained by taking the expectation of the conditional with respect to marginal distributions over the conditioning set:
\begin{align}
    q(\mby \in \mbY, \xex \in \mbB_\mbv(\mbS_\mbv)) &  = \int_{\mbB_\mbv(\mbS_\mbv)} q(\mby \in \mbY \g \xex = (\mbv, \mba)) q(d\xex) \nonumber
    \\
    &  = \int_{\mbS_\mbv} q(\mby \in \mbY \g \xex = (\mbv, \mba)) q(\xex = (\mbv, \mba)) d \mba
    \label{eq:xex-version}
\end{align}
and 
\begin{align}
    q(\mby \in \mbY, \mbE_\mbv=1, \xv \in \mbS_\mbv) &  =  \int_{\mbS_\mbv} q(\mby \in \mbY, \mbE_\mbv=1 \g \xv=\mba) q(\xv=\mba) d \mba \nonumber 
\\
    &  =  \int_{\mbS_\mbv} q(\mby \in \mbY \g \mbE_\mbv=1, \xv=\mba) q(\mbE_\mbv=1, \xv=\mba) d \mba \label{eq:ev-version}.
\end{align}
Due to \cref{eq:event-eq}, the LHS terms of the two equations above are equal and so are the probability measures over the integrating variables in~\cref{eq:xex-version,eq:ev-version}.
Letting $\xv$ be defined on a Borel sigma algebra, these two integrals~\cref{eq:xex-version,eq:ev-version} are equal if and only if for any Borel set $\mbS_\mbv$, for almost every $\mba\in \mbS_\mbv$
\begin{align*}
   q(\mby \in \mbY \g \xex = (\mbv, \mba) ) = q(\mby \in \mbY \g \mbE_\mbv=1, \xv=\mba).
\end{align*}
That is, in more plain terms, the conditional distributions are equal $q(\mby \g \xex = (\mbv, \mba)) = q(\mby \g \mbE_\mbv=1, \xv = \mba)$.

\subsection{With non-encoding explanations "what you see is what you get"}\label{appsec:wysiwyg}

\encdef{} says that an explanation $e(\mbx)$ is encoding if there exists an $\mbS$ where $q(\xex \in \mbS) > 0$ such that for every $(\mbv, \mba) \in \mbS$, :
        \begin{align}\label{eq:app-a11-dependence}
            \mby \,\, \nindep \,\, \mbE_\mbv \g \xv = \mba.
        \end{align}
For a non-encoding explanation,~\encdef{} does not hold. Here, we derive the implications of violating~\encdef{}.
Define the set $\mbA$ to contain all $(\mbv, \mba)$ where \cref{eq:app-a11-dependence} is violated:
\begin{align}\label{eq:indep-in-set-A}
	\mbA = \{(\mbv, \mba): \mby \,\, \indep \,\, \mbE_\mbv \g \xv = \mba\}.
\end{align}
By definition, the complement $\mbA^C$ is such that
\[\forall (\mbv, \mba) \in \mbA^C, \quad   \mby \,\, \nindep \,\, \mbE_\mbv \g \xv = \mba.\]
Such a set cannot have positive measure when~\encdef{} is violated which means
\[q(\xex \in \mbA^C) = 0.\]
In turn,
\[q(\xex\in \mbA) = 1 - q(\xex\in \mbA^C) = 1.\]

Thus, $\mbA$ is such that $q(\xex \in \mbA) = 1$, and by \cref{eq:indep-in-set-A} for all $(\mbv,\mba) \in \mbA$
\[\mby \,\, \indep \,\, \mbE_\mbv \g \xv = \mba,\]
which in turn guarantees
 \[q(\mby \g \xex=(\mbv,\mba)) = q(\mby\g \xv =\mba, \mbE_\mbv=1) = q(\mby\g \xv=\mba).
 \]

\subsection{Helpful Lemmas and their proofs}
\label{sec:proof-lemma}

\setcounter{lemma}{0}
\subsubsection{Alternate conditions equivalent to \texorpdfstring{\encdef{}}{}}\label{appsec:defeq}
The dependence in \encdef{} occurs due to two reasons, understanding which sheds more light on the definition.
First, for some selection $e(\mbx) = \mbv$, the explanation's values $\mbx_{\mbv}$ do not provide enough information to reveal that the explanation should select the inputs denoted by $\mbv$. 
In other words, the indicator of the selection is variable even after fixing the explanation's values themselves.
Second, this indicator is predictive of the label for the data with the explanation $\mbv$.
These two properties provide intuition on the definition of encoding:
\newcommand{\defeq}{
 \encdef{} holds for an explanation $e(\mbx)$ if and only if there exists a selection $\mbv$ such that $q(e(\mbx) = \mbv) > 0$ and a set $\mbS_\mbv \subseteq \{\mbx_\mbv: e(\mbx) = \mbv\}$ such that $q(\mbx_\mbv \in \mbS_\mbv) > 0$ where both of the following conditions hold for almost every $\mba \in \mbS_\mbv$:
        \begin{align*}
\textbf{Unpredictability of Explanation} & \qquad
             q(\mbE_\mbv = 1 \g \mbx_\mbv = \mba) \neq 1 ;
\\
\textbf{Additional Information from Explanation} & \qquad 
            q(\mby \g \mbx_\mbv= \mba, \mbE_\mbv = 1) \neq q(\mby \g \mbx_\mbv = \mba, \mbE_\mbv = 0). 
        \end{align*}
}
\begin{lemma}
\label{lemma:defeq}
    \defeq{}
\end{lemma}

\begin{proof}

First,~\Cref{lemma:connection-lemma} shows the~\encdef{} holds if only if there exists a selection $\mbv$ such that $q(e(\mbx) = \mbv)>0$ and a set $\mbS_\mbv \subseteq \{\mbx_\mbv: e(\mbx) = \mbv\}$ such that $q(\mbx_\mbv \in \mbS_\mbv) > 0$ where
\[\forall \mba \in \mbS_\mbv, \quad   \mby \,\, \nindep \,\, \mbE_\mbv \g \mbx_\mbv.\]
We use this alternate definition in what follows.

Given a non-measure zero set $\mbS_\mbv$, by \Cref{lemma:positiveind}, almost everywhere in $\mbS_\mbv$ it holds that $q(\mbE_\mbv = 1 \g \mbx_\mbv) > 0$.

\paragraph{Conditional dependence implies Unpredictability and Additional information (the only if part).}
If $q(\mbE_\mbv =1 \g \mbx_\mbv) = 1$ almost everywhere (under $q(\mbx)$), then $\mbE_\mbv$ is constant  given $\mbx_\mbv$, and therefore independent of any variable given $\mbx_\mbv$:
\[ q(\mbE_\mbv = 1 \g \mbx_\mbv) = 1 \implies \mby\indep \mbE_\mbv \g \mbx_\mbv. \]
Then, it follows that conditional dependence implies the unpredictability property
\[\mby\nindep \mbE_\mbv \g \mbx_\mbv \implies q(\mbE_\mbv = 1 \g \mbx_\mbv) <  1.\]

Second, with the result from \Cref{lemma:positiveind}, we have $q(\mbE_\mbv = 1 | \mbx_\mbv) \in (0, 1)$. Thus $q(\mby | \mbx_\mbv, \mbE_\mbv = 1)$ and $q(\mby | \mbx_\mbv, \mbE_\mbv = 0)$ exist almost every where in $\mbS_\mbv$.
Then, by definition of conditional dependence, there is additional information about the label in the explanation:
\[\mby\nindep \mbE_\mbv \g \mbx_\mbv \implies q(\mby \g \mbx_\mbv, \mbE_\mbv=1) \not= q(\mby\g \mbx_\mbv, \mbE_\mbv=0).\]
This shows that \encdef{} implies the additional information property.

\paragraph{Conditional dependence implied by Unpredictability and Additional information (the if part).}
Now, if $q(\mbE_\mbv = 1 \g \mbx_\mbv) \in (0, 1)$, then the following two conditional distributions exist almost everywhere in $\mbS_\mbv$
\[q(\mby \g \mbx_\mbv, \mbE_\mbv=1) \qquad , \qquad q(\mby\g \mbx_\mbv, \mbE_\mbv=0).\]
Then, by definition of dependence almost everywhere $\mbS_\mbv$:
\[q(\mby \g \mbx_\mbv, \mbE_\mbv=1) \not= q(\mby\g \mbx_\mbv, \mbE_\mbv=0) \Longrightarrow \mby\nindep \mbE_\mbv \g \mbx_\mbv.\]
Thus, the unpredictability and the additional information properties imply \encdef{}.

\end{proof}

\begin{lemma}\label{lemma:connection-lemma}
\encdef{} holds for an explanation $e(\mbx)$ if and only if there exists a selection $\mbv$ such that $q(e(\mbx) = \mbv)>0$ and a set $\mbS_\mbv \subseteq \{\mbx_\mbv: e(\mbx) = \mbv\}$ such that $q(\mbx_\mbv \in \mbS_\mbv) > 0$ where
        \begin{align}\label{eq:appsec-dependence}
\forall \mba \in \mbS_\mbv, \quad   \mby \,\, \nindep \,\, \mbE_\mbv \g \mbx_\mbv = \mba.
        \end{align}
\end{lemma}
\begin{proof}
\encdef{} says that the explanation $e(\mbx)$ is encoding if there exists an $\mbS$ where $q(\xex \in \mbS) > 0$ such that for every $(\mbv, \mba) \in \mbS$ \cref{eq:appsec-dependence} holds.
This proof works by showing that $\mbS$ having a positive measure implies the existence of $\mbv$ and $\mbS_\mbv$ as in~\Cref{lemma:connection-lemma} such that \cref{eq:appsec-dependence} holds.

Decompose $q(\xex \in \mbS)$ by introducing an expectation over $\mbv \sim q(e(\mbx))$,
\[q(\xex \in \mbS) = \E_{\mbv \sim q(e(\mbx))}q(\xex \in \mbS \g e(\mbx) = \mbv ).\]
As there are only finitely many $\mbv$, 
\[q(\xex \in \mbS) > 0 \quad  \Longleftrightarrow \quad  \exists \mbv \quad  \text{s.t.} \quad  q(e(\mbx) = \mbv) > 0  \quad \text{ and } \quad q(\xex \in \mbS \g e(\mbx) = \mbv) > 0. 
\]

\paragraph{The "only if" direction.}
Pick any $\mbv$ such that the RHS above holds and define $\mbS_\mbv = \{\mba : (\mbv, \mba) \in \mbS\}$.
By definition,
\[\mbS_\mbv = \{\mbx_\mbv: (\mbv, \mbx_\mbv) \in \mbS\} \cap \{\mbx_\mbv : e(\mbx) = \mbv\} \subseteq \{\mbx_\mbv : e(\mbx) = \mbv\}.\]
This proves that $\mbS_\mbv$ has positive measure:
\[q(\mbx_\mbv \in \mbS_\mbv) = q(\xex\in \mbS, e(\mbx) = \mbv) = q(\xex \in \mbS \g e(\mbx) = \mbv) * q(e(\mbx) = \mbv) >0.\]

Finally, as $\mba\in \mbS_\mbv\implies (\mbv, \mba)\in \mbS$, \cref{eq:appsec-dependence} holds:
 \[\forall \mba \in \mbS_\mbv, \quad   \mby \,\, \nindep \,\, \mbE_\mbv \g \mbx_\mbv = \mba.\]
This completes the "only if"  direction.

\paragraph{The "if" direction.}
Assume that there exists $\mbv$ such that $ q(e(\mbx) = \mbv) > 0$   and $\mbS_\mbv\subseteq \{\mbx_\mbv: e(\mbx)=\mbv\}$ such that $q(\mbx_\mbv\in \mbS_\mbv) > 0$ where
\[\forall \mba \in \mbS_\mbv \qquad \mby \nindep \mbE_\mbv \g \mbx_\mbv = \mba.\]
Define $\mbS = \{(\mbv, \mba) : \mba\in \mbS_\mbv\}$.
By this construction, $\mbS$ has positive measure:
\begin{align*}
	q(\xex \in \mbS) & = q( (\mbv, \mbx_\mbv) \in \mbS)
\\
	 & = q(e(\mbx) = \mbv)q((\mbv , \mbx_\mbv) \in \mbS \g e(\mbx) = \mbv) 
\\
 	 & =  q(e(\mbx) = \mbv)q(\mbx_\mbv\in \mbS_\mbv \g e(\mbx) =\mbv) 
\\
 	 & =  q(e(\mbx) = \mbv) q(\mbx_\mbv\in \mbS_\mbv) \qquad \{\text{ as } \mbS_\mbv\subseteq \{\mbx_\mbv: e(\mbx)=\mbv\}
\\
 	& > 0,
\end{align*}
where the last inequality holds because by assumption 
\[q(e(\mbx) = \mbv) > 0\qquad  q(\mbx_\mbv\in \mbS_\mbv) > 0.\]

Finally, as $(\mbv, \mba)\in \mbS\implies \mba \in \mbS_\mbv$, \cref{eq:appsec-dependence} holds:
 \[\forall (\mbv, \mba)\in \mbS, \quad   \mby \,\, \nindep \,\, \mbE_\mbv \g \mbx_\mbv = \mba.\]

This completes the "if" directions and with that the proof.
\end{proof}

\begin{lemma}
\label{lemma:positiveind}
    For any set $\mbS_\mbv \subseteq \{\mbx_\mbv:  e(\mbx) = \mbv\}$ such that $q( \mbx_\mbv \in \mbS_\mbv) > 0$, then for almost every $\mba \in \mbS_\mbv $, $q(\mbE_\mbv=1 \g \mbx_\mbv= \mba) >0$.
\end{lemma}
\begin{proof}
Define the set $\mbA_\mbv = \{\mba: q(\mbE_\mbv=1 \g \mbx_\mbv = \mba) = 0 \}$. Next compute the joint probability
\begin{align*}
    q(\mbx_\mbv \in \mbA_\mbv \cap \mbS_\mbv) = q(\mbx_\mbv \in \mbA_\mbv) q(\mbx_\mbv \in \mbS_\mbv \g \mbx_\mbv \in \mbA_\mbv).
\end{align*}
Now, noting that $\mbS_\mbv$ is a subset of $\{\mbx_\mbv: e(\mbx) = \mbv\}$, which is equivalent to $\{\mbx_\mbv: \mbE_\mbv = 1\}$, thus
\begin{align*}
    &q( \mbx_\mbv \in \mbS_\mbv \g \mbx_\mbv \in \mbA_\mbv) \\
    &= \int q(\mbx_\mbv \in \mbS_\mbv \g \mbx_\mbv = \mba, \mbx_\mbv \in \mbA_\mbv) q(\mbx_\mbv = \mba \g \mbx_\mbv \in \mbA_\mbv) d\mba
    \\
    &= \int q(\mbx_\mbv \in \mbS_\mbv \g \mbx_\mbv = \mba) q(\mbx_\mbv = \mba \g \mbx_\mbv \in \mbA_\mbv)d\mba
    \\
    &\leq \int q(\mbE_\mbv = 1 \g \mbx_\mbv = \mba) q(\mbx_\mbv = \mba \g \mbx_\mbv \in \mbA_\mbv)d\mba
    \\
    &= \int q(\mbE_\mbv = 1 \g \mbx_\mbv = \mba) q(\mbx_\mbv = \mba \g \mbx_\mbv \in \{\mba: q(\mbE_\mbv=1 \g \mbx_\mbv = \mba) = 0 \})d\mba
    \\
    &= 0.
\end{align*}
The probability $q( \mbx_\mbv \in \mbS_\mbv \g \mbx_\mbv \in \mbA_\mbv)$ is non-negative, so it must be zero.
Plugging this conditional back into the joint gives $q(\mbx_\mbv \in \mbA_\mbv \cap \mbS_\mbv) = 0$. Then expanding yields
\begin{align*}
0 = q(\mbx_\mbv \in \mbA_\mbv \cap \mbS_\mbv) = q(\mbx_\mbv \in \mbA_\mbv \g \mbx_\mbv \in \mbS_\mbv) q(\mbx_\mbv \in \mbS_\mbv).
\end{align*}
Since $q(\mbx_\mbv \in \mbS_\mbv) > 0$, $q(\mbx_\mbv \in \mbA_\mbv \g \mbx_\mbv \in \mbS_\mbv)$ must be zero and thus, $q(\mbx_\mbv \notin \mbA_\mbv \g \mbx_\mbv \in \mbS_\mbv) = 1$, where expanding out the definition of $\mbA_\mbv$  gives the desired result that $q(\mbE_\mbv=1 \g \mbx_\mbv = \mba) > 0$ for almost $\mba \in \mbS_\mbv$: 
\begin{align*}
1 &= q(\mba \notin \mbA_\mbv \g \mba \in \mbS_\mbv) &
\\
    & = q(\mba \notin \{\mba: q(\mbE_\mbv=1 \g \mbx_\mbv = \mba) = 0 \} \g \mba \in \mbS_\mbv) 
\\
    & =  q(\mba \in \{\mba: q(\mbE_\mbv=1 \g \mbx_\mbv = \mba) > 0 \} \g \mba \in \mbS_\mbv).
\end{align*}
\end{proof}

\subsubsection{Optimal value and the optimal gap under \texorpdfstring{\acrshort{EVAL-X}}{}}

\newcommand{\evaloptlemma}{
The \acrshort{EVAL-X} optimality gap value for $e(\cdot)$ is an averaged $\kld$ between $q(\mby\g \mbx)$ and $q(\mby\g\mbx_\mbv)$:
$
    \sum_{\mbv \in {\cal V}} q(e(\mbx)=\mbv) \E_{q(\mbx \g e(\mbx)=\mbv)} \kld(q(\mby \g \mbx) \| q(\mby \g \mbx_{\mbv})).
$
This gap is zero, i.e. $e(\mbx)$ is optimal with the score $\acrshort{EVAL-X}^*=\E_q [\log q(\mby \g \mbx)]$ if for all $\mbv$ such that $q(e(\mbx)= \mbv) > 0,$
    \begin{align*}
    q(\mby \g \mbx) = q(\mby \g \mbx_\mbv) \quad \text{ a.e. in } \quad \{\mbx : e(\mbx) = \mbv \}.
\end{align*} 
}
\begin{lemma}\label{lemma:evalopt}
\evaloptlemma{}
\end{lemma}
\begin{proof}
Let $p$ be a generic conditional distribution and let $\mbx_{-\mbv}$ be the values outside the explanation.
\begin{align*}
     \max_{e}  \, \acrshort{EVAL-X}(q, e) &= 
\max_{e} \E_{(\mbv, \mba) \sim q(\xex)} \E_{q( \mby \g \xex = (\mbv, \mba))}\log q\left(\mby \g \mbx_\mbv=\mba\right)  
\\  &= 
\max_{e} \E_{(\mbv, \mba) \sim q(\xex)}
 			\E_{q(\mbx_{-\mbv}\g \xex=(\mbv, \mba))}\E_{q( \mby \g \xex=(\mbv, \mba),\mbx_{-\mbv})}\log q\left(\mby \g \mbx_\mbv=\mba\right)     
\\  &= 
\max_{e} \E_{(\mbv, \mba) \sim q(\xex)}
 			\E_{q(\mbx\g \xex=(\mbv, \mba))}\E_{q( \mby \g \mbx )}\log q\left(\mby \g \mbx_\mbv=\mba\right)     
\\ & 
	\, \leq \max_{p}  \E_{(\mbv, \mba) \sim q(\xex)}
 			\E_{q(\mbx\g \xex=(\mbv, \mba))}\E_{q( \mby \g \mbx )}[\log p(\mby \g \mbx)]
\\ & 
	\, \leq \max_{p}
 			\E_{q(\mbx)}\E_{q( \mby \g \mbx )}[\log p(\mby \g \mbx)]
\\ & 
	\,= \max_{p} -\E_{q(\mbx)}
    \kld
    (q(\mby \g \mbx) || p(\mby \g \mbx)) + \E_q [\log q(\mby \g \mbx)]
    \\ & \,= \E_q [\log q(\mby \g \mbx)].
\end{align*}
This upper bound is achievable by an explanation that selects all inputs, so the maximum \acrshort{EVAL-X} denoted as $\acrshort{EVAL-X}^* = \E_q \log q(\mby \g \mbx)$. 

As in the math above, the \acrshort{EVAL-X} score for an explanation method can be expanded as 
\begin{align*}
    \acrshort{EVAL-X}^e &= \E_{(\mbv, \mba) \sim q(\xex)} \E_{q( \mby \g \xex = (\mbv, \mba))}\log q\left(\mby \g \mbx_\mbv=\mba\right)     
    \\
    &= \E_{(\mbv, \mba) \sim q(\xex)}
 			\E_{q(\mbx\g \xex=(\mbv, \mba))}\E_{q( \mby \g \mbx )}\log q(\mby \g \mbx_\mbv=\mba)
    \\
    &= \E_{\mbv \sim q(e(\mbx))} \E_{\mba \sim q(\mbx_\mbv\g e(\mbx)=\mbv)}
 			\E_{q(\mbx\g \mbx_\mbv=\mba, e(\mbx)=\mbv))}\E_{q( \mby \g \mbx )}\log q(\mby \g \mbx_\mbv=\mba)
    \\
    &= \E_{\mbv \sim q(e(\mbx))} \E_{q(\mbx\g e(\mbx)=\mbv))}\E_{q( \mby \g \mbx )}\log q(\mby \g \mbx_\mbv),
\end{align*}
where in the last step, we dropped $\mba$ because it equals $\mbx_\mbv$ almost surely.
Similarly, the optimal score $\acrshort{EVAL-X}^*$ expands to
\begin{align*}
   \E_{\mbv \sim q(e(\mbx))} \E_{q(\mbx\g e(\mbx)=\mbv))}\E_{q( \mby \g \mbx )} \log q(\mby \g \mbx).
\end{align*}
Let ${\cal V}$ be the set of values that explanations can take on, then taking the difference from optimality 
\begin{align*}
    &\acrshort{EVAL-X}^*  - \acrshort{EVAL-X}^{e} \nonumber
\\
    &=  \E_{\mbv \sim q(e(\mbx))} \E_{q(\mbx\g e(\mbx)=\mbv))}\E_{q( \mby \g \mbx )}\log \frac{q(\mby\g \mbx)}{q(\mby\g \mbx_\mbv=\mbx)}
\\
	 &=\E_{\mbv \sim q(e(\mbx))} \E_{q(\mbx\g e(\mbx)=\mbv))}
 			\KL{q(\mby \g \mbx)}{q(\mby\g \mbx_\mbv)}
\\
    &= \sum_{\mbv \in {\cal V}} q(e(\mbx)=\mbv) \E_{q(\mbx\g e(\mbx)=\mbv))}
 			\KL{q(\mby \g \mbx)}{q(\mby\g \mbx_\mbv)}.
\end{align*} 
As each $\kld$ term is non-negative, each term in the sum being set to $0$ simultaneously achieves the optimum, which happens when for all $\mbv$ such that $q(e(\mbx)= \mbv) > 0,$
\begin{align*}
    q(\mby \g \mbx) = q(\mby \g \mbx_\mbv) \quad \text{ for almost every } \quad \{\mbx : e(\mbx) = \mbv \}.
\end{align*} 
\end{proof}

In ~\Cref{sec:proof-eval-x}, we use the results from \Cref{lemma:defeq} and \Cref{lemma:evalopt} to prove that the optimal score of \evalx{} can only be achieved by non-encoding explanations.

\subsection{Proof of\texorpdfstring{ \Cref{thm:opt-removes-encoding}}{}}
\label{sec:proof-eval-x}

\setcounter{thm}{0}
\begin{thm}
\optencthm{}    
\end{thm}
\begin{proof}
Note only $q(e(\mbx) = \mbv) > 0$ are of interest, since $q(e(\mbx) = \mbv) = 0$ implies that $\mbE_\mbv = 0$ almost surely and thus $\mby \indep \mbE_\mbv \g \mbx_\mbv$.

Then if $e(\mbx)$ achieves \acrshort{EVAL-X}$^*$, then by \Cref{lemma:evalopt}, for all $\mbv$ such that $q(e(\mbx)= \mbv) > 0,$
    \begin{align*}
    q(\mby \g \mbx) = q(\mby \g \mbx_\mbv) \quad \text{ for almost every } \quad \{\mbx : e(\mbx) = \mbv \}.
\end{align*} 
First, this optimality criteria can incorporate $\mbE_\mbv=1$ on the lefthand side by first conditioning on $e(\mbx)$ and then noting that the equality holds for $\mbx$ where $e(\mbx) = \mbv$.
\begin{align*}
    q(\mby \g \mbx) &= q(\mby \g \mbx_\mbv) \quad \text{ for almost every } \quad \{\mbx : e(\mbx) = \mbv \}
    \\ \iff
    q(\mby \g \mbx, e(\mbx)) &= q(\mby \g \mbx_\mbv) \quad \text{ for almost every } \quad \{\mbx : e(\mbx) = \mbv \}
    \\ \iff
    q(\mby \g \mbx, e(\mbx)=\mbv) &= q(\mby \g \mbx_\mbv) \quad \text{ for almost every } \quad \{\mbx : e(\mbx) = \mbv \}
    \\ \iff
    q(\mby \g \mbx, \mbE_\mbv=1) &= q(\mby \g \mbx_\mbv) \quad \text{ for almost every } \quad \{\mbx : e(\mbx) = \mbv \}.
\end{align*}

To understand if the optimality criterion disallows encoding, integrate the left and right-hand sides of this optimality criterion with the respect to complement of the inputs in $\mbx_\mbv$, $q(\mbx_\mbv^c \g \mbx_\mbv, \mbE_\mbv=1)$ yields
\begin{align*}
    &\int q(\mby \g \mbx, \mbE_\mbv=1) q(\mbx_\mbv^c \g \mbx_\mbv, \mbE_\mbv=1) d\mbx_\mbv^c  
    = 
    \int q(\mby \g \mbx_\mbv)q(\mbx_\mbv^c \g \mbx_\mbv, \mbE_\mbv=1)  d\mbx_\mbv^c \\ & \qquad \qquad \qquad \qquad \qquad \qquad \qquad \quad \quad \,\, \text{ for almost every } \{\mbx : e(\mbx) = \mbv \}
    \\
    &\iff \int q(\mby \g \mbx_\mbv^c, \mbx_\mbv, \mbE_\mbv=1) q(\mbx_\mbv^c \g \mbx_\mbv, \mbE_\mbv=1) d\mbx_\mbv^c  
    = 
    q(\mby \g \mbx_\mbv) \int q(\mbx_\mbv^c \g \mbx_\mbv, \mbE_\mbv=1)  d\mbx_\mbv^c 
\\ & \qquad \qquad \qquad \qquad \qquad \qquad \qquad \quad \quad \,\, \text{ for almost every } \{\mbx : e(\mbx) = \mbv \} \\
    &\iff
    q(\mby \g \mbx_\mbv, \mbE_\mbv=1) = q(\mby \g \mbx_\mbv) \quad \text{ for almost every } \{\mbx : e(\mbx) = \mbv \} \\
    &\iff
    q(\mby \g \mbx_\mbv, \mbE_\mbv=1) = q(\mby \g \mbx_\mbv) \quad \text{ for almost every } \{\mbx_\mbv : e(\mbx) = \mbv \}.
\end{align*}
Now expanding the right-hand side gives
\begin{align*}
    q(\mby \g \mbx_\mbv) = q(\mby, \mbE_\mbv=1 \g \mbx_\mbv) + q(\mby, \mbE_\mbv=0 \g  \mbx_\mbv).
\end{align*}
Combing the two equations gives
\begin{align}
    q(\mby \g \mbx_\mbv, \mbE_\mbv=1) = q(\mby, \mbE_\mbv=1 &\g \mbx_\mbv) + q(\mby, \mbE_\mbv=0 \g \mbx_\mbv) \quad
    \text{ for almost every } \{\mbx_\mbv : e(\mbx) = \mbv \}.
    \label{eq:optimaility-new}
\end{align}
We show that this equality implies that $\mby \indep \mbE_\mbv \g \mbx_\mbv$ by splitting the analysis into cases based on $q(\mbE_\mbv=1 \g \mbx_\mbv)=1$ and $q(\mbE_\mbv=1 \g \mbx_\mbv)<1$. In turn, the condition in \encdef{} is violated and the explanation $e(\cdot)$ is not encoding.

\paragraph{Case 1: \acrshort{EVAL-X} optimality holds when the explanation is predictable.}
The first case to consider is when the event that the explanation takes the value $\mbv$ is determined by $\mbx_\mbv$ for all samples with the explanation $\mbv$. That is, $q(\mbE_\mbv=1 \g \mbx_\mbv)=1$:
\begin{align*}
    q(\mbE_\mbv=1 \g \mbx_\mbv)=1 \iff q(\mbE_\mbv=0 \g \mbx_\mbv)=0.
\end{align*}
Then expanding this marginal into the joint shows that the joint $q(\mby, \mbE_\mbv=0 \g \mbx_\mbv)$ has to be zero as well. 
\begin{align*}
    q(\mbE_\mbv=0 \g \mbx_\mbv) = \int q(\mby, \mbE_\mbv=0 \g \mbx_\mbv) d\mby = 0,
\end{align*}
because an integral of non-negative terms being zero implies that each term itself is zero almost surely.

Then, we can show that the determinism condition  $q(\mbE_\mbv=1 \g \mbx_\mbv)=1$ is sufficient for the optimality criterion \cref{eq:optimaility-new}:
\begin{align*}
    q(\mby \g \mbx_\mbv, \mbE_\mbv=1)
    & = q(\mby \g \mbx_\mbv, \mbE_\mbv=1) \times 1
\\
    & = q(\mby \g \mbx_\mbv, \mbE_\mbv=1)  q(\mbE_\mbv=1 \g \mbx_\mbv)
\\
    & = q(\mby, \mbE_\mbv=1 \g \mbx_\mbv) \nonumber \\
    & = q(\mby, \mbE_\mbv=1 \g \mbx_\mbv) + q(\mby, \mbE_\mbv=0 \g \mbx_\mbv) \nonumber \quad
    \text{ for almost every } \{\mbx_\mbv : e(\mbx) = \mbv \}.
\end{align*}

This shows that the \acrshort{EVAL-X} optimality criteria is satisfied when the $q(\mbE_\mbv=1 \g \mbx_\mbv)=1$, thus the explanation is completely predictable from the explanation for examples with that explanation. By \Cref{lemma:defeq}, we have
\[q(\mbE_\mbv=1 \g \mbx_\mbv)=1 \implies \mby\indep \mbE_\mbv \g \mbx_\mbv,\]
which violates \encdef{}.
So there is no encoding in this case. 

\paragraph{Case 2: When the explanation is unpredictable, \acrshort{EVAL-X} optimality requires that the explanation provide no extra information.}
Now consider the alternative case, $q(\mbE_\mbv=1 \g \mbx_\mbv)<1$. Here the explanation does not determine the explanation opening the possibility that the \acrshort{EVAL-X}-optimal explanation method can encode information in the explanation.

Because $q(e(\mbx) = \mbv) > 0$, we have $q(\mbx_\mbv \in \{\mbx_\mbv: e(\mbx) = \mbv\}) > 0$. Thus, by \Cref{lemma:positiveind}, for almost every $\{\mbx_\mbv: e(\mbx) = \mbv\}$ it holds that $q(\mbE_\mbv = 1 \g \mbx_\mbv) > 0.$
Putting this result together with alternative case ($q(\mbE_\mbv=1 \g \mbx_\mbv)<1$) gives: $0 < q(\mbE_\mbv=1 \g \mbx_\mbv) < 1$ for almost every $\{\mbx_\mbv: e(\mbx) = \mbv\}$.

Now, expanding out the optimality criterion \cref{eq:optimaility-new}:
\begin{align*}
    &q(\mby \g \mbx_\mbv, \mbE_\mbv=1) = q(\mby, \mbE_\mbv=1 \g \mbx_\mbv) \times 1 + q(\mby, \mbE_\mbv=0 \g \mbx_\mbv) \times 1 
    \\
     & \qquad \qquad \qquad \qquad \qquad \qquad \qquad \qquad \qquad \qquad
    \text{ for almost every } \{\mbx_\mbv : e(\mbx) = \mbv \} \\
    &\iff q(\mby \g \mbx_\mbv, \mbE_\mbv=1) = q(\mby \g \mbE_\mbv=1, \mbx_\mbv) q(\mbE_\mbv=1 \g \mbx_\mbv) 
    \\ &\qquad\qquad\qquad\qquad\quad\quad\,\, + q(\mby \g \mbE_\mbv=0, \mbx_\mbv) (1-q(\mbE_\mbv=1 \g \mbx_\mbv)) \nonumber 
\\
     & \qquad \qquad \qquad \qquad \qquad \qquad \qquad \qquad \qquad \qquad
    \text{ for almost every } \{\mbx_\mbv : e(\mbx) = \mbv \} \\
    &\iff
    (1 - q(\mbE_\mbv=1 \g \mbx_\mbv)) q(\mby \g \mbx_\mbv, \mbE_\mbv=1) 
    =(1-q(\mbE_\mbv=1 \g \mbx_\mbv)) q(\mby \g \mbx_\mbv, \mbE_\mbv=0) 
\\
     & \qquad \qquad \qquad \qquad \qquad \qquad \qquad \qquad \qquad \qquad
    \text{ for almost every } \{\mbx_\mbv : e(\mbx) = \mbv \}
    \\
    &\iff
    q(\mby \g \mbx_\mbv, \mbE_\mbv=1) = q(\mby \g \mbx_\mbv, \mbE_\mbv=0)
    \quad \text{ for almost every } \{\mbx_\mbv : e(\mbx) = \mbv\}.
\end{align*}
This equality says for all samples with the explanation $\mbv$, knowing $\mbE_\mbv$ does not change the distribution of the label $\mby$.
By \Cref{lemma:defeq}, this equality implies that
the independence $\mby \,\,\indep \,\, \mbE_\mbv \g \mbx_\mbv$ holds which violates \encdef{}.
\end{proof}

\subsection{Proof of \texorpdfstring{\Cref{prop:encmeaszero}}{} and \texorpdfstring{\Cref{thm:strong-det}}{}}\label{appsec:hy}
\setcounter{prop}{2}
\begin{prop}
    \encmeaszero{}
\end{prop}
\begin{proof}
First, \encdef{} is violated if and only if there exists a set $\mbA$ such that $q(\xex \in \mbA)=1$ and 
        \[ \forall (\mbv,\mba)\in \mbA \qquad \mby \indep \mbE_\mbv \g \mbx_\mbv = \mba.\]

For the if direction, note that if \encmeas{} $\phi_q(e)=0$,
\begin{align*}
\E_{(\mbv, \mba) \sim q(\xex)}  \mbI(\mby; \mbE_\mbv \g \mbx_\mbv = \mba) &= 0.
\end{align*}
To show the forward direction, the above equality means that if $\phi_q(e)=0$, almost surely for every $(\mbv, \mba)\sim q(\xex)$, the instantaneous mutual information is $0$ which implies the desired conditional independence
\[\mbI(\mby; \mbE_\mbv \g \mbx_\mbv = \mba)= 0 \implies \mby \indep \mbE_\mbv \g \mbx_\mbv = \mba.\]
By definition of almost surely, there exists a set $\mbA$ such that $q(\xex \in \mbA)=1$ the independence above holds; this completes the "if" direction.

To show the reverse direction, let there exist a set $\mbA$ such that $q(\xex \in \mbA)=1$, for every $(\mbv,\mba)\in \mbA$,
\[\mby \indep \mbE_\mbv \g \mbx_\mbv.\]
In turn, for all $(\mbv,\mba)\in \mbA$,
\[\mbI(\mby; \mbE_\mbv \g \mbx_\mbv = \mba)=0.\]
Then, the fact that $q(\xex \in \mbA)=1$ implies that expectations with respect to $q(\xex)$ over the whole support equal expectations over $q(\xex\g \xex\in \mbA)$, which is  $q(\xex)$ restricted to $\mbA$: 
\[\E_{(\mbv, \mba) \sim q(\xex)}  \mbI(\mby; \mbE_\mbv \g \mbx_\mbv = \mba) = \E_{(\mbv, \mba) \sim q(\xex\g \xex \in \mbA)}  \mbI(\mby; \mbE_\mbv \g \mbx_\mbv = \mba) = 0.\]
This completes the "only if" direction.
\end{proof}

\begin{thm}
	\strongdet{}
\end{thm}
\begin{proof}
Recall that \detx{} is 
\[\detx{}_\alpha(q,e) := \E_{(\mbv, \mba) \sim q(\xex)} \E_{q( \mby \g \xex = (\mbv, \mba))}[\log q\left(\mby \g \mbx_\mbv = \mba\right)]  - \alpha \phi_q(e),\]
  where the \encmeas{}
  \[ \phi_q(e) := \E_{(\mbv, \mba)\sim q(\xex)}\mbI\left( \mbE_\mbv; \mby \g \mbx_\mbv = \mba \right).\] 
We first show bounds for the first term in \acrshort{stripe-x} and then derive the \acrshort{stripe-x} scores for encoding and non-encoding explanations.
\paragraph{Bounds on \acrshort{EVAL-X} scores.}
We lower bound the \acrshort{EVAL-X} score, which is the first term in \acrshort{stripe-x}, for non-encoding explanations.

For non-encoding explanations, almost surely over $\mbv, \mba\sim q(\xex)$
 \[q(\mby\g \xex = (\mbv,\mba)) = q(\mby\g \mbx_\mbv=\mba).\]
Then,
\begin{align*}
	&\evalx(q,e) - \E_{q(\mby, \mbx_\mbv = \mba)}
	\log q(\mby)
\\
 &  =\E_{(\mbv, \mba) \sim q(\xex)} \E_{q( \mby \g \xex = (\mbv, \mba))}\log q\left(\mby \g \mbx_\mbv = \mba\right)  
 -  \E_{q(\mby, \xex)} \log q(\mby)
	\\
& = \E_{(\mbv, \mba) \sim q(\xex)} \E_{q( \mby \g \xex = (\mbv, \mba))}[\log q\left(\mby \g \xex = (\mbv, \mba)\right)] -  \E_{q(\mby, \xex)}
	\log q(\mby)
\\	
& = \E_{q(\mby, \xex)}[\log q\left(\mby \g \xex \right)] -  \E_{q(\mby, \xex)}
	\log q(\mby)
\\
 & = \E_{q(\mby, \xex)} \log\frac{q\left(\mby \g \xex \right)}{q(\mby)}
\\
& 	= \mbI(\mby; \xex).
\end{align*} 
The above inequality implies that
\begin{align*}
	&	\evalx(q,e) - \E_{(\mbv, \mba) \sim q(\xex)}  \E_{\mby \sim q(\mby \g \xex = (\mbv, \mba))} \log q(\mby) = \mbI(\mby; \xex) \geq  0 
\\
 & \implies  \evalx(q, e) + \mbH_q(\mby) \geq 0
\\
	 & \implies \evalx(q, e) \geq - \mbH_q(\mby) 	.
\end{align*}
Every inequality in the derivation above becomes strict when the explanation selects inputs that are predictive of the label because
\[ \mbI(\mby; \xex) >0.\] 
Thus, non-encoding explanations have \evalx{} scores that are at least $-\mbH_q(\mby)$.

The optimal \evalx{} score for any explanation (see~\Cref{lemma:evalopt}) equals the negative conditional entropy  which is upper bounded by some finite number:
\[\E_q\left[\log q(\mby\g \mbx)\right] = -\mbH_q(\mby\g \mbx) = C.\]

\paragraph{Comparing explanations via \acrshort{stripe-x}.}

For any encoding explanation, by \Cref{prop:encmeaszero}, for some $c>0$
\[ \phi_q(e) > c.\]
Now, consider $\alpha^*=  \frac{\mbI(\mby;\mbx)}{c}  \geq 0$, which is finite because each term in the ratio is finite.
Then, for all $\alpha > \alpha^*$
\[\alpha \phi_q(e) > \alpha^* \phi_q(e) > \mbI(\mby;\mbx).\]
Thus, 
\[- \alpha \phi_q(e) <  -\mbI(\mby;\mbx).\] 
As \evalx{} scores are below $C=-\mbH(\mby\g\mbx)$ for any encoding explanation,
\begin{align*}
	\detx{}_\alpha(q,e) &= \evalx{}(q,e) - \alpha\phi_q(e) 
\\
 	& < -\mbH(\mby\g\mbx) -\mbI(\mby;\mbx)
\\
 	& < -\mbH(\mby\g\mbx) -(\mbH_q(\mby) - \mbH(\mby\g \mbx))
\\
	 & = -\mbH_q(\mby).
\end{align*}

Finally, for any non-encoding explanation, $\phi_q(e^\prime)=0$ by \Cref{prop:encmeaszero}, \detx{} scores equal \evalx{} scores, which are lower bounded at $-\mbH_q(\mby)$.

Together, for every non-encoding explanation $e^\prime(\mbx)$ and encoding explanation $e(\mbx)$, it holds that
\[\detx{}_\alpha(q,e^\prime) \geq-\mbH_q(\mby) > \detx{}_\alpha(q,e).\]
This proves that \detx{} is a strong detector of encoding.
\end{proof}

\section{Encoding examples, non-detection of \textsc{roar,fresh}, and non-strong detection of \texorpdfstring{\evalx{}}{} }\label{appsec:example-proofs}

\subsection{An illustrative \texorpdfstring{\gls{dgp}}{} for \texorpdfstring{\encdef{}}{}}
\label{sec:definition-example}

With $\cB(0.5)$ being a Bernoulli distribution, consider the following example
\begin{align*}
&\mby \sim \cB(0.5)\quad,\quad \mbz \sim \cB(0.5) \quad, \quad  \epsilon_1, \epsilon_2, \epsilon_3 \sim \cN\left(0, \mbI\right) ,
\\
& \mbx \quad  = \quad
\begin{cases}
	[\mby + \epsilon_1, \qquad \epsilon_3, 0, \epsilon_2 ] \quad \text{ if } \mbz = 0 ,
	\\
	[\qquad\epsilon_3, \mby + \epsilon_1, 1, \epsilon_2 ] \quad \text{ if } \mbz = 1 .
\end{cases}
\end{align*} 
For this problem, if the third coordinate $\mbx_3=0$, all the information between the label and the covariates is in the first coordinate $\mbx_1$, and if $\mbx_3=1$, the information is between the label and the second coordinate $\mbx_2$.
The corresponding explanation function is $e(\mbx) = \mathbbm{1}[\mbx_3=0]\xi_1 + \mathbbm{1}[\mbx_3=1]\xi_2$.
This explanation is encoding because neither explanation's values $\mbx_1$ nor $\mbx_2$ determine the explanation function because it depends on $\mbx_3$.
Formally
\[ q( \mby =1 \g \mbx_1, \mbx_3=1) \neq q( \mby =1 \g \mbx_1, \mbx_3=0) \implies \mby \nindep \mbx_3 \g \mbx_1 \implies \mby \nindep \mbE_\mbv \g \mbx_1,\]
which meets \encdef{}.
Consider an alternate 
non-encoding 
explanation function  $e(\mbx) = [\mathbbm{1}[\mbx_4 > 0],\mathbbm{1}[\mbx_4 \leq 0],0,0]$; $\mbx_1, \mbx_2$ do not determine $e(\mbx)$ that depends on the noise $\epsilon_2$ in $\mbx_4$. That means the unpredictability property in \Cref{lemma:defeq} holds.
However, by construction,
\[(\mby, \mbx_1, \mbx_2) \,\, \indep \,\, \epsilon_2 \implies (\mby, \mbx_1, \mbx_2) \,\, \indep \,\, \mbE_\mbv \,\, \implies \,\, \mby \indep \mbE_\mbv \g \mbx_1 \quad \text{and} \quad \mby \indep \mbE_\mbv \g \mbx_2. \]
So no additional information about the label is encoded: \[ \mby  \,\, \indep \,\, \mbE_\mbv \g \mbx_\mbv.
\]
The additional information property in \Cref{lemma:defeq} avoids such cases where the explanations keeps  additional information that is irrelevant to the label.

\subsection{Encoding explanations conceal predictive inputs that affect the explanation}
\label{sec:bewilderment-example}

Consider the following \gls{dgp}
\begin{align}
    \mbx & = [\mbx_1, \mbx_2, \mbx_3]  \sim \cB(0.5)^{\otimes 3 }, \qquad \quad
        \mby = \begin{cases}
                     \mbx_1  \quad  \text{ if }  \mbx_3 = 1,
                          \\
                    \mbx_2 \quad  \text{ if }  \mbx_3 = 0.
                \end{cases}
    \label{eq:simple-dgp}
\end{align}

Let $e$ be an encoding explanation that selects the first coordinate if $\mbx_3 = 1$ and the second coordinate otherwise. We never observe $\mbx_3$ when looking only at the explanation. \Cref{tab:bewilderment-example} shows all possible values of this explanation. Notice that in the third and fourth rows, the value of $\mbx_{e(\mbx)}$ changes to match the label $\mby$ exactly, even though the values of the first two coordinates that we can observe stay constant. It is impossible to understand the perfect predictiveness of $\mbx_{e(\mbx)}$, as the encoding explanation conceals the control flow feature $\mbx_3$ that determines which of the first two features should be picked to predict the label.

\begin{table}[ht]
    \centering
    \caption{Possible values of the inputs, label, and explanation for the \gls{dgp} in \cref{eq:simple-dgp}}
    \vspace{5pt}
    \begin{tabular}{cccccc}
        \toprule
        & & & & \multicolumn{2}{c}{$\mbx_{e(\mbx)} = (\mbv, \mba)$} \\
        \textcolor{lightgray}{$\mbx_1$} & \textcolor{lightgray}{$\mbx_2$} & \textcolor{lightgray}{$\mbx_3$} & $\mby$ & $\mbv$ & $\mba$ \\ \midrule
        \textcolor{lightgray}{0} & \textcolor{lightgray}{0} & \textcolor{lightgray}{0} & 0 & [0, 1, 0] & 0 \\
        \textcolor{lightgray}{0} & \textcolor{lightgray}{0} & \textcolor{lightgray}{1} & 0 & [1, 0, 0] & 0 \\
        \textcolor{lightgray}{0} & \textcolor{lightgray}{1} & \textcolor{lightgray}{0} & 1 & [0, 1, 0] & 1 \\
        \textcolor{lightgray}{0} & \textcolor{lightgray}{1} & \textcolor{lightgray}{1} & 0 & [1, 0, 0] & 0 \\
        \textcolor{lightgray}{1} & \textcolor{lightgray}{0} & \textcolor{lightgray}{0} & 0 & [0, 1, 0] & 0 \\
        \textcolor{lightgray}{1} & \textcolor{lightgray}{0} & \textcolor{lightgray}{1} & 1 & [1, 0, 0] & 1 \\
        \textcolor{lightgray}{1} & \textcolor{lightgray}{1} & \textcolor{lightgray}{0} & 1 & [0, 1, 0] & 1 \\
        \textcolor{lightgray}{1} & \textcolor{lightgray}{1} & \textcolor{lightgray}{1} & 1 & [1, 0, 0] & 1 \\ \bottomrule  
    \vspace{-5pt}
    \end{tabular}
    \label{tab:bewilderment-example}
\end{table}

\subsection{Position-based encoding fits \texorpdfstring{\encdef{}}{}}
\label{sec:position-based-example}
Recall the perceptual task that classifying images of dogs versus classifying images of cats, and consider the encoding explanation $e_{\textrm{position}}(\mbx)$ that is
\begin{align*}
    e_{\textrm{position}}(\mbx) &= \xi_1 \quad \text{ if } \quad q(\mby = \textrm{dog} \g \mbx) = 1, \\
    e_{\textrm{position}}(\mbx) &= \xi_2 \quad \text{ if } \quad q(\mby = \textrm{cat} \g \mbx) = 1.
\end{align*}

Assume that the inputs in the top leftmost pixels are always background, meaning that the values of these inputs provide no information about the label $\mby \indep \mbx_1, \mbx_2$. Now we check if this intuitively-defined position-encoded explanation meets the definition for encoding (\encdef{}). 
To condition on $\mbx_{\xi_1}, \mbE_{\xi_1}=0$, we need $q(\mby = \textrm{dog}) \neq 1$.
Note that
\begin{align*}
    q(\mbE_{\xi_1} = 1 \g \mbx_{\xi_1}) = q(\mby = \textrm{dog} \g \mbx_{\xi_1}) = q(\mby = \textrm{dog}) \neq 1.
\end{align*}
\encdef{} holds because the indicator of which explanation was chosen $\mbE_{\xi_1}$ determines the label.
\begin{align*}
    q(\mby = \textrm{dog} \g \mbx_{\xi_1}, \mbE_{\xi_1} = 1) = 1 \neq 
    0=
    q(\mby = \textrm{dog} \g \mbx_{\xi_1}, \mbE_{\xi_1} = 0) 
    \implies \mby \,\, \nindep \,\, \mbE_{\xi_1} \g \mbx_{\xi_1}.
\end{align*}
This example shows how the encoding definition \encdef{} captures the informally described position-based encoding from the literature.

\subsection{Prediction-based encoding fits \texorpdfstring{\encdef{}}{}}
\label{sec:prediction-based-example}
The informal example of prediction-based encoding from \Cref{subsec:examples} selects a single input that makes the prediction from all of the input have the highest confidence when given the single input.
One way to mathematically express such a selection is as follows:
\begin{align}
\label{eq:predenc}
\begin{split}
    e_\textrm{prediction}(\mbx) &= \xi_{\text{argmax}_{i} q(\mby = 1 \g \mbx_i)} \text{ if } q(\mby = 1 \g \mbx) > 0.5 , \\
    e_\textrm{prediction}(\mbx) &= \xi_{\text{argmin}_{i} q(\mby = 1 \g \mbx_i)} \text{ if } q(\mby = 1 \g \mbx) \leq 0.5.
\end{split}
\end{align}

\newcommand{\mbxxi}[1]{\mbx_{\xi_{#1}}}
\newcommand{\mbexi}[1]{\mbE_{\xi_{#1}}}

Here, we describe one set of conditions on the distribution $q(\mby, \mbx)$ for which the explanation in \cref{eq:predenc} fits the definition of encoding in \encdef{}.
Assume that there exists a non-measure-zero set $\mbU \subseteq \{\mbx : q(\mby = 1 \g \mbx) > 0.5\}$ and an index $k$ such that
 \begin{align}
     \mbx & \in \mbU \implies q(\mby=1 \g \mbx) \geq \rho ,\label{eq:rhomax}
\\
     \mbx & \in \mbU \implies \forall i \quad  q(\mby = 1 \g \mbxxi{i} ) < q(\mby = 1\g \mbxxi{k}) ,\label{eq:udetpos}
\\
     \mbx & \not\in \mbU \implies q(\mby=1 \g \mbx) < \rho, \label{eq:rholess}
\\
     \mbx & \not\in \mbU \implies \exists i,j \quad  q(\mby = 1 \g \mbxxi{i} ) > q(\mby = 1\g \mbxxi{k}) > q(\mby = 1 \g \mbxxi{j}). \label{eq:udet}
 \end{align}
Further, assume that 
$\mbxxi{k}$ alone does not determine $\mbx\in \mbU$:
\begin{align}
0 & < \E\left[\mathbbm{1}[\mbx\in \mbU] \g \mbx_{\xi_k}\right ]  < 1. \label{eq:boundedaway}
\end{align}

The assumptions above imply the facts below about $e_\textrm{prediction}(\mbx)$:
\begin{enumerate}
\item By~\cref{eq:udetpos,eq:udet}
        \begin{align*}
            \mbx \in \mbU \Leftrightarrow  e_\textrm{prediction}(\mbx) = \xi_k.
        \end{align*}
    Define the explanation indicator $\mbE_\mbv = \mathbbm{1}[ e_\textrm{prediction}(\mbx) = \mbv]$.
\item  By~\cref{eq:udetpos,eq:udet} and the definition of $\mbE_\mbv$
    \begin{align}
        \mbx \in \mbU \Leftrightarrow \mbexi{k}=1. \label{eq:uandeimplication}
    \end{align}
\item By \cref{eq:uandeimplication,eq:boundedaway}
\begin{align}
    0 < q(\mbexi{k} = 1 \g \mbxxi{k}) = q(\mbx \in\mbU \g \mbxxi{k})  < 1 .\label{eq:boundedek1}
\end{align}
\item  
    By \cref{eq:boundedek1}, $q(\mby = 1 \g \mbxxi{k}, \mbexi{k}=1)$ and  $q(\mby = 1 \g \mbxxi{k}, \mbexi{k}=0)$ are well defined.
Then, by~\cref{eq:rhomax,eq:rholess}, for $\mbx\in \mbU$
\begin{align*}
    q(&\mby=1 \g \mbx_{\xi_k}, \mbE_k=1)
\\ & = \E_{q(\mbx \g \mbx_{\xi_k}, \mbE_k=1)} q(\mby \g \mbx) 
\\ &  \geq \E_{q(\mbx \g \mbx_{\xi_k}, \mbE_k=1)} \rho  \qquad \qquad \{\text{as } \mbE_k=1 \implies \mbx \in \mbU\}
\\ & = \rho.
\\ 
\\ q(&\mby=1 \g \mbx_{\xi_k}, \mbE_k=0) 
\\ & = \E_{q(\mbx \g \mbx_{\xi_k}, \mbE_k=0)} q(\mby \g \mbx) 
\\ &< \E_{q(\mbx \g \mbx_{\xi_k}, \mbE_k=0)} \rho \qquad\qquad \{ \text{as } \mbE_k=0 \implies \mbx \not\in \mbU\}
\\ & = \rho.
\end{align*}
\end{enumerate}
Thus, for all elements of $\{\mbx_{\xi_k} : \mbx \in \mbU\}$
\begin{align}
    q(\mby = 1 \g \mbx_{\xi_k}, \mbE_{\xi_k} = 1) > q(\mby = 1 \g \mbx_{\xi_k}, \mbE_{\xi_k} = 0). \label{eq:mismatch}
\end{align}
By~\Cref{lemma:defeq}, the properties in  \cref{eq:mismatch,eq:boundedek1} imply that $\forall \mba \in \{\mbx_{\xi_k} : \mbx \in \mbU\}$
\[\mby\nindep \mbexi{k} \g \mbxxi{k} = \mba.\]
Finally, the set $\mbU_k = \{\mbx_{\xi_k} : \mbx \in \mbU\}$ is non-measure-zero: as $\mathbbm{1}[\mbx \in \mbU] =1 \implies \mathbbm{1}[\mbx_{\xi_k}\in \mbU_k]=1$, accumulating $q(\mbx)$ with the restriction $\mbx_{\xi_k}\in \mbU_k$ leads to at least as much mass as accumulating with the stricter restriction $\mbx\in \mbU$:
\[ q(\mbx_{\xi_k} \in \mbU_k) = \int q(\mbx) \mathbbm{1}[\mbx_{\xi_k}\in \mbU_k] d \mbx \geq \int q(\mbx) \mathbbm{1}[\mbx \in \mbU] d \mbx  = q(\mbx \in \mbU)> 0.\]
Together, the last two equations implies that \encdef{} holds for $e_\textrm{prediction}(\mbx)$ from~\cref{eq:predenc}.

\subsection{\texorpdfstring{\acrshort{margenc}}{} explanations are encoding}\label{appsec:margenc}

\begin{figure}[t]
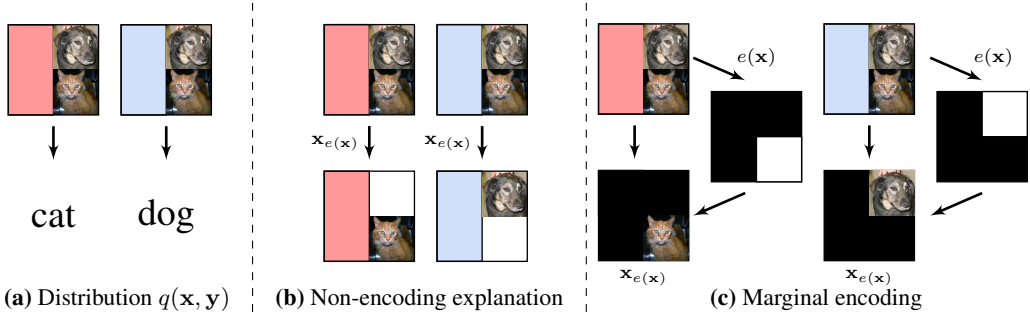

\vspace{-5pt}
\def\scale{0.6}
\centering
\begin{subfigure}[b]{0.24\textwidth}
\centering
\begin{tikzpicture}[
        >=latex',
        circ/.style={draw=white, shape=circle, node distance=1cm,minimum width=0.9cm,fill=CornflowerBlue!30},
        rect/.style={draw=black, shape=rectangle, node distance=1cm, line width=0.5pt},
        inner/.style={shape=rectangle, node distance=1cm, line width=0.0pt}]
    \node[rect,fill=red!40,minimum width=\scale cm,minimum height=2*\scale cm] at (0.5*\scale,-\scale) {};
    \node[inner] at (1.5*\scale,-0.5*\scale) 
        {\includegraphics[width=\scale cm]{figures/dog.jpeg}};
    \node[inner] at (1.5*\scale,-1.5*\scale) 
        {\includegraphics[width=\scale cm]{figures/cat.jpeg}};
    \node[rect,draw,minimum width=2*\scale cm, minimum height=2*\scale cm] at (\scale,-\scale) {};

    \node[rect,fill=CornflowerBlue!30,minimum width=\scale cm,minimum height=2*\scale cm] at (3*\scale,-\scale) {};
    \node[inner] at (4*\scale,-0.5*\scale) 
        {\includegraphics[width=\scale cm]{figures/dog.jpeg}};
    \node[inner] at (4*\scale,-1.5*\scale) 
        {\includegraphics[width=\scale cm]{figures/cat.jpeg}};
    \node[rect,draw,minimum width=2*\scale cm, minimum height=2*\scale cm] at (3.5*\scale,-\scale) {};
    \node[inner,minimum height=6.25*\scale cm] at (\scale,-4.25*\scale) {\Large cat};
    \node[inner] at (3.5*\scale,-4.25*\scale)  {\Large dog};
    \draw[->, line width=1pt] (\scale,-2.25*\scale) -- (\scale,-3*\scale) {};

    \draw[->, line width=1pt] (3.5*\scale,-2.25*\scale) -- (3.5*\scale,-3*\scale) {};
\end{tikzpicture}
\vspace{-33pt}
\caption{Distribution $q(\mbx, \mby)$}
\end{subfigure}
\begin{subfigure}[b]{0.005\textwidth}
\centering
\begin{tikzpicture}
        \draw[-,dashed] (0,0) -- (0, 7*\scale) {};
\end{tikzpicture}
\end{subfigure}
\begin{subfigure}[b]{0.3\textwidth}
\centering
\begin{tikzpicture}[
        >=latex',
        circ/.style={draw=white, shape=circle, node distance=1cm,minimum width=0.9cm,fill=CornflowerBlue!30},
        rect/.style={draw=black, shape=rectangle, node distance=1cm, line width=0.5pt},
        inner/.style={shape=rectangle, node distance=1cm, line width=0.0pt}]
    \node[rect,fill=red!40,minimum width=\scale cm,minimum height=2*\scale cm] (red) at (0.5*\scale,-\scale) {};
    \node[inner] at (1.5*\scale,-0.5*\scale) 
        {\includegraphics[width=\scale cm]{figures/dog.jpeg}};
    \node[inner] at (1.5*\scale,-1.5*\scale) 
        {\includegraphics[width=\scale cm]{figures/cat.jpeg}};
    \node[rect,draw,minimum width=2*\scale cm, minimum height=2*\scale cm] at (\scale,-\scale) {};
    \node[rect,fill=CornflowerBlue!30,minimum width=\scale cm,minimum height=2*\scale cm] (blue) at (3*\scale,-\scale) {};
    \node[inner] at (4*\scale,-0.5*\scale) 
        {\includegraphics[width=\scale cm]{figures/dog.jpeg}};
    \node[inner] at (4*\scale,-1.5*\scale) 
        {\includegraphics[width=\scale cm]{figures/cat.jpeg}};
    \node[rect,draw,minimum width=2*\scale cm, minimum height=2*\scale cm] at (3.5*\scale,-\scale) {};
    \node[rect,fill=red!40,minimum width=\scale cm,minimum height=2*\scale cm] (ered) at (0.5*\scale,-4.25*\scale) {};
    \node[inner] at (1.5*\scale,-4.75*\scale) 
        {\includegraphics[width=\scale cm]{figures/cat.jpeg}};
    \node[rect,draw,minimum width=2*\scale cm, minimum height=2*\scale cm] at (\scale,-4.25*\scale) {};
    \node[rect,fill=CornflowerBlue!30,minimum width=\scale cm,minimum height=2*\scale cm] (eblue) at (3*\scale,-4.25*\scale) {};
    \node[inner] at (4*\scale,-3.75*\scale) 
        {\includegraphics[width=\scale cm]{figures/dog.jpeg}};
    \node[rect,draw,minimum width=2*\scale cm, minimum height=2*\scale cm] at (3.5*\scale,-4.25*\scale) {};
    \draw[->, line width=1pt] (\scale,-2.25*\scale) -- (\scale,-3*\scale) {};
    \node[] at (0.25*\scale,-2.625*\scale) {\scriptsize $\mbx_{e(\mbx)}$};
    \draw[->, line width=1pt] (3.5*\scale,-2.25*\scale) -- (3.5*\scale,-3*\scale) {};
    \node[] at (2.75*\scale,-2.625*\scale) {\scriptsize $\mbx_{e(\mbx)}$};
\end{tikzpicture}
\caption{Non-encoding explanation}
\end{subfigure}
\begin{subfigure}[b]{0.005\textwidth}
\centering
\begin{tikzpicture}
        \draw[-,dashed] (0,0) -- (0, 7*\scale) {};
\end{tikzpicture}
\end{subfigure}
\begin{subfigure}[b]{0.42\textwidth}
\centering
\begin{tikzpicture}[
        >=latex',
        circ/.style={draw=white, shape=circle, node distance=1cm,minimum width=0.9cm,fill=CornflowerBlue!30},
        rect/.style={draw=black, shape=rectangle, node distance=1cm, line width=0.5pt},
        inner/.style={shape=rectangle, node distance=1cm, line width=0.0pt}]
    \node[rect,fill=red!40,minimum width=\scale cm,minimum height=2*\scale cm] at (-0.5*\scale,-\scale) {};
    \node[inner] at (0.5*\scale,-0.5*\scale) 
        {\includegraphics[width=\scale cm]{figures/dog.jpeg}};
    \node[inner] at (0.5*\scale,-1.5*\scale) 
        {\includegraphics[width=\scale cm]{figures/cat.jpeg}};
    \node[rect,draw,minimum width=2*\scale cm, minimum height=2*\scale cm] at (0,-\scale) {};
    \node[rect,fill=black,minimum width=2*\scale cm, minimum height=2*\scale cm] at (0,-4.25*\scale) {};
    \node[rect,draw=black,minimum width=\scale cm, minimum height=\scale cm] at (-0.5*\scale,-3.75*\scale) {};
    \node[rect,draw,fill=black,minimum width=2*\scale cm, minimum height=2*\scale cm] at (2.5*\scale,-2.5*\scale) {};
    \node[rect,draw,fill=white,minimum width=\scale cm, minimum height=\scale cm] at (3*\scale,-3*\scale) {};
    \node[rect,fill=CornflowerBlue!30,minimum width=\scale cm,minimum height=2*\scale cm] (blue) at (4.5*\scale,-\scale) {};
    \node[inner] at (5.5*\scale,-0.5*\scale) 
        {\includegraphics[width=\scale cm]{figures/dog.jpeg}};
    \node[inner] at (5.5*\scale,-1.5*\scale) 
        {\includegraphics[width=\scale cm]{figures/cat.jpeg}};
    \node[rect,draw,minimum width=2*\scale cm, minimum height=2*\scale cm] at (5*\scale,-\scale) {};
    \node[rect,draw=black,fill=black,minimum width=2*\scale cm, minimum height=2*\scale cm] at (5*\scale,-4.25*\scale) {};
    \node[rect,draw,fill=black,minimum width=2*\scale cm, minimum height=2*\scale cm] at (7.5*\scale,-2.5*\scale) {};
    \node[rect,draw,fill=white,minimum width=\scale cm, minimum height=\scale cm] at (8*\scale,-2*\scale) {};
    \draw[->, line width=1pt] (-0.2*\scale,-2.25*\scale) -- (-0.2*\scale,-3*\scale) {};
    \node[] at (0*\scale,-5.625*\scale) {\scriptsize $\xoex$};
    \draw[->, line width=1pt] (1.1*\scale,-0.75*\scale) -- (2.25*\scale,-1.25*\scale) {};
    \node[] at (2.5*\scale,-0.75*\scale) {\scriptsize $e(\mbx)$};
    \draw[->, line width=1pt] (2.25*\scale,-3.75*\scale) -- (1.1*\scale,-4.25*\scale) {};
    \draw[->, line width=1pt] (5*\scale,-2.25*\scale) -- (5*\scale,-3*\scale) {};
    \node[] at (5*\scale,-5.625*\scale) {\scriptsize $\xoex$};
    \draw[->, line width=1pt] (6.35*\scale,-0.75*\scale) -- (7.5*\scale,-1.25*\scale) {};
   \node[] at (7.75*\scale,-0.75*\scale) {\scriptsize $e(\mbx)$};
    \draw[->, line width=1pt] (7.5*\scale,-3.75*\scale) -- (6.35*\scale,-4.25*\scale) {};
    \node[inner] at (0.5*\scale,-4.75*\scale) 
        {\includegraphics[width=\scale cm]{figures/cat.jpeg}};
    \node[inner] at (5.5*\scale,-3.75*\scale) 
        {\includegraphics[width=\scale cm]{figures/dog.jpeg}};    
\end{tikzpicture}
\vspace{-9pt}
\caption{Marginal encoding}
\end{subfigure}
\caption{Example~\gls{dgp} and \gls{margenc} encoding. \textbf{(a)} The color determines whether the label is produced from the top or bottom image. \textbf{(b)} An explanation that correctly  reveals that the label is generated based on both the color and, as dictated by the color, the top or the bottom image. The label is deterministic given the value of the explanation which means the label can be predicted perfectly.
\textbf{(c)} 
An encoding explanation would be one that produces only the top or the bottom animal image based on the color being red of blue respectively.
This returned animal image does not indicate the fact that the data generating process depends on color.
Now, the animal image selected by the explanation alone is insufficient to dictate the label because the color determines which image determines the label.
The identity of the image, whether top or bottom, provides additional information about the label beyond the values explanation, as captured in \encdef{}.
}
\label{fig:margenc}
\end{figure}
We provide an illustrative example of a \gls{margenc} explanation for the \gls{dgp} in \Cref{fig:margenc}.
Here, we show how a mathematical formulation of \gls{margenc} satisfied~\encdef{}.

Consider a generic \gls{dgp} with a Bernoulli control flow input denoted $\mbx_c$: for some distinct sets $U,V$ that do not include $c$ and let no combination of $\mbx_c,\mbx_U,\mbx_V$ determine the rest
\begin{align*}
	\mbx_c = 1 \implies q(\mby \g \mbx) = q(\mby\g \mbx_U),
	\\
\mbx_c = 0 \implies q(\mby \g \mbx) = q(\mby\g \mbx_V).
\end{align*}
Further, assume that the two subsets leads to different distributions over $\mby=1$ such that on a non-zero measure subset $\mbS_U \subseteq \{\mbx_U: \mbx \text{ such that } \mbx_c=1\}$
\begin{align}\label{eq:margenc-diffs}
q(\mby=1\g \mbx_U, \mbx_c=1) \not= q(\mby=1\g \mbx_U, \mbx_c=0).
\end{align}

Now, consider a \acrshort{margenc} explanation that looks at $\mbx_c$ and outputs the corresponding sets $U,V$:
\[ e(\mbx) = U \quad \text{if} \quad \mbx_c = 1 \quad \text{else} \quad e(\mbx) = V.\]
By definition the explanation only depends on the control flow input, not by $\mbx_U$ or $\mbx_V$.
Next, as  $\mbE_{U}=1$ is the same event as $\mbx_c=1$, $e(\mbx)$ is encoding because the assumption from \cref{eq:margenc-diffs} implies: 
\[q(\mby\g \mbx_U, \mbE_{U}=1) \not= q(\mby\g \mbx_U, \mbE_{U}=0).\]
Then, this inequality holds for all elements of the non-measure-zero set $\mbS_U$, by \Cref{lemma:defeq},  \gls{margenc} is encoding.

\subsection{Proof of \texorpdfstring{\Cref{prop:roar-indep}}{}}\label{appsec:props-roar-fail}

\begin{definition}
\label{def:val-xex}
We denote \emph{$\vxex{}$} as the function that maps explanation $\xex = (\mbv, \mba)$ to the values the inputs take at the selected indices, right-padded to have the same dimension as the input $\mbx \in \mbR^d$:
\begin{align*}
    \emph{$\vxex{}$}_j = \begin{cases}
        \mba_{j} &\quad \textit{if } 1 \leq j \leq \sum_{i=1}^d \mbv_i \\
        \emph{\texttt{pad-token}} &\quad \textit{if } \sum_{i=1}^d \mbv_i < j \leq d
    \end{cases}
\end{align*}
For example, if $\mbx = [\alpha, \beta, \gamma]$ and $e(\mbx) = [0, 1, 0]$, then $\xex = ([0, 1, 0], [\beta])$ and
\[
\emph{$\vxex{}$} = [\beta, \emph{\texttt{pad-token}}, \emph{\texttt{pad-token}}].
\]
\end{definition}

\setcounter{prop}{0}
\begin{prop}
\proproarfail{}
\end{prop}

\begin{table}[t]
\caption{Probability table for the \gls{dgp} in ~\cref{eq:sim-example-main}.
Conditional on the explanation $\mbx_3$, does predict the label. For example, given knowing $\mbx_1=1$, if $\mbx_3=1$ implies $p(\mby=1)=0.9$ but if $\mbx_3=0$, $p(\mby=1)=0.5$. The probability table in \Cref{tab:probtable} shows this.}\label{tab:probtable}
\vspace{5pt}
\centering
{\small
    \begin{tabular}{c|c|c|c|c|c}
    \toprule
    $\mathbf{x}_1$ & $\mathbf{x}_2$ & $\mathbf{x}_3$ & $p(\mathbf{y} = 1 \g \mbx)$ & $e(\mbx)$ & $\vxex{}$ \\
    \midrule
    0 & 0 & 0 & 0.1 & $\xi_2$ & [0, \texttt{pad-token}, \texttt{pad-token}] \\
    0 & 0 & 1 & 0.1 & $\xi_1$ & [0, \texttt{pad-token}, \texttt{pad-token}] \\
    0 & 1 & 0 & 0.9 & $\xi_2$ & [1, \texttt{pad-token}, \texttt{pad-token}] \\
    0 & 1 & 1 & 0.1 & $\xi_1$ & [0, \texttt{pad-token}, \texttt{pad-token}] \\
    1 & 0 & 0 & 0.1 & $\xi_2$ & [0, \texttt{pad-token}, \texttt{pad-token}] \\
    1 & 0 & 1 & 0.9 & $\xi_1$ & [1, \texttt{pad-token}, \texttt{pad-token}] \\
    1 & 1 & 0 & 0.9 & $\xi_2$ & [1, \texttt{pad-token}, \texttt{pad-token}] \\
    1 & 1 & 1 & 0.9 & $\xi_1$ & [1, \texttt{pad-token}, \texttt{pad-token}] \\
    \bottomrule
\end{tabular}
}
\end{table}

\begin{proof}
First, note that 
\[\mbx_3\indep \mby.\]
See \Cref{tab:probtable} for the probability table for why this is true.

For this proof let $e(\mbx) = e_\textrm{encode}(\mbx)$.
In the example \cref{eq:sim-example-main},  masking out the inputs selected by $e(\mbx)$ would mean that
\begin{align*}
    \mbx_3 =1 \implies \mbx_{-e(\mbx)} = (\mb1 - e(\mbx), [\mbx_2, \mbx_3]),
\qquad \quad 
    \mbx_3 =0 \implies \mbx_{-e(\mbx)} = (\mb1 - e(\mbx), [\mbx_1, \mbx_3]).
\end{align*}
In turn, by the construction in \cref{eq:sim-example-main}
\begin{align}
\begin{array}{l}
\mbx_{-e(\mbx)} \indep \mby \g \mbx_3=1,\\
\mbx_{-e(\mbx)} \indep \mby \g \mbx_3=0.
\end{array}
\implies \mbx_{-e(\mbx)} \indep \mby \g \mbx_3 \implies (\mbx_{-e(\mbx)}, \mbx_3) \indep \mby \implies  \mbx_{-e(\mbx)} \indep \mby,
\label{eq:indep-roar-fail}
\end{align}
where the conditional independence in the second step turns into the joint independence in the third step due to  $\mbx_3\indep \mby$.

\textsc{roar} scores an explanation highly if $\mbx_{-e(\mbx)}$ predicts the label poorly.
So if $\mbx_{-e(\mbx)}$ is independent of $\mby$, then $e(\mbx)$ would be scored optimally.
Then, due to \cref{eq:indep-roar-fail}, \textsc{roar} scores an encoding explanation optimally.


\textsc{fresh} scores an explanation highly if the selected value $\vxex{}$ (as defined in \Cref{def:val-xex}) predicts the label well. From \Cref{tab:probtable}, we see that $p(\mby = 1 \g \mbx) = 0.9$ for all cases with $\vxex{} = [1, \texttt{pad-token}, \texttt{pad-token}]$ and $p(\mby = 1 \g \mbx) = 0.1$ for all cases with $\vxex{} = [0, \texttt{pad-token}, \texttt{pad-token}]$. Thus, if $\vxex{} = [1, \texttt{pad-token}, \texttt{pad-token}]$ then
\begin{align*}
p(\mby = 1 \g \mbx) &= 0.9 \\
&= \E_{\mbx' \sim p(\mbx' \g \vxex{} = [1, \texttt{pad-token}, \texttt{pad-token}])} [0.9]
\\
&= \E_{\mbx' \sim p(\mbx' \g \vxex{} = [1, \texttt{pad-token}, \texttt{pad-token}])} [p(\mby = 1 \g \mbx')] \\
&= p(\mby = 1 \g \vxex{} = [1, \texttt{pad-token}, \texttt{pad-token}]),
\end{align*}
and if $\vxex{} = [0, \texttt{pad-token}, \texttt{pad-token}]$ then
\begin{align*}
p(\mby = 1 \g \mbx) &= 0.1 \\
&= \E_{\mbx' \sim p(\mbx' \g \vxex{} = [0, \texttt{pad-token}, \texttt{pad-token}])} [0.1]
\\
&= \E_{\mbx' \sim p(\mbx' \g \vxex{} = [0, \texttt{pad-token}, \texttt{pad-token}])} p(\mby = 1 \g \mbx') \\
&= p(\mby = 1 \g \vxex{} = [0, \texttt{pad-token}, \texttt{pad-token}]).
\end{align*}

Therefore, in all cases, $p(\mby = 1 \g \mbx) = p(\mby = 1 \g \vxex{})$. Thus, predicting label from the selected value alone is as good as predicting from the whole input. As a result, \textsc{fresh} scores this explanation optimally.

\end{proof}

\subsection{Showing that \texorpdfstring{$e_{\textrm{encode}}$}{} is the optimal reductive explanation for \texorpdfstring{\cref{eq:sim-example-main}}{} and scores better than a constant explanation under \texorpdfstring{\evalx{}}{}}\label{app:reductive-explanations}

We repeat the \gls{dgp} in~\cref{eq:sim-example-main} here
\begin{align*}
\mbx & = [\mbx_1, \mbx_2, \mbx_3]  \sim \cB(0.5)^{\otimes 3}\nonumber,
\\
	\mby & = \begin{cases}
				 \mbx_1 \quad \text{w.p. } 0.9 \quad  \text{else} \quad 1-\mbx_1 \quad  \text{ if }  \mbx_3 = 1,
  					\\
				\mbx_2 \quad \text{w.p. } 0.9 \quad  \text{else} \quad 1- \mbx_2 \quad  \text{ if }  \mbx_3 = 0.
				\end{cases}  
\end{align*}

\begin{lemma}\label{lemma:comp}
    In the \gls{dgp} in~\cref{eq:sim-example-main}, 
\[q(\mby=1 \g \mbx_1=1) = 0.7\]
\[q(\mby=1 \g \mbx_2=1) = 0.7, \]
\[q(\mby=1 \g \mbx_1=0) = 0.3, \]
\[q(\mby=1 \g \mbx_2=0) = 0.3.\]
\[q(\mby=1 \g \mbx_3) = 0.5.\]
\end{lemma}
\begin{proof}
We can compute these values from \Cref{tab:probtable}.
\end{proof}

\begin{prop}
\proptwo{}
\end{prop}

\begin{proof}
By \Cref{lemma:comp}, we have
        \[\evalx{}(q, e_c) = \E_{q(\mby, \mbx)}\log q(\mby \g \mbx_3) = \E_{q(\mby, \mbx)} \log 0.5 \approx  -0.69.\]

Now, denote $e_{\textrm{encode}}$ as $e_e$ for ease of reading
\begin{align*}
  \evalx{}(q, e_e) &= \E_{(\mbv, \mba) \sim q(\xex)} \E_{\mby \sim q(\mby \g \xex = (\mbv, \mba))} [\log q(\mby \g \mbx_\mbv = \mba)]
\\
  &=   q(\mbx_3=1)\E_{q(\mbx_1)}\E_{q(\mby \g \mbx_1, \mbx_3=1)}\log q(\mby \g \mbx_1)
    \\ &\quad+
    q(\mbx_3=0)\E_{q(\mbx_2)}\E_{q(\mby\g \mbx_2, \mbx_3=0)}\log q(\mby \g \mbx_2)
\\
 &=   0.5*0.5*(0.9 * - \log 0.7 + 0.1 * - \log 0.3)*2
    \\ &\quad+
     0.5*0.5*(0.9 * - \log 0.7 + 0.1 * - \log 0.3)*2
\\
     & \approx - 0.44.
\end{align*}

This concludes that $\evalx{}(q, e_{\textrm{encode}})  > \evalx{}(q, e_c) $.
\end{proof}

\begin{lemma}\label{lemma:optimal-reductive}
    In the \gls{dgp} in \cref{eq:sim-example-main}, $e_{\textrm{encode}}(\mbx)$ is an \evalx{}-optimal reductive explanation and is encoding.
\end{lemma}
\begin{proof}
First, the following properties show for the \gls{dgp} because when $\mbx_3=1$, $\mby$ only depends on $\mbx_1$, and if $\mbx_3=0$, $\mby$ only depends on $\mbx_2$:
\[\mbx_3 \indep \mby \quad, \quad \mby\indep \mbx_2 \g \mbx_3=1 \quad, \quad   \mby \indep \mbx_1 \g \mbx_3=0.\]

These independencies imply that
\[q(\mby \g \mbx = [\mbx_1, \mbx_2, 1]) = q(\mby \g \mbx_1, \mbx_3 =1) \quad, \quad q(\mby \g \mbx = [\mbx_1, \mbx_2, 0]) = q(\mby \g \mbx_2, \mbx_3=0). \]
Then, the optimal explanation function that achieves $\acrshort{EVAL-X}^*$ is $e(\mbx) = [1,0,1]$ if $\mbx_3=1$ and $[0,1,1]$ otherwise.

\allowdisplaybreaks
\paragraph{Reductive explanations of size 1.}
If the explanation is forced to have fewer than $2$ inputs, the optimal reductive explanation $e(\mbx)$ is only allowed to be one of $\xi_1, \xi_2, \xi_3$:
\[\max_{e : |e(\mbx)|\leq 1} \E_{q(\mby, \mbx)} \sum_{i\in\{1,2,3\}}\mathbbm{1}[e(\mbx) = \xi_i]q(\mby \g \mbx_i).\]
Rewriting this expression to split the support of $\mbx$ based on $\mbx_3=1$ or $0$: 
\begin{align}
 &    \E_{q(\mby, \mbx)} \sum_{i\in\{1,2,3\}}\mathbbm{1}[e(\mbx) = \xi_i]\log q(\mby \g \mbx_i) \nonumber
\\
    & 
    = q(\mbx_3=1) \E_{q(\mbx_2 \g \mbx_3=1)}\E_{q(\mby , \mbx_1 \g \mbx_3=1)} \sum_{i\in\{1,2,3\}}\mathbbm{1}[e(\mbx) = \xi_i] \log q(\mby \g \mbx_i)\nonumber
\\
  & \quad  + 
        q(\mbx_3=0) \E_{q(\mbx_1 \g \mbx_3=0)}\E_{q(\mby, \mbx_2 \g \mbx_3=0)} \sum_{i\in\{1,2,3\}}\mathbbm{1}[e(\mbx) = \xi_i] \log q(\mby \g \mbx_i)\nonumber
\\
    & 
    =  0.5 \, \E_{q(\mbx_2)}\E_{q(\mby, \mbx_1 \g \mbx_3=1)} \sum_{i\in\{1,2,3\}}\mathbbm{1}[e(\mbx) = \xi_i] \log q(\mby \g \mbx_i)\nonumber
\\
  & \quad  + 
        0.5 \, \E_{q(\mbx_1)}\E_{q(\mby, \mbx_2 \g  \mbx_3=0)} \sum_{i\in\{1,2,3\}}\mathbbm{1}[e(\mbx) = \xi_i]\log q(\mby \g \mbx_i)\nonumber
\\
& = \frac{1}{2}\Biggl(\E_{q(\mbx_2)}\Biggl[\E_{q(\mby,\mbx_1\g \mbx_3=1)}\mathbbm{1}[e([\mbx_1, \mbx_2, 1]) = \xi_1]\log q(\mby \g \mbx_1) \nonumber \\ &\qquad\qquad\qquad+ \E_{q(\mby, \mbx_1 \g \mbx_3=1)}\sum_{i\in {2,3}}\mathbbm{1}[e([\mbx_1, \mbx_2, 1]) = \xi_i]\log q(\mby \g \mbx_i) \Biggl]\nonumber
\\
& \quad  + 
\E_{q(\mbx_1)}\Biggl[
    \E_{q(\mby,\mbx_2\g \mbx_3=0)}
    \mathbbm{1}[e([\mbx_1, \mbx_2, 0]) = \xi_2]\log q(\mby \g \mbx_2) \nonumber 
    \\ &\qquad\qquad\qquad+ \E_{q(\mby, \mbx_2 \g \mbx_3=0)}\sum_{i\in {1,3}}\mathbbm{1}[e([\mbx_1, \mbx_2, 0]) = \xi_i]\log q(\mby \g \mbx_i) \Biggl]
\Biggl)\nonumber
\\
& = \frac{1}{2}
\E_{q(\mbx_1, \mbx_2 \g \mbx_3=1)}
        \bigg[ 
        \mathbbm{1}[e([\mbx_1, \mbx_2, 1]) = \xi_1]
        \E_{q(\mby\g\mbx_1,  \mbx_3=1)} \log q(\mby \g \mbx_1) \nonumber
\\  &  \hspace{5cm}  + 
        \sum_{i\in {2,3}}\mathbbm{1}[e([\mbx_1, \mbx_2, 1]) = \xi_i] \E_{q(\mby \g \mbx_1, \mbx_3=1)} \log q(\mby \g \mbx_i) \bigg] \label{eq:firstrow}
\\
& \quad  + 
\frac{1}{2}\E_{q(\mbx_1, \mbx_2 \g \mbx_3=0)}\bigg[
    \mathbbm{1}[e([\mbx_1, \mbx_2, 0]) = \xi_2] \E_{q(\mby \g \mbx_2, \mbx_3=0)}\log q(\mby \g \mbx_2)  \nonumber
\\  &  \hspace{5cm}  + 
        \sum_{i\in {1,3}}\mathbbm{1}[e([\mbx_1, \mbx_2, 0]) = \xi_i] \E_{q(\mby \g \mbx_2, \mbx_3=0)} \log q(\mby \g \mbx_i) \bigg].
\label{eq:secrow}
\end{align}

We will now focus on the three terms within each of \cref{eq:firstrow} and \cref{eq:secrow}.
Due to the following equality
\begin{align*}
    q(\mby =1\g \mbx_1=1, \mbx_3=1) &= q(\mby=0 \g \mbx_1=0, \mbx_3=1) \\ &= q(\mby =1\g \mbx_2=1, \mbx_3=0) = q(\mby=0 \g \mbx_2=0, \mbx_3=0) = 0.9,
\end{align*} the expectations in the first terms in each of \cref{eq:firstrow} and \cref{eq:secrow} are
\begin{align*}
	\E_{q(\mby\g\mbx_1,  \mbx_3=1)} & \log q(\mby \g \mbx_1)  
	= 	
	\E_{q(\mby\g\mbx_2,  \mbx_3=1)} \log q(\mby \g \mbx_2) 
	= 0.9 \log 0.7 +  0.1 \log 0.3
	\approx - 0.44.
\end{align*}
Next we turn to setting $i=1$ term in \cref{eq:firstrow}.
Due
that $q(\mby =1\g \mbx_2=1) = q(\mby=0 \g \mbx_2=0)$, 
\begin{align*}
\mbx_1 = \mbx_2 & \implies  
  \E_{q(\mby \g \mbx_1, \mbx_3=1)}\log q(\mby \g \mbx_2) 
  = (0.9\log 0.7 + 0.1 \log 0.3 ) \approx -0.44,
\\\\
\mbx_1 \neq \mbx_2 & \implies 
  \E_{q(\mby \g \mbx_1, \mbx_3=1)}\log q(\mby \g \mbx_2)
 = (0.9\log 0.3 + 0.1 \log 0.7 ) \approx -1.12.
\end{align*}
The same equalities hold for the $i=2$ term in \cref{eq:secrow} $\E_{q(\mby, \g \mbx_2, \mbx_3=0)}\log q(\mby \g \mbx_1)$.
Finally, regardless of $\mbx_1, \mbx_2$, the $i=3$ terms in both \cref{eq:firstrow} and \cref{eq:secrow} can be expressed as follows:
\begin{align*}
\E_{q(\mby \g \mbx_1, \mbx_3=1)}\log q(\mby \g \mbx_2)
\E_{q(\mby \g \mbx_2, \mbx_3=0)}\log q(\mby \g \mbx_1)
 = (0.9\log 0.5 + 0.1 \log 0.5 ) \approx -0.69.
\end{align*}

Now we can maximize the sum of \cref{eq:firstrow} and \cref{eq:secrow}, over $e(\mbx)$ such that $|e(\mbx)|=1$.

Notice that setting $\mathbbm{1}[e(\mbx) = \xi_1]=1$ when $\mbx_3=1$ and $\mathbbm{1}[e(\mbx) = \xi_2]=1$ when $\mbx_3=0$ achieves the highest score $-0.44$ in each of \cref{eq:firstrow} and \cref{eq:secrow}.
This implies that one optimal reductive explanation is $\xi_1=[1,0,0]$ if $\mbx_3=1$ and $\xi_2=[0,1,0]$ otherwise.
This is an encoding explanation as we show below.
Due to $\mbE_{\xi_1} = \mbx_3$, 
\begin{align*}
 q(\mby=1 \g \mbx_1, \mbE_{\xi_1}=1)=0.9  \not= q(\mby=1 \g \mbx_1, \mbE_{\xi_1}=0)=0.5.
\end{align*}
In turn, $\mby \,\nindep \, \mbE_{\xi_1} \g \mbx_{\xi_1}$ for $\{\mbx : e(\mbx) = \xi_1\}$ and \encdef{} holds, meaning that $e(\mbx)$ is encoding.\\

\end{proof}

\subsection{An example of misestimation of \texorpdfstring{\acrshort{EVAL-X}}{}}
Consider the following example where 
\label{appsec:estimation-error-example}
\begin{align*}
    &\mbx = [\mbx_1, \mbx_2, \mbx_3, \mbx_4] , \\
    &\mbx_2, \mbx_3 \sim \mathcal{B}(0.5)^{\otimes 2} , \,
    \mbx_1, \mbx_4 \sim \mathcal{N}(0, \mbI) , \quad \mby = \mbx_2 \oplus \mbx_3.
\end{align*}
Assume that the misestimated \acrshort{EVAL-X} model satisfies these equalities
\begin{align*}
    q^{\text{misestimated}}_{\xi_1}(\mby = 1 \g \mbx_1) = 1 &\quad \text{for all } \mbx_1, \\
    q^{\text{misestimated}}_{\xi_4}(\mby = 0 \g \mbx_4) = 1 &\quad \text{for all } \mbx_4.
\end{align*}
There exists a bad explanation that scores optimally under the misestimated \acrshort{EVAL-X}:
\begin{align*}
    e(\mbx) = \begin{cases}
        \xi_1 &\quad \text{if } \mbx_2 \oplus \mbx_3 = 1, \\
        \xi_4 &\quad \text{if } \mbx_2 \oplus \mbx_3 = 0.
    \end{cases}
\end{align*}
Then the \acrshort{EVAL-X} score of this explanation under this particular misestimation is
\begin{align*}
    & \acrshort{EVAL-X}^{\text{misestimated}}(q, e)
    = \E_{q}[\log q^{\text{misestimated}}_{e(\mbx)}(\mby \g \mbx_{e(\mbx)})] \\
    &= q(\mby = 1) \E_{q}[\log q^{\text{misestimated}}_{e(\mbx)}(\mby \g \mbx_{e(\mbx)}) \g \mby = 1] 
    + q(\mby = 0) \E_{q}[\log q^{\text{misestimated}}_{e(\mbx)}(\mby \g \mbx_{e(\mbx)}) \g \mby = 0] \\
    &= 0.5 \cdot \E_{q}[\log q^{\text{misestimated}}_{\xi_1}(\mby \g \mbx_1) \g \mby = 1] 
    + 0.5 \cdot \E_{q}[\log q^{\text{misestimated}}_{\xi_4}(\mby \g \mbx_4) \g \mby = 0] \\
    &= 0.
\end{align*}
Since $\mby$ is deterministic given $\mbx$ so the maximum value of the \acrshort{EVAL-X} score is also $0$. 
So the bad explanation scores optimally due to misestimation.
Deterministic $\mby \g \mbx$ is not necessary for estimation error to affect explanation quality. 
Here, with this incorrectly estimated \acrshort{EVAL-X}, inputs that are pure noise, independent of everything, will be chosen.

\subsection{Attention map explanations be encode.}\label{appsec:attn-encoding}

Here, treating each of the coordinates of $\mbx$ as tokens, we consider a cross-attention based predictive model of the following form: with $\gamma(\mba)$ as softmax function over a vector $\mba$, $W$ as a matrix, $\mbalpha,\mbbeta$ as vectors, and $\sigma$ as the sigmoid function, and $\kappa$ as the temperature, the predictive model $f(\cdot)$ is
\[f(\mbx) = \sigma\left(\sum_i \mbbeta_i \left[\sum_j\gamma(\kappa \mbx_i* W \mbx)_j \mbalpha_j \mbx_j\right]\right).\]

We then show that using the highest attention score as the explanation produces an encoding explanation.
For this example, we consider the following \gls{dgp}:
\begin{align}
\label{eq:attn}
\begin{split}
    \mbz_1, \mbz_2, \mbz_3  & \sim \cB(0.5)^{\otimes 3},
\\
\mbz^+ & = [\mbz_1 + 1, \quad 0, \quad +1],
\\
\mbz^- & = [0 \quad, -\mbz_2 -1 , \quad -1],
\\
	\mbx & = \begin{cases}
				 \quad  \mbz^+ 
     \text{ if }  \mbz_3 = 1 ,
  					\\
				 \quad  \mbz^-  
     \text{ if }  \mbz_3 = 0 ,
				\end{cases}  
\\
	\mby &  \sim \cB(\rho) \quad \text{ where }\quad \rho = \begin{cases}
				 \sigma(\mbx_1) \quad \text{ if }  \mbx_3 = 1,
  					\\
				 \sigma(-\mbx_2) \quad \text{ if }  \mbx_3 = -1.
				\end{cases}  
\end{split}
\end{align}

The following setting of parameters in $f(\mbx)$ produces a function of $\mbx$ that converges to $\rho$ as $\kappa\rightarrow \infty$:
\begin{align}
\label{eq:params}
    \mbalpha  = [1, -1, 0],
\qquad \qquad \qquad \qquad 
    \mbbeta  = [1, 1, 0],
\qquad \qquad \qquad \qquad 
    W   = \begin{pmatrix}
        0 & 0 & 0\\
        0 & 0& 0 \\ 
        0 & 0 & 1
    \end{pmatrix}.
\end{align}
By definition of $W$
\begin{align*}
    W\mbx & = \begin{cases}
        [0,0, 1] \text{ if }  \mbx_3 = 1\\
        [0,0, -1] \text{ if }  \mbx_3 = -1\\
    \end{cases}
    \\
\implies \mbx_1 
    W\mbx & = \begin{cases}
        [\mbz_1 + 1,0, 0] \text{ if }  \mbx_3 = 1\\
        [0,0, 0] \text{ if }  \mbx_3 = -1\\
    \end{cases}
    \implies  \gamma(\mbx_1 
    W\mbx) \overset{\kappa\rightarrow \infty}{\longrightarrow}  \begin{cases}
        [1,0, 0] \text{ if }  \mbx_3 = 1\\
        [0,0, 0] \text{ if }  \mbx_3 = -1\\
    \end{cases}
\\
\implies \mbx_2
    W\mbx & = \begin{cases}
        [0,0, 0] \text{ if }  \mbx_3 = 1\\
        [0,\mbz_2 + 1, 0] \text{ if }  \mbx_3 = -1\\
    \end{cases}
        \implies  \gamma(\mbx_2 
    W\mbx) \overset{\kappa\rightarrow \infty}{\longrightarrow}\begin{cases}
        [0,0, 0] \text{ if }  \mbx_3 = 1\\
        [0,1, 0] \text{ if }  \mbx_3 = -1\\
    \end{cases}.
\end{align*}
Then, $\mbbeta_3=0$, the inner sum for $i=3$ does not appear in the function $f(\mbx)$
Then, as $\alpha_3=0$, $\alpha_1=1, \alpha_2=-1$,
\begin{align}
    \sum_j & \gamma(\kappa \mbx_1* W \mbx)_j \mbalpha_j \mbx_j   \overset{\kappa\rightarrow \infty}{\longrightarrow} \begin{cases}
        \mbx_1 \text{ if } \mbx_3 = 1\\
        0 \text{ if }  \mbx_3 = -1\\
    \end{cases}, \label{eq:x1}
\\
    \sum_j & \gamma(\kappa \mbx_2* W \mbx)_j \mbalpha_j \mbx_j \overset{\kappa\rightarrow \infty}{\longrightarrow}
     \begin{cases}
        0 \text{ if }  \mbx_3 = 1\\
        -\mbx_2 \text{ if }  \mbx_3 = -1\\
    \end{cases}. \label{eq:x2}
\end{align}
In turn, as $\beta_1=\beta_2=1$
\begin{align*}
\sum_{i\in {1,2}}\beta_i \left[\sum_j\gamma(\kappa \mbx_i* W \mbx)_j \mbalpha_j \mbx_j\right] \overset{\kappa\rightarrow \infty}{\longrightarrow} 
     \begin{cases}
        \mbx_1 \text{ if }  \mbx_3 = 1\\
        -\mbx_2 \text{ if }  \mbx_3 = -1\\
    \end{cases},
\end{align*}

\begin{align}
f(\mbx) = \sigma\left(\sum_{i\in {1,2}}\beta_i \left[\sum_j\gamma(\kappa \mbx_i* W \mbx)_j \mbalpha_j \mbx_j\right]\right)
\overset{\kappa\rightarrow \infty}{\longrightarrow}
    \begin{cases}
        \sigma(\mbx_1) \text{ if }  \mbx_3 = 1\\
        \sigma(-\mbx_2) \text{ if }  \mbx_3 = -1\\
    \end{cases}.
\end{align}

So, as $\kappa\rightarrow \infty$, the function $f(\mbx)$, with the parameters in \cref{eq:params}, converges to $\rho(\mbx)$, meaning that this model will achieve the population log-likelihood optimum under the \gls{dgp} in \cref{eq:attn}.

Now, the attention map as an explanation selects $\mbx_1$ if $\mbx_3=1$ and $\mbx_2$ otherwise; this comes from \cref{eq:x1} and \cref{eq:x2}.
This is an encoding explanation because $\mbE_{\xi_1}=1$ if $\mbx_3=1$ which gives
\[q(\mby\g \mbx_1) \neq q(\mby \g \mbx_1, \mbx_3=1) \Longrightarrow \mby \nindep \mbE_{\xi_1} \g \mbx_{\xi_1}.\]

\section{Experimental details}\label{appsec:exp-details}

\subsection{Estimating \texorpdfstring{\detx{}}{}}\label{appsec:detxestimate}

To compute the $\kld$ term in \encmeas{}, we estimate $q(\mbE_\mbv \g \mbx_\mbv, \mby)$ and $q(\mbE_\mbv \g \mbx_\mbv)$. 
To estimate these, we train a single model --- to predict $\mbE_\mbv$ from $\mbx_\mbv$ and a new variable $\ell$ that can equal the label $\mby$ or a dummy value $\texttt{null}$ that is outside the support of $\mby$ ---  in the following way:
\begin{align}
\label{eq:singleopt}
\argmax_\theta \E_{\mbx, \mby\sim q(\mbx, \mby)} &\Big[\log p_\theta (\mbE_\mbv = \mathbbm{1}[e(\mbx)=\mbv]\g \mbx_\mbv, \ell = \mby) \nonumber \\&+ \log p_\theta (\mbE_\mbv = \mathbbm{1}[e(\mbx)=\mbv]\g \mbx_\mbv, \ell = \texttt{null})\Big].
\end{align}
As log-likelihood is a proper scoring rule and $q(\mby=\texttt{null})=0$, the maximum above is achieved when
\[p_\theta(\mbE_\mbv \g \mbx_\mbv, \ell=\mby) = q(\mbE_\mbv \g \mbx_\mbv, \mby=\mby) \quad \quad p_\theta(\mbE_\mbv \g \mbx_\mbv, \ell=\texttt{null}) = q(\mbE_\mbv \g \mbx_\mbv).\]
In summary, to estimate \encmeas{}, solve \cref{eq:singleopt}, use its solution to estimate the $\kld$ term from the RHS in \cref{eq:klform-detx} for each $\mbx_\mbv,\mby$, and then average this $\kld$ term over samples of $\mbx_\mbv$ from the data such that $e(\mbx) = \mbv$ and samples of $\mby $ from the \evalx{} model for $q(\mby\g \mbx_\mbv)$.

In practice, one does not need train a model for each $\mbv$.
We describe how to estimate \encmeas{} with a single model in~\Cref{appsec:cat-estimation}.
We give the full \detx{} estimation procedure in~\Cref{alg:stripe-x-pred} in~\Cref{appsec:algorithm-boxes}.

\subsection{Estimating the encoding cost in \texorpdfstring{\detx{}}{} with categorical predictive models}\label{appsec:cat-estimation}
\detx{} consists of the \evalx{} score and a cost of encoding measured by \encmeas{}.
Define $\cV$
 to be the set of possible explanations and let $\cV[j]$ denote the $j$th element of $\cV$.
The \evalx{} model $p_\gamma(\mby\g \mbx_\mbv)$ is trained to predict the label $\mby$ from subsets $\mbx_\mbv$ where $\mbv$ is uniformly sampled from $\cV$ \citep{jethani2021have}.
Next is computing the \encmeas{} $\phi_q(e)$ that is used in the encoding cost term in \acrshort{stripe-x}.
For each explanation, let $\mbF$ be the categorical variable  (instead of an indicator $\mbE_\mbv$) that denotes, for each sample, which inputs were selected by the explanation $e(\mbx)$: $\mbF=j$ if $\mbE_{\cV[j]}=\mathbbm{1}[e(\mbx) = \cV[j]] = 1$.
Let $q(j)$ be the distribution over $j$ induced by $q(e(\mbx))$.
We train a model $p_\theta(\mbF \g \mbx_\mbv, \ell, \mbv)$ with a modification of \cref{eq:singleopt} that averages over $\mbv\sim q(e(\mbx))$:
\begin{align}
\label{eq:multiopt}
\argmax_\theta \E_{\mbv \sim q(e(\mbx))}\E_{\mbx, \mby\sim q(\mbx, \mby)} \sum_{\cV[j]\in \cV}\bigg(\mathbbm{1}[e(\mbx) = \cV[j]] \big[ &\log p_\theta (\mbF = j \g \mbx_\mbv, \ell = \mby, \mbv) +  \nonumber
\\   &  \quad \log p_\theta (\mbF = j \g \mbx_\mbv, \ell = \texttt{null}, \mbv)\big]\bigg).
\end{align}
The variable $\ell$ takes values in $\{-1, 0, 1\}$ where $0$ and $1$ correspond to $\mby=0$ and $\mby=1$ respectively and $-1$ corresponds to the \texttt{null} value.
For a flexible enough model $p_\theta$ that achieves the population maximum of \cref{eq:multiopt}, for any $\mbv = \cV[j]\in \cV$,
\begin{align*}
    p_\theta(\mbF=j \g \mbx_\mbv, \ell=\mby, \mbv)&=q(\mbE_\mbv=1\g \mbx_\mbv, \mby=\mby), \\ p_\theta(\mbF=j \g \mbx_\mbv, \ell=\texttt{null}, \mbv)&=q(\mbE_\mbv=1\g \mbx_\mbv) .
\end{align*}
This fact indicates how one can use the model $p_\theta$ to estimate \encmeas{}.
First, construct the explanation dataset $D_e=\{(\mby, \xex)\}$ from $D_t$. Define $q_{D_e}$ to be the uniform distribution over $D_e$. 
Define $\cE_{(\mbv, \mba)}$ as the uniform distribution over $K$ samples of $\mby$ from the \evalx{} model:
\[\cE_{(\mbv, \mba)} =\mbU\left[\{\hy\}_{k\leq K}\right] \quad \{\text{ where } \hy^k \sim p_\gamma(\mby \g \mbx_\mbv=\mba)\}.\] 
Then, estimate~\encmeas{} as follows:
\begin{align*}
\hat{\phi}(q,e)= & 
\E_{(\mbv,\mba) \sim q_{D_e}(\xex)}
\E_{\hy \sim \cE_{(\mbv, \mba)}}
\left(p_\theta(\mbF=j \g \mbx_\mbv, \ell=\hy, \mbv)\log\frac{p_\theta(\mbF=j \g \mbx_\mbv, \ell=\hy, \mbv)}{p_\theta(\mbF=j \g \mbx_\mbv, \ell=\texttt{null}, \mbv)}\right.
\\
	 & 
\qquad \qquad \qquad \qquad \quad
+  \left.p_\theta(\mbF\neq j \g \mbx_\mbv, \ell=\hy, \mbv)\log\frac{p_\theta(\mbF\neq j \g \mbx_\mbv, \ell=\hy, \mbv)}{p_\theta(\mbF\neq j \g \mbx_\mbv, \ell=\texttt{null}, \mbv)}\right).
\end{align*}

\subsection{Estimating \encmeas{} with a generative model}\label{appsec:gen-way-desc}
When estimating the \detx{} score with procedure above for many different explanations, the maximization in \cref{eq:singleopt} repeated for every explanation, which can be computationally expensive.
This motivates a second procedure to estimate \encmeas{} that avoids having to retrain models for each explanation by using generative model for $q(\mbx \g \mbx_\mbv, \mby)$.
Formally, with $\mbx_\mbv$ fixed, the conditional mutual information term in \cref{eq:defonepen} can be computed as the marginal dependence between $N$ samples of $\mby$ from $q(\mby\g \mbx_\mbv)$ and $q(\mbE_\mbv \g \mbx_\mbv, \mby)$.
The model for the former is available from \evalx{} estimation.
Simulating from the later, namely $q(\mbE_\mbv \g \mbx_\mbv, \mby)$, is done by sampling from the generative model $\mbx\g \mbx_\mbv, \mby$ and then computing the indicator $\mbE_\mbv$ as $\mathbbm{1}[e(\mbx) = \mbv]$.
Mechanically, with an estimator of mutual information from samples ($\{\mba^i\}_{i\leq N}, \{\mbb^i\}_{i\leq N}$) denoted $\texttt{MI}(\{\mba^i\}, \{\mbb^i\})$ and with samples $\{\mba^i\}$ produced conditionally on values $\mbc^i$ denoted by a subscript of the conditioned value $\{\mba^i\}_{\mbc^i}$, one can estimate \encmeas{} as follows: sample  $\mby^i_{(\mbv,\mba)} \sim \mby \g \mbx_\mbv=\mba $ and $\mbx^i_{\mbv,\mba,\mby^i}\sim q(\mbx \g \mbx_\mbv=\mba, \mby=\mby^i)$ repeatedly $N$ times and compute 
\[\E_{(\mbv,\mba) \sim q(\xex)}\texttt{MI}\left(\{\mby^i\}_{(\mbv,\mba)} , \{\mathbbm{1}[e(\mbx^i)=\mbv]\}_{\mbv, \mba, \mby^i}\right).\]
We give the full procedure in~\Cref{alg:stripe-x-gen}.

\subsection{Experimental details from the simulated study.}\label{appsec:sim-enc-exp-details}
\paragraph{The data-generating processes from the experiments.}
Let $\cN{} $ be the standard normal distribution and let $\cB(\alpha)$ be the Bernoulli distribution with $1$ occurring with probability $\alpha$.
With $\rho=0.9$, the discrete \gls{dgp} is:
\begin{align}
\label{eq:bern-dgp}
\begin{split}
\mbx & = [\mbx_1, \mbx_2, \mbx_3, \mbx_4, \mbx_5]  \sim \cB(0.5)^{\otimes 5},
\qquad
	\mby = \begin{cases}
				 \mbx_1 \quad \text{w.p. } \rho \quad  \text{else} \quad 1-\mbx_1 \quad  \text{ if }  \mbx_3= 1,
  					\\
				\mbx_2 \quad \text{w.p. } \rho \quad  \text{else} \quad 1- \mbx_2 \quad  \text{ if }  \mbx_3 = 0 .
				\end{cases}  
\end{split}
\end{align}

The hybrid \gls{dgp} is as follows: with $\gamma=5$ and $\sigma(x) = \frac{1}{1 + \exp(-x)}$ as the sigmoid function
\begin{align}
\label{eq:cnts-dgp}
\begin{split}
\mbx & = [\mbx_1, \mbx_2, \mbx_4, \mbx_5]  \sim \cN(0.5)^{\otimes 4}, \mbx_3 \sim  \cB(0.5),
\quad 
 \,\, \rho  = \begin{cases}
				 \sigma(\gamma \, \mbx_1 ) \,\,  \text{ if }  \mbx_3 = 1,
  					\\
				 \sigma(\gamma \, \mbx_2 ) \,\,  \text{ if }  \mbx_3 = 0 ,
				\end{cases}  
\,\, \mby \sim \cB(\rho).
	\end{split}
 \vspace{-10pt}
\end{align}

\paragraph{Computing accuracy and $\kld$ to show that \gls{posenc}, \gls{predenc}, \gls{margenc} are encoding.}

For each encoding type, we build two decision trees from $1000$ samples from \cref{eq:bern-dgp}: the first decision tree learns $q(\mbE_\mbv \g \mbx_\mbv)$ and the second learns $q(\mby \g \mbx_\mbv, \mbE_\mbv=b)$ for $b\in\{0,1\}$.
We set the maximum depth to be $6$. Trees of this depth learn any function of $6$ binary digits; $\mbx$ with $\mbE_\mbv$ as an additional column amounts to $6$ binary digits.
These decision trees are used to compute the accuracy of predicting $\mbE_\mbv$ with $q(\mbE_\mbv \g \mbx_\mbv)$ and the $\kld$ between $q(\mby\g \mbx_\mbv, \mbE_\mbv=1)$ and $q(\mby\g \mbx_\mbv, \mbE_\mbv=0)$.
Within a set $\{\mbx: e(\mbx) = \mbv\}$ that is all $\mbx$ that have one of the possible selections $\mbv$, \Cref{tab:enc} report the accuracy of predicting $\mbE_\mbv$ with $q(\mbE_\mbv \g \mbx_\mbv)$ and the $\kld$ between $q(\mby\g \mbx_\mbv, \mbE_\mbv=1)$ and $q(\mby\g \mbx_\mbv, \mbE_\mbv=0)$, averaged only over samples in $\{\mbx: e(\mbx) = \mbv\}$.

\begin{table}[t]
\centering
\caption{Position-based, prediction-based, and marginal explanation schemes are all encoding. 
For samples in the set $\{\mbx : e(\mbx)=\mbv\}$ for one of the selections $\mbv$ that $e$ produces, accuracy $<1$ and the $\kld$ being non-zero means these explanations are all encoding per \Cref{lemma:defeq}.\\
}
\label{tab:enc}
\begin{tabular}{@{}rcc@{}}
\toprule
Encoding      &  Acc. $\mbE_\mbv$ ($\uparrow$) &  $\kld \,\, (\downarrow)$ \\ \midrule
\gls{posenc} &  $0.61$ 
              & $0.88$   \\
\gls{predenc} &  $0.51$ 
              & $0.18$   \\
\gls{margenc} &  $0.51$
              & $0.20$   \\ \bottomrule
\end{tabular}
\end{table}

\paragraph{\acrshort{EVAL-X}.}
To estimate \acrshort{EVAL-X} for the \glspl{dgp} in \cref{eq:bern-dgp} and \cref{eq:cnts-dgp}, we compute conditionals $q(\mby= 1 \g \mbx_{\mbv})$ via Monte Carlo approximation.
Due to the different coordinates of $\mbx$ being  independent, one can compute $q(\mby= 1 \g \mbx_{\mbv})$  as a marginal expectation over the inputs except those in $\mbv$:
\[q(\mby\g \mbx_\mbv) = \E_{q(\mbx_{\mbv}^c \g \mbx_\mbv)} q(\mby \g \mbx_\mbv, \mbx_\mbv^c) =  \E_{q(\mbx_\mbv^c)} q(\mby \g \mbx).\]
We Monte Carlo estimate the RHS of this equation over $500$ resamples of $\mbx^c_{\mbv}$.
We take $5000$ samples from each \gls{dgp} to estimate \acrshort{EVAL-X} scores with respect to $q(\mby,\mbx)$. In \Cref{appsec:thresmetrics} we also show experiment results where we use the \evalx{} accuracy and AUROC as the score instead.

\paragraph{\acrshort{REAL-X}.}
We solve \acrshort{REAL-X} for any specified explanation size $K$ as follows.
In the case of the discrete \gls{dgp}, for each possible value of $x\in \supp(q(\mbx))$ (of which there are finitely many), we make $e(\mbx)$ output the subset of at most size $K$ that achieves that maximum averaged log-likelihood over the samples that equal said value $\mbx=x$.
This produces the optimally-scoring explanation $e(\mbx)$ that maps each finite value in the support of $q(\mbx)$ to one subset of the coordinates of $\mbx$.
To do the same in the continuous \gls{dgp} in \cref{eq:cnts-dgp}, we round $\mbx$ to integers and then use the same type of optimization as in the discrete case.

\paragraph{\acrshort{stripe-x}.}
In estimating \encmeas{}, the model $p_\theta(\mbE_\mbv \g \mbx_\mbv, \ell)$ is a decision tree of depth at most 5, which is then used to estimate averaged $\kld$ in the RHS of \cref{eq:klform-detx} with a single sample from $\mby \g \mbx_\mbv$.
The process is repeated for each $\mbv$ and averaged to produce the \encmeas{}.
The simulated experiments were done on a CPU with the whole runtime around $10$ minutes.

\subsection{Experiments with \texorpdfstring{\acrshort{EVAL-X}}{} accuracy and \texorpdfstring{\acrshort{EVAL-X}}{} AUROC}\label{appsec:thresmetrics}

\begin{figure*}
\vspace{-5pt}
\begin{subfigure}[b]{0.48\textwidth}
\centering
    \begin{subfigure}[b]{0.49\textwidth}
    \includegraphics[width=1\linewidth]{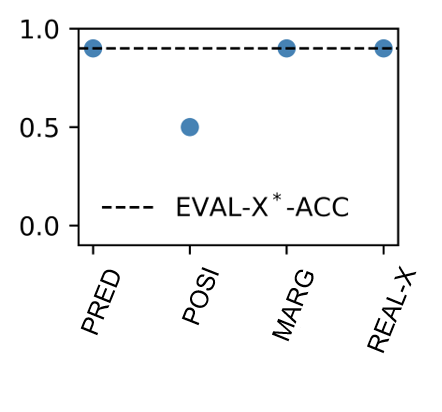}
    \end{subfigure}
    \begin{subfigure}[b]{0.49\textwidth}
    \centering
    \includegraphics[width=1\linewidth]{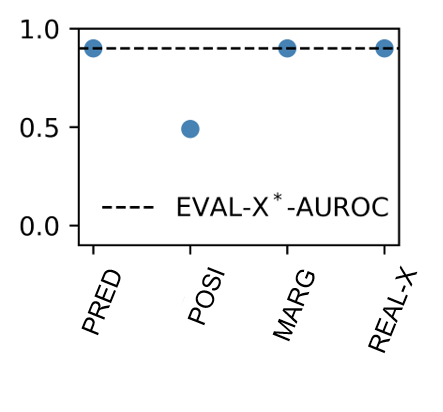}
    \end{subfigure}
\caption{Results: Discrete \gls{dgp}.} 
\label{fig:acc-score-discrete}
\end{subfigure}
\hspace{5pt}
\begin{subfigure}[b]{0.48\textwidth}
\centering
    \begin{subfigure}[b]{0.49\textwidth}
    \includegraphics[width=1\linewidth]{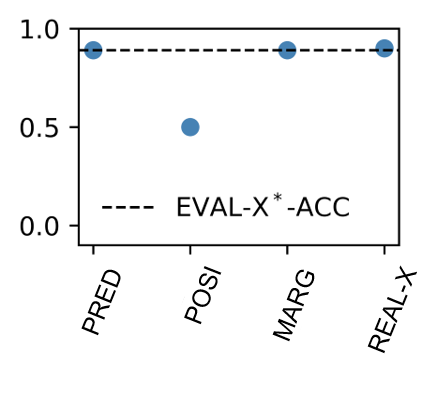}
    \end{subfigure}
    \begin{subfigure}[b]{0.49\textwidth}
    \centering
    \includegraphics[width=1\linewidth]{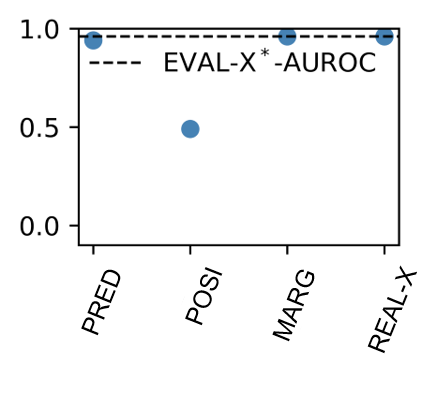}
    \end{subfigure}
\caption{Results: hybrid \gls{dgp}.} 
\label{fig:acc-score-hybrid}
\end{subfigure}
\caption{\small 
The \acrshort{EVAL-X}-ACC and AUROC scores for the different explanations for the discrete \gls{dgp} are on the left and the scores for the hybrid \gls{dgp} are in the right. 
In both, multiple encoding explanations (\gls{predenc}, \gls{margenc}, and the reductive one from \realx{}) all achieve the same score as the corresponding  \acrshort{EVAL-X}$^*$ score.
Thus, ranking metrics like accuracy and AUROC are not sensitive to encoding explanations like \acrshort{EVAL-X} log-likelihoods, and can fail to even weakly detect encoding.
This stems from the fact that accuracy and AUROC only depend on the ranking of the datapoint, and therefore are not sensitive to differences in log probabilities that do not change ranks.}
\label{fig:acc-fails}
\vspace{-5pt}
\end{figure*}

Here, instead of \acrshort{EVAL-X} log-likelihoods, we use the accuracy and AUROC of the \acrshort{EVAL-X} model as the score.
We call these \acrshort{EVAL-X}-ACC and \acrshort{EVAL-X}-AUROC scores.
These metrics only depend on the ranking of the datapoints, and therefore are not sensitive to differences in log probabilities that do not change ranks.
\Cref{fig:acc-fails} shows that, due to this insensitivity, multiple encoding explanations (\gls{predenc}, \gls{margenc}, and the excessively reductive one) all achieve the same score as the corresponding \acrshort{EVAL-X}$^*$ score.
In summary, ranking metrics like accuracy and AUROC are not sensitive to encoding explanations like \acrshort{EVAL-X} log-likelihoods.

\subsection{Classifying dogs and cats.}\label{appsec:catsvdogs}

The \gls{posenc} explanation selects the upper or the lower color patch depending on whether $q(\mby=1\g\mbx) > 0.5$ or not.
The \gls{predenc} explanation selects the patch predicting from which best matches the prediction from $q(\mby=1\g\mbx)$.
The \gls{margenc} explanation selects the top or the bottom image patch based on the color as in the \gls{dgp} in \Cref{fig:margenc}.
We consider two non-encoding explanations.
The first explanation, denoted \texttt{optimal}, selects exactly the features that occur in the \gls{dgp}: $\{\mbx_1, \mbx_2\}$ if the color patch $\mbx_1$ is \texttt{blue} and $\{\mbx_1, \mbx_4\}$ otherwise.
As $\mby$ is determined by the explanation, meaning $\mby\indep \mbE_\mbv \g \mbx_\mbv$ for all $\mbv$ and values $\mbx_\mbv$, this explanation is non-encoding.
The second one, denoted \texttt{fixed}, always outputs the bottom right patch $\mbx_4$; this explanation is constant which violates the first criterion in \Cref{lemma:defeq} meaning there is no encoding.

We also run the \realx{} method from \cite{jethani2021have} to produce an explanation. \realx{} was run over explanations that select one of the four quarter patches and exact marginalization over the selections.

The base cat and dog images were obtained from the \texttt{cats\_vs\_dogs} dataset from the Tensorflow datasets package.
To construct images like in \Cref{fig:margenc}, the color and the two images are sampled independently.
The color being blue/red determines that the label associated with the top/bottom image becomes the label for the constructed image.
The training, validation, and test dataset consist of $8000, 1000,$ and $1000$ samples respectively.

We follow the procedure in \Cref{appsec:cat-estimation} to estimate \detx{}.
The \acrshort{EVAL-X} model is a pre-trained 18-layer residual network. The model $p_\theta(\mbF \g \mbx_\mbv, \ell, \mbv)$ used in computing the \encmeas{} term in \acrshort{stripe-x} (\cref{eq:stripe-x-main}) are 34-layer Residual neural networks.
The \evalx{} model is trained for $100$ epochs with a batch size of $100$ with the Adam optimizer, with the learning rate and weight decay parameters set to $10^{-3}$ and $0$ respectively.
The $p_\theta(\mbF \g \mbx_\mbv, \ell, \mbv)$ model is trained for $50$ epochs with a batch size of $200$ with the Adam optimizer, with the learning rate and weight decay parameters set to $5\times 10^{-5}$ and $1$ respectively.
The  $p_\theta$ model sees variable $\ell$ through an entire extra channel where all the pixels take the value $\ell$.
We used validation loss as the metric to early stop.
The \evalx{} and \detx{} scores are computed on the test dataset.
The cats vs. dogs experiment were done on an A100 GPU where the whole training and evaluation ran in less than 20 minutes.

\paragraph{Remark on the gap between~\fresh{} and \evalx{} scores for the optimal explanation.}
As the optimal explanation selects features sufficient to produce the label, meaning $\mby \indep \mbx \g \mbx_\mbv$ or $\mby \indep \mbx \g \vxex$, \fresh{} and \evalx{} log-likelihoods should be the same as predicting from the full feature set.
One potential reason there is a gap between the two scores in~\cref{tab:real} is that the \fresh{} and \evalx{} scores are computed with ResNet18 models that solve prediction problems of different levels of difficulty
On one hand, \fresh{} is computed with a model trained for a single prediction task: predict $\mby$ from $\vxex$.
On the other hand, \evalx{} is computed with a model trained for a more complicated task: for a range possible $\mbv$, predict $\mby$ from $\xv$.
Using large models with appropriate regularization, such as weight decay, should mitigate the gap in scores.


\subsection{\texorpdfstring{\gls{llm}}{} experiment details}\label{appsec:llmexp}

We generate $10,000$ reviews of the following type: with \texttt{ADJ1} and \texttt{ADJ2} as adjectives, the review is
\begin{itemize}[leftmargin=25pt]
	\item  \texttt{\small 'My day was <ADJ1> and the movie was <ADJ2>. that is it'}  $\quad $ or
	\item  \texttt{\small 'My day was <ADJ1> and the movie was <ADJ2>. oh wait, reverse the adjectives'}.
\end{itemize}
The second sentence in the review acts as a "control flow" input and determines whether \texttt{ADJ1} or \texttt{ADJ2} describes the sentiment about the movie.
We prompted  Llama 3 to predict the sentiment and select words relevant to predicting the sentiment.
In~\Cref{appsec:prompts}, we give the prompts we used to make Llama 3 produce explanations from.
For this problem, the inputs $\mbx$ are the reviews and Llama 3  produces explanations $e(\mbx)$ that select a subset of words in the review.
The summaries and explanations were generated for all $10,000$ samples but to estimate \detx{}, we only used data from the $5$ most common explanations (we restricted to inputs whose explanations $\mbv$ had high $q(e(\mbx)=\mbv)$).
This resulted in a dataset of size $8136$, which we split into a training, validation, and test datasets of sizes $6102,1017,$ and $1017$ respectively.

Both the \evalx{} model and the model for $p_\theta(\mbF \g \mbx_\mbv, \mbv, \ell)$ (see~\Cref{appsec:cat-estimation}) used in estimating the \encmeas{} term in \detx{} were finetuned GPT-2 models.
For the \evalx{} model, we used the AdamW optimizer with a batch size of $100$ and trained for $50$ epochs with the learning rate set to $5e-5$, weight decay set to $0$, and a Cosine learning rate scheduler with the number of cycles set to $1$.
For the $p_\theta$ model used in estimating \encmeas{}, we used the AdamW optimizer with a batch size of $50$ and trained for $25$ epochs with the learning rate set to $5e-5$, weight decay set to $0$, and a Cosine learning rate scheduler with number of cycles set to $1$.
The $p_\theta$ model sees variable $\ell$ through the following word added to the input sequence of words: \texttt{positive} if $\ell=\mby=1$, \texttt{negative} if $\ell=\mby=0$, and nothing if $\ell=\texttt{null}$.
We used validation loss to early stop. 
We follow the procedure in \Cref{appsec:cat-estimation} to compute \encmeas{} with $p_\theta(\mbF \g \mbx_\mbv, \mbv, \ell)$ on the test data with the averaging over $\mby \g \mbx_\mbv$ estimated using a $5$ samples per value of $\mbx_\mbv$.
All training and inference for this experiment was done on an A100. The explanation step and the estimation for both parts of \detx{} together took under 2 hours.
The \gls{llm}-generated explanations achieves an \evalx{} score of $-0.497$ and an \encmeas{} value was $0.114$.

\subsection{Prompts used to predict sentiment and produce explanation}\label{appsec:prompts}
In \Cref{box:summs}, we provide the prompt we used to predict the sentiment from a review and generate an explanation for that prediction.

\vspace{10pt}
\definecolor{shadecolor}{RGB}{245,245,245}
{\captionsetup{hypcap=false}
\captionof{figure}{Llama 3 prompt used to predict sentiment and generate an explanation for that prediction.}\label{box:summs}}
\begin{lstlisting}[backgroundcolor=\color{shadecolor},numbers=none]
System: You are a helpful and honest assistant. Please, respond concisely and truthfully.

You are asked to summarize movie reviews of the form "first sentence. second sentence". 
The following are examples along with the reasoning.


Consider 'My day was moving and the movie was overblown. that is it.'
The second sentence means the second adjective 'overblown' describes the movie.
Due to this description, the sentiment is negative.


Consider 'My day was moving and the movie was overblown. oh wait, reverse the adjectives.'
The second sentence means the first adjective 'moving' describes the movie.
Due to this description, the sentiment is positive.


These are all examples.

user: What is the sentiment about the movie in this review `<REVIEW>`?

Think step-by-step about this latest review. If the second sentence instructs it, switch the adjectives and then based on the new descriptor of the movie, answer either 'positive' or 'negative'. 

Explain why you chose those this sentiment by selecting as few words as possible from the review. Include all the words that you looked at.

Use this helpful format: "the sentiment is <positive/negative> and the explanation is <words, ...>. END. "
\end{lstlisting}

\section{Algorithms}\label{appsec:algorithm-boxes}
\Cref{alg:stripe-x-gen} describes an alternate way to estimate the~\encmeas{} component of \detx{} with a conditional generative model.
\Cref{alg:stripe-x-pred} describes the predictive version of \detx{} estimation, which we used in our experiments.

{\small 
\begin{algorithm}[t]
    \SetAlgoLined
    \KwIn{Training data $D\sim q(\mby, \mbx)$ and test data $D_t\sim q(\mby, \mbx)$, explanation function $e(\mbx)$, penalty weight $\lambda$. 
    \evalx{} model $p_\gamma(\mby \g \mbx_\mbv)$.
	Conditional generative model $p_\theta(\mbx \g \mbx_\mbv, \mby)$ and mutual information estimator that takes two sets as arguments $\texttt{MI}[\{c_i\},\{d_i\}]$;
    }
    \KwResult{Return estimate of \detx{} : }

Define $\mbJ_{(\mbv,\mba)}$ as the set of $K$ random samples of $\mby$ from the \evalx{} model:
\[\mbJ_{(\mbv,\mba)} =\{\hy^k\}_{k\leq K} \quad \{\text{ where } \hy^k \sim p_\gamma(\mby \g \mbx_\mbv=\mba)\}\]

Define $\mbL_{(\mbv,\mba,\mbJ_{(\mbv,\mba)})}$ as the set of $K$ random samples of $\mbx$ from $p_\theta$ conditioned on $\mba$ and $\hy$:
\[\mbL_{(\mbv,\mba,\mbJ_{(\mbv,\mba)})} =\{\mathbbm{1}[e(\hx^k) = \mbv]\}_{k\leq K} \quad \{\text{ where } \hx^k \sim p_\theta(\mbx \g \mbx_\mbv=\mba, \mby = \hy^k)\}\]

Construct the explanation dataset $D_e=\{(\xex=(e(\mbx), \mba))\}$ from $D_t$.

Define $q_{D_e}$ to be the uniform distribution over $D_e$.

Compute the following averaging of estimated mutual information between, $\mbJ,\mbL$
\begin{align*}
\hat{\phi}_q(e) & = 
\E_{(\mbv, \mba) \sim q_{D_e}(\xex)}
\texttt{MI}\left[\mbJ_{(\mbv,\mba)}, \mbL_{(\mbv,\mba,\mbJ_{(\mbv,\mba)})}\right]
\end{align*}

Return $\hat{\phi}(q,e)$ as the \encmeas{} estimate.

     \caption{\encmeas{}, generative version.}\label{alg:stripe-x-gen}
    \end{algorithm}
}

{\small 
\begin{algorithm}[t]
  \SetKwFunction{algo}{\evalx{}}
  \SetKwFunction{proc}{\encmeas{}}
  \SetKwProg{myalg}{Estimate}{}{}
      \SetKwInOut{Input}{Input}
    \SetKwInOut{Output}{Output}
    \SetKwInOut{Define}{Define}
    \SetAlgoLined
    \KwIn{Training data $D\sim q(\mby, \mbx)$ and test data $D_t\sim q(\mby, \mbx)$, explanation function $e(\mbx)$, penalty weight $\lambda$. 
    Specifications for the models $p_\gamma(\mby\g \mbx_\mbv)$ and $p_\theta(\mbF \g \mbx_\mbv, \ell)$.
    }
    \KwResult{Return estimate of \detx{} : }

Define $q_D$ to be the uniform distribution over $D$
    
Construct the explanation dataset $D_e=\{(\mby, \xex)\}$ from $D_t$

Define $q_{D_e}$ to be the uniform distribution over $D_e$.

\myalg{\algo{}}{   
    
    Solve the following minimization problem to learn $p_\gamma(\mby\g \mbx_\mbv)$
     \begin{align*}
        \argmax_\gamma \E_{\mbv \sim q_D(e(\mbx))}\E_{\mbx, \mby\sim q_D(\mbx, \mby)} \big[ \log p_\gamma (\mby \g \mbx_\mbv)\big]
    \end{align*}
    
    \Output{$p_\gamma$}
}
\myalg{\proc{}}{       

	 Construct the set of possible selections $\cV = \{\mbv : q(e(\mbx) = \mbv) 
     > 0 \}$

	Construct data of the form $(\mbx,\mbF)$ where $\mbF = j$ if $\mbE_{\cV[j]}=1$.

    Fit the model $p_\theta(\mbF\g \mbx_\mbv, \ell)$ via the following log-likelihood maximization:
    \begin{align*}
        \argmax_\theta \E_{\mbv \sim q_D(e(\mbx))}\E_{\mbx, \mby\sim q_D(\mbx, \mby)} \sum_{\cV[j]\in \cV}\bigg(\mathbbm{1}[e(\mbx) = \cV[j]] &\big[ \log p_\theta (\mbF = j \g \mbx_\mbv, \ell = \mby, \mbv)  \nonumber
    \\   &+ \log p_\theta (\mbF = j \g \mbx_\mbv, \ell = \texttt{null}, \mbv)\big]\bigg)
    \end{align*}

Define $\cE_{(\mbv, \mba)}$ as the uniform distribution over $K$ samples of $\mby$ from the \evalx{} model:
\[\cE_{(\mbv, \mba)} =\mbU\left[\{\hy\}_{k\leq K}\right] \quad \{\text{ where } \hy^k \sim p_\gamma(\mby \g \mbx_\mbv=\mba)\}\]

\Output{The following nested expectation over $q_{D_e}$ and $\cE(\cdot)$:

\begin{align*}
& 
\E_{(\mbv,\mba) \sim q_{D_e}(\xex)}
\E_{\hy \sim \cE_{(\mbv, \mba)}}
\left(p_\theta(\mbF=j \g \mbx_\mbv, \ell=\hy, \mbv)\log\frac{p_\theta(\mbF=j \g \mbx_\mbv, \ell=\hy, \mbv)}{p_\theta(\mbF=j \g \mbx_\mbv, \ell=\texttt{null}, \mbv)}\right.
\\
	 & 
\qquad \qquad \qquad \qquad \qquad
+  \left.p_\theta(\mbF\neq j \g \mbx_\mbv, \ell=\hy, \mbv)\log\frac{p_\theta(\mbF\neq j \g \mbx_\mbv, \ell=\hy, \mbv)}{p_\theta(\mbF\neq j \g \mbx_\mbv, \ell=\texttt{null}, \mbv)}\right)
\end{align*}
}
}

Learn the \evalx{} model $p_\gamma \leftarrow $\algo{}.

Estimate \encmeas{} as the $\hat{\phi}_q(e) \leftarrow $\proc{}.

Return the following as the \detx{} estimate: 
\[\E_{(\mbv,\mba) \sim q_{D_e}(\xex)}\E_{\mby \sim q_{D_e(\mby \g \xex = (\mbv,\mba))}}\left[\log p_\gamma\left(\mby = \mby \g \mbx_{\mbv} = \mba\right)\right]
 - \lambda \hat{\phi}_q(e)  \]
     \caption{\detx{}, predictive version.}\label{alg:stripe-x-pred}
\end{algorithm}
}

\end{document}